# State-of-the-art AI-based Learning Approaches for Deepfake Generation and Detection, Analyzing Opportunities, Threading through Pros, Cons, and Future Prospects

Harshika Goyal[1], Mohammad Saif Wajid[2†], Mohd Anas Wajid[3†], Akib Mohi Ud Din Khanday[4,5†], Mehdi Neshat[6,8], Amir Gandomi[6,7*]

[1]Indian Institute of Technology, Kharagpur, India.
[2]Department of Computer Science and Intelligent Systems, Tecnol´ogico de Monterrey, Monterrey , Mexico.
[3]Institute for the Future of Education, TEC de Monterrey, Monterrey, Mexico.
[4]Department of Computer Science and Software Engineering, United Arab Emirates University, Al Ain, Abu Dhabi, UAE.
[5]Department of Computer Science, Samarkand International University of Technology, Samarkand, Uzbekistan.
[6]Faculty of Engineering and Information Technology, University of Technology Sydney, Ultimo, Sydney, 2007, NSW, Australia.
[7]University Research and Innovation Center (EKIK), Obuda University, Budapest, 1034, Hungary.
[8]School of Computer and Mathematical Sciences, University of Adelaide, Adelaide, 5005, Australia.

*Corresponding author(s). E-mail(s): gandomi@uts.edu.au;
Contributing authors: harshikagoyal@kgpian.iitkgp.ac.in ; a00831364@exatec.tec.mx ; anas wajid@tec.mx; akib.khanday@siut.uz; mehdi.neshat@uts.edu.au; †These authors contributed equally to this work.

**Abstract**

The rapid expansion and advancement of deepfake technology, which is specifically designed to create incredibly lifelike facial imagery and video content, has ignited a remarkable level of interest and curiosity across a multitude of fields, including but not limited to the realms of the entertainment industry, forensic analysis, cybersecurity and the innovative creation of digital characters. By harnessing the latest breakthroughs in deep learning methodologies, especially through the implementation of techniques such as Generative Adversarial Networks, Variational Autoencoders, Few-Shot Learning Strategies, and Transformer Neural Networks, the outcomes achieved in the generation of deepfakes have been nothing short of astounding and transformative. Furthermore, the recent advancements in diffusion models have served to invigorate and catalyze ongoing research efforts in this fascinating domain. At the same time, the ongoing evolution of detection technologies is being developed to counteract the potential for misuse associated with deepfakes, effectively addressing critical concerns that range from political manipulation to the dissemination of fake news and the ever-growing issue of cyberbullying. This comprehensive review paper meticulously investigates the most recent developments in both deepfake generation and detection, including around 400 publications, providing an in-depth analysis of the cutting-edge innovations that are shaping this rapidly evolving landscape. Starting with a thorough examination of systematic literature review methodologies, we embark on a journey that delves into the complex technical intricacies inherent in the various techniques used for deepfake generation, comprehensively addressing the challenges faced, potential solutions available, and the nuanced details surrounding manipulation formulations. Besides, task definitions are standardized to ensure clarity, while datasets and metrics are diligently introduced, paving the way for a deeper understanding of emerging technologies poised to impact the field. The subsequent sections of the paper are dedicated to accurately benchmarking leading approaches against prominent datasets, offering thorough assessments of the contributions that have significantly impacted these vital domains. Ultimately, we engage in a thoughtful discussion of the existing challenges that persist in the field and outline potential future research trajectories for both deepfake generation and detection, thereby paving the way for continuous advancements in this critical and ever-dynamic area of study.

**Keywords:** Deepfake Technology, Generative Adversarial Networks, Deep Learning, Diffusion Models, Detection Technologies, Cybersecurity, Ethical Implications

# 1 Introduction

Deepfake technology has emerged as a groundbreaking advancement in the realm of synthetic media, driven by the convergence of Artificial Intelligence (AI) and Machine Learning (ML). The term ”deepfake” itself is a portmanteau of ”Deep Learning” (DL) and ”fake,” encapsulating the essence of this transformative technology (Rana et al., 2022). At its core, a deepfake refers to content that is convincingly generated by an AI system, designed to appear genuine in the eyes of human observers. The rise of deepfake technology has been nothing short of remarkable, gaining significant momentum since its emergence in 2017 (Westerlund, 2019). Rapid developments in DL techniques, coupled with the availability of vast amounts of visual data, have paved the way for increasingly sophisticated and realistic deep fakes. Consequently, the implications and consequences of this technology have garnered widespread attention and sparked debates surrounding its potential applications, as well as its malicious misuse. Deepfake technology originated in the academic and research communities but quickly found its way into various aspects of society, such as entertainment, digital art, and historical preservation (Seow et al., 2022) These beneficial applications have showcased the potential of deepfakes to boost creativity, create immersive virtual experiences, and enable new forms of expression.

Traditionally, the limitations imposed on the realism and sheer volume of facial alterations were predominantly due to the absence of sophisticated editing tools, the necessity for specialized skills and knowledge, as well as the intricate and time-consuming nature of the entire process involved in making such modifications. To illustrate, an early pioneering study (Bregler et al., 2023) achieved the remarkable feat of synchronizing a person’s lip movements with an entirely unrelated audio track by ingeniously correlating the phonetic elements of the audio with the respective facial shapes that corresponded to those sounds. However, in a striking transformation over the past few years, the pace of advancements in this field has seen an extraordinary surge, propelling us into an era where the automatic creation of nonexistent faces or the manipulation of a person’s visage in photographs or videos has become significantly more accessible and straightforward. This remarkable shift can be attributed to two key factors: first, the widespread availability of extensive public datasets that provide a rich resource for training models, and second, the revolutionary progress in deep learning methodologies, including innovations such as Autoencoders (AE) and Generative Adversarial Networks (GANs) (Kingma, 2013; Goodfellow et al., 2014), which have dramatically reduced the reliance on labour-intensive manual editing processes. Consequently, as a result of these technological advancements, a plethora of user-friendly software applications and mobile platforms, like ZAO (de Seta, 2021) and FaceApp (Neyaz et al., 2020), have emerged onto the scene, allowing individuals from all walks of life, regardless of their technical expertise or background, to effortlessly create and generate deceptive images and videos that can easily fool the unsuspecting observer. Thus, we find ourselves at a fascinating intersection of technology and creativity, where the possibilities for visual manipulation seem virtually limitless, transforming the way we perceive and interact with digital content in our daily lives. In this ever-evolving landscape, it is imperative to remain aware of the implications and ethical considerations surrounding the use of such powerful tools, as they redefine the boundaries of authenticity in our increasing digital world.

However, there is a darker side to deepfake technology. Malicious individuals have recognized its power and are using deepfakes to deceive, manipulate, and spread misinformation. The ease of creating deepfakes combined with their highly realistic appearance has raised concerns about the potential for significant social and political disruptions, violations of personal privacy, and a decline in trust towards media and information sources.

## 1.1 Benefits of Deepfake Technology

When harnessed for ethical purposes, deepfake technology, which involves the use of AI to create realistic video and audio simulations, possesses remarkable capabilities and can provide a plethora of advantages in various sectors. In the realm of education (Hwang et al., 2021), deepfakes have the potential to revolutionize the learning landscape by authentically simulating historical figures (Turan, 2021). This would enable students to interact with virtual representations of influential characters, thus deepening their grasp of diverse historical eras. These digital reenactments can foster engaging experiences that render learning more captivating and unforgettable, offering insights into cultural, social, and political frameworks (Pantserev, 2020) that conventional educational approaches might fail to communicate.

Deepfakes have emerged as a valuable tool in the entertainment industry (Putra et al., 2024), aiding in the creation of immersive visual experiences. They enable filmmakers to seamlessly blend live-action footage with computer-generated elements or digitally resurrect beloved actors for new performances. This technology has significantly reduced production costs and saved time Hasani et al. (2024) while ensuring visual consistency in films and television series. More importantly, deepfake technology empowers creators to push the boundaries of artistic expression, enabling them to explore innovative visual aesthetics and experiment with unconventional narrative techniques.

Deepfake technology also holds exciting prospects in the realm of personalized content. It can tailor educational resources to fit individual learning styles, thereby enhancing the educational journey. In the sphere of virtual reality (VR) and augmented reality (AR), deepfakes can enhance immersion by crafting lifelike avatars and interactions, making VR/AR experiences more credible and interactive (Aliman and Kester, 2022). However, it's important to note that deepfake technology also raises significant ethical concerns, particularly in the context of misinformation and privacy. As this technology continues to evolve, it's crucial to consider these risks and work towards responsible use and regulation. Furthermore, the capacity of deepfakes to evoke nostalgia and provide solace is another intriguing aspect. For instance, it can resurrect legendary figures, such as Freddie Mercury (Westerlund, 2019), permitting audiences to relive the charm and aura of their favorite icons in a digital setting. This has implications beyond entertainment, extending into therapeutic environments were engaging with a virtual likeness of a cherished individual could offer emotional reassurance. As this technology continues to evolve, the potential uses of deepfakes are poised to grow, unveiling fresh avenues for innovation across a multitude of industries.

## 1.2 Malicious Applications of Deepfake Technology

The malicious use of deepfake technology poses a significant threat due to its potential for wide-scale impact and the various ways it can be misused. One concerning aspect is its exploitation on social media platforms, where the rapid spread of conspiracies, rumors, and misinformation can lead to severe consequences. Deepfakes have been deployed as a tool for propaganda, political manipulation, and character assassination. By manipulating videos or images, malicious actors can fabricate false evidence, undermine trust in public figures, and sow discord among communities.

In 2018, a widely circulated deepfake video featuring Barack Obama served as a notable example of the potential impact of deepfakes. In the video, Obama appeared to be warning about the dangers associated with deepfakes themselves (Chesney and Citron, 2019). Another notorious example of malicious use occurred in late 2017 when an anonymous Reddit user harnessed DL techniques to replace the faces of individuals in pornographic videos with the faces of unsuspecting individuals, creating photo-realistic fake videos. This not only violated the privacy and consent of the victims but also highlighted the potential for deepfakes to be exploited for revenge, harassment, or blackmail.

Deepfakes pose significant risks, as they can be exploited for nefarious purposes such as tarnishing individuals' reputations, coercing or blackmailing them, influencing elections, inciting conflict, spreading fake speeches to fuel political or religious unrest, disrupting financial markets, and facilitating identity theft (Gu¨era and Delp, 2018; Diakopoulos and Johnson, 2021; Pantserev, 2020; Figueira and Oliveira, 2017; Zhou and Zafarani, 2020; Kietzmann et al., 2020). The prevalence of malicious applications far outweighs the beneficial ones. Recent advancements have made it increasingly easy to create deepfakes from still images, leading to their widespread misuse by cybercriminals (Zakharov et al., 2019).

Furthermore, the proliferation of deepfakes and facial manipulations exacerbates issues with Automated Face Recognition Systems (AFRS). Studies indicate alarmingly high error rates in AFRS performance when faced with deepfakes, morphing, makeup manipulation, partial face tampering, digital beautification, adversarial examples, and Generative Adversarial Network (GANs) (Korshunov and Marcel, 2019; Scherhag et al., 2017; Rathgeb et al., 2021; Majumdar et al., 2019; Ferrara et al., 2013; Yang et al., 2021; Colbois et al., 2021). Similarly, the accuracy of automated speaker verification significantly decreases under adversarial examples (Huang et al., 2021). These findings underscore the urgent need for robust solutions to mitigate the negative impacts of deepfakes.

## 1.3 Evolution of Deepfake Technologies

The journey of developing techniques for detecting deepfakes, which has been marked by significant fluctuations and transformations throughout the years, can be seen in Figure 2. In the initial stages of this endeavour, the primary focus of detection methods revolved around pinpointing visible artefacts and inconsistencies that could be detected in manipulated videos (Pei et al., 2024), which proved to be quite effective against those low-quality deepfakes that were more rudimentary in nature. Nevertheless, as the technology behind the deepfake generation progressed and became increasingly sophisticated, enabling the creation of highly realistic and believable content (Zhang, 2022), many of the earlier detection methods fell into obsolescence, rendered ineffective against these new, high-quality challenges (Patel et al., 2023).

Subsequent years saw a rapid rise in deepfake quality and the need for detection methods. From 2019 to 2021, innovations such as DeepFaceLab (Perov et al., 2020), GAN inversion techniques, and detection models like EfficientNet (Coccomini et al., 2022; Cunha et al., 2024) and ResNet (Ahmed et al., 2022; Min-Jen and Cheng-Tao, 2024) were introduced. During this period, large-scale awareness also increased, prompting initiatives like Facebook's Deepfake Detection Challenge (DFDC) (Juefei-Xu et al., 2022).

The emergence of cutting-edge models such as Multi-Attention Transformers (Waseem et al., 2023; Li et al., 2023) (MAT) and frequency-based detectors (Jeong et al., 2022; Tan et al., 2024a), exemplified by innovations like F3-NET, signified a turning point in this ongoing struggle. These advanced methodologies provided more robust and accurate detection capabilities by meticulously focusing on subtle inconsistencies and employing complex temporal analysis techniques. By 2022, advancements included using Transformers (Coccomini et al., 2022) and Attention mechanisms (Ganguly et al., 2022), Multimodal analysis (Wang et al., 2022), and Diffusion-based models (Ulhaq and Akhtar, 2022; Chung et al., 2022) for improved realism. In 2023, real-time deepfake detection (Mittal et al., 2024; Lanzino et al., 2024) became a focus, alongside methods to improve model robustness, such as adversarial training and integrating explainable AI (XAI) (Mathews et al., 2023) techniques. In 2024, the integration of Neural Radiance Fields (NeRF) (Gao et al., 2023; Li et al., 2023) and the development of collaborative frameworks, combined with a focus on adversarial robustness and longitudinal analysis, continued the push toward more effective detection and management of deepfake technology. Yet, despite these advancements, deepfake creation techniques continued to evolve rapidly, frequently outsmarting detection efforts by exploiting the vulnerabilities and weaknesses present in existing models and frameworks. Recently, innovative approaches have emerged, such as multimodal detection strategies that seamlessly integrate visual, audio, and temporal clues into the analysis process. These have made remarkable strides in addressing these persistent challenges. However, the relentless advancement of generative models ensures that the ongoing battle between the forces of creation and detection remains a dynamic and ever-evolving challenge, characterized by periodic advancements and setbacks on both sides as they vie for dominance in this digital landscape. Ultimately, the interplay between these two opposing forces continues to shape the future of media authenticity and the integrity of digital content.

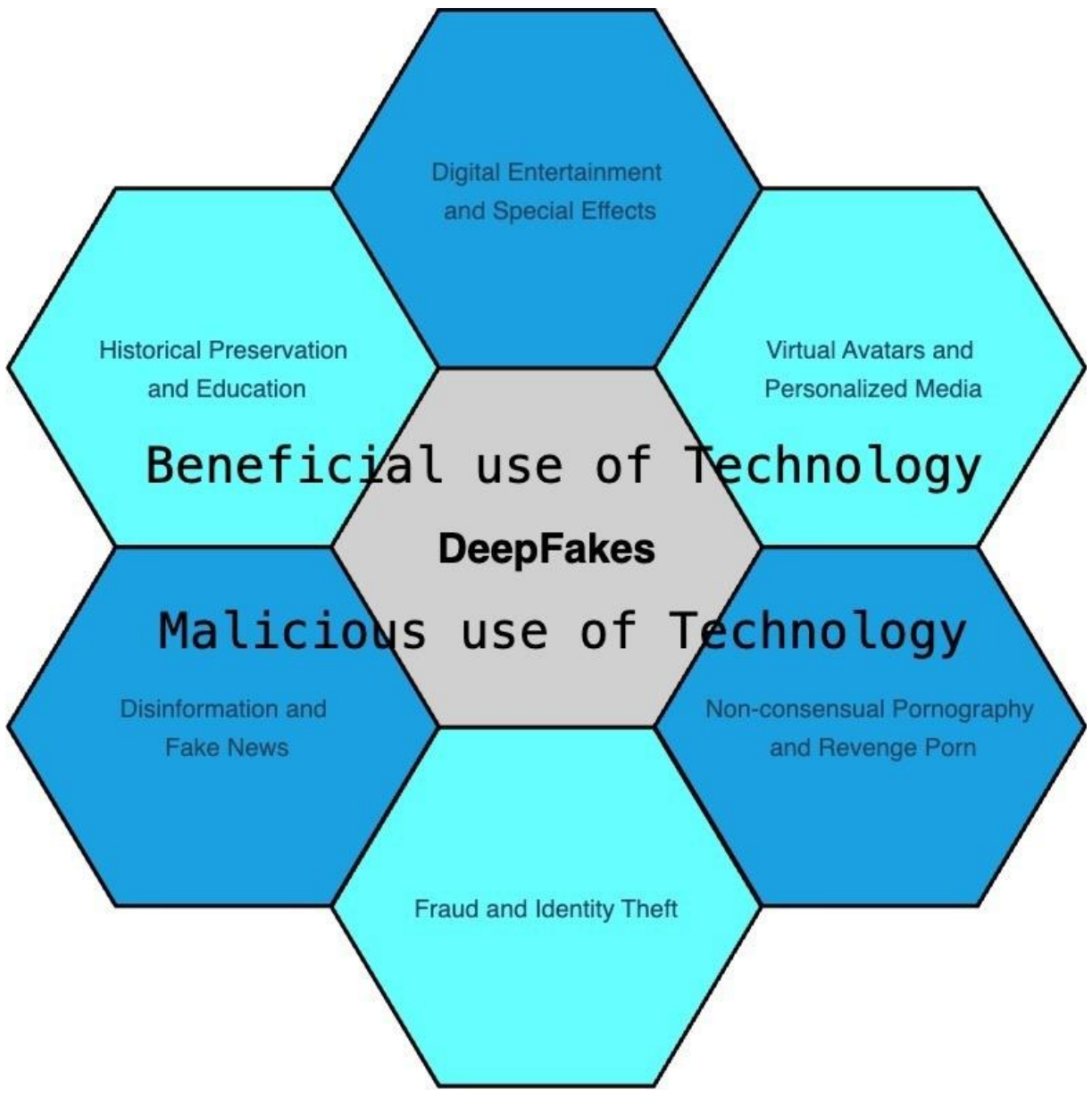

**Fig. 1**: Beneficial and Malicious use of Technology

**Fig. 2**: Evolution of Deepfake technologies over the last two decades, highlighting key advancements in generation and detection techniques from 2014 to 2024

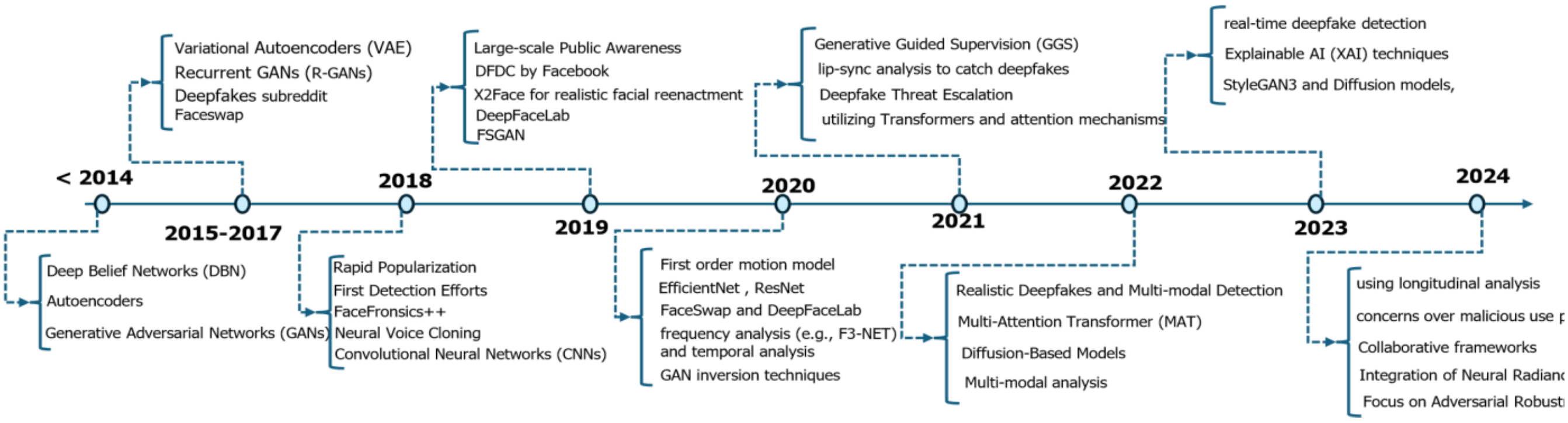

## 1.4 Major contributions

The significant contributions of this review paper are summarized as follows:

- Comprehensive Survey of Deepfake Generation and Detection Techniques: The paper presents an extensive overview of the current methods for both deepfake generation and detection. Techniques like Generative Adversarial Networks (GANs), Variational Autoencoders, Few-Shot Learning Strategies, Transformer Neural Networks, and diffusion models are explored, providing an in-depth understanding of the state-of-the-art technologies used in the deepfake field.
- Unification of Task Definitions and Standardization: This review plays a crucial role in unifying various definitions related to deepfake tasks, ensuring clarity and understanding in the field. The introduction of standardized datasets and metrics is a significant step, critical for benchmarking and comparing the performance of different approaches.
- Comprehensive Dataset and Metric Comparison: This review evaluates a wide range of datasets and metrics used in training and benchmarking deepfake generation and detection models. The detailed comparison provided here is a valuable resource, aiding researchers in choosing suitable datasets for their studies.
- In-Depth Analysis of Related Techniques: This review provides a thorough investigation of multiple mainstream deepfake technologies, including face swapping, face reenactment, talking face generation, and facial attribute editing. It also discusses related fields such as head swapping, face super-resolution, face reconstruction, body animation, and adversarial detection.
- Ethical and Responsible Use Discussion: The paper includes an essential discussion on ethical implications and responsible use of deepfake technology. This section highlights the risks of misuse, such as political manipulation, misinformation, and privacy invasion, emphasizing the need for ethical standards and policies.
- Benchmarking and Evaluation of Recent Innovations: The review benchmarks leading approaches against prominent datasets, evaluating their efficacy. It also pays special attention to diffusion-based models, which have shown significant advances in recent years.
- Integration of Explainability in Detection Models: The review discusses the importance of explainability in deepfake detection models using tools like LRP and LIME. This integration aims to build trust in the technologies and ensure their responsible use in various domains.
- Comprehensive Discussion of Challenges and Solutions: It identifies and discusses the challenges faced in deepfake generation and detection, along with potential solutions for these issues, contributing to the development of more robust models.

- Identification of Future Research Trajectories: The review outlines potential research directions for the field of deepfake technology, inspiring advancements in generalizability, ethical standards, advanced detection techniques, and collaboration across industries. These contributions make paper a valuable resource for advancing the understanding, development, and responsible use of deepfake technologies.

# 2 Research Methodology

## 2.1 Analysis Framework and Approach

In order to develop this systematic and comprehensive review (SLR) study in the realm of deepfake generation and detection, it entails a series of meticulous steps that we followed and are essential for ensuring a thorough and insightful analysis as follows.

- First and foremost, we established the research objectives with absolute clarity, delineating the precise questions and goals that the review work seeks to explore within the realm of deepfake generation and detection. This involves pinpointing the specific facets of deepfake technology that will be scrutinized and evaluated.
- Following the delineation of research objectives, the next step involves an extensive literature search. We delved deep into academic databases, search engines, and specialized repositories to unearth a wide array of research papers, articles, and conference proceedings pertinent to the domain of deepfake generation and detection. This exhaustive search is imperative to ensure the review work is founded on a comprehensive and up-to-date understanding of the existing literature.
- Subsequently, we established stringent inclusion and exclusion criteria that will serve as guiding principles in selecting studies for the review. These criteria should be intricately linked to the research objectives and may consider factors such as publication dates, research quality, relevance to the topic, and the methodologies employed in the studies.
- Once the criteria are in place, the study selection process commences, wherein the researcher meticulously screens handpicks studies that align with the predefined inclusion criteria. We focused on abstracts and summaries and scrutinized them to ascertain the suitability of each study for inclusion, with reasons for exclusion being meticulously documented to ensure transparency and reproducibility in the review process.
- Following the selection of studies, the next phase involved extracting pertinent data from these chosen sources. This includes a thorough examination of the research methodologies utilized, the datasets employed, the techniques for generating deepfakes, the detection algorithms, the evaluation metrics applied, and the outcomes obtained. We applied a systematic and organized approach to data extraction, which is crucial for facilitating subsequent analysis and comparison. With the extracted data in hand, we embarked on an in-depth analysis to discern common themes, trends, and patterns prevalent in the various approaches and methods employed for deepfake generation and detection. Studies with similar characteristics or research focuses were grouped together, and qualitative analysis techniques such as thematic analysis or content analysis may be employed to unearth critical insights and findings. • In the subsequent steps in the research methodology, we considered Critical evaluation and synthesis wherein the reviewed approaches and methods' strengths, weaknesses, and limitations are meticulously scrutinized. The reliability, effectiveness, and robustness of different detection techniques are critically assessed, and a synthesis of the findings from the analyzed studies is conducted, shedding light on significant discoveries and identifying gaps in the existing body of literature.
- Upon completion of the analysis, we proceeded to draw conclusions and formulate recommendations based on the key findings derived from the review. These findings have implications for the field of deepfake generation and detection, and suggestions for future research directions, potential enhancements in techniques, and areas warranting further exploration are provided.

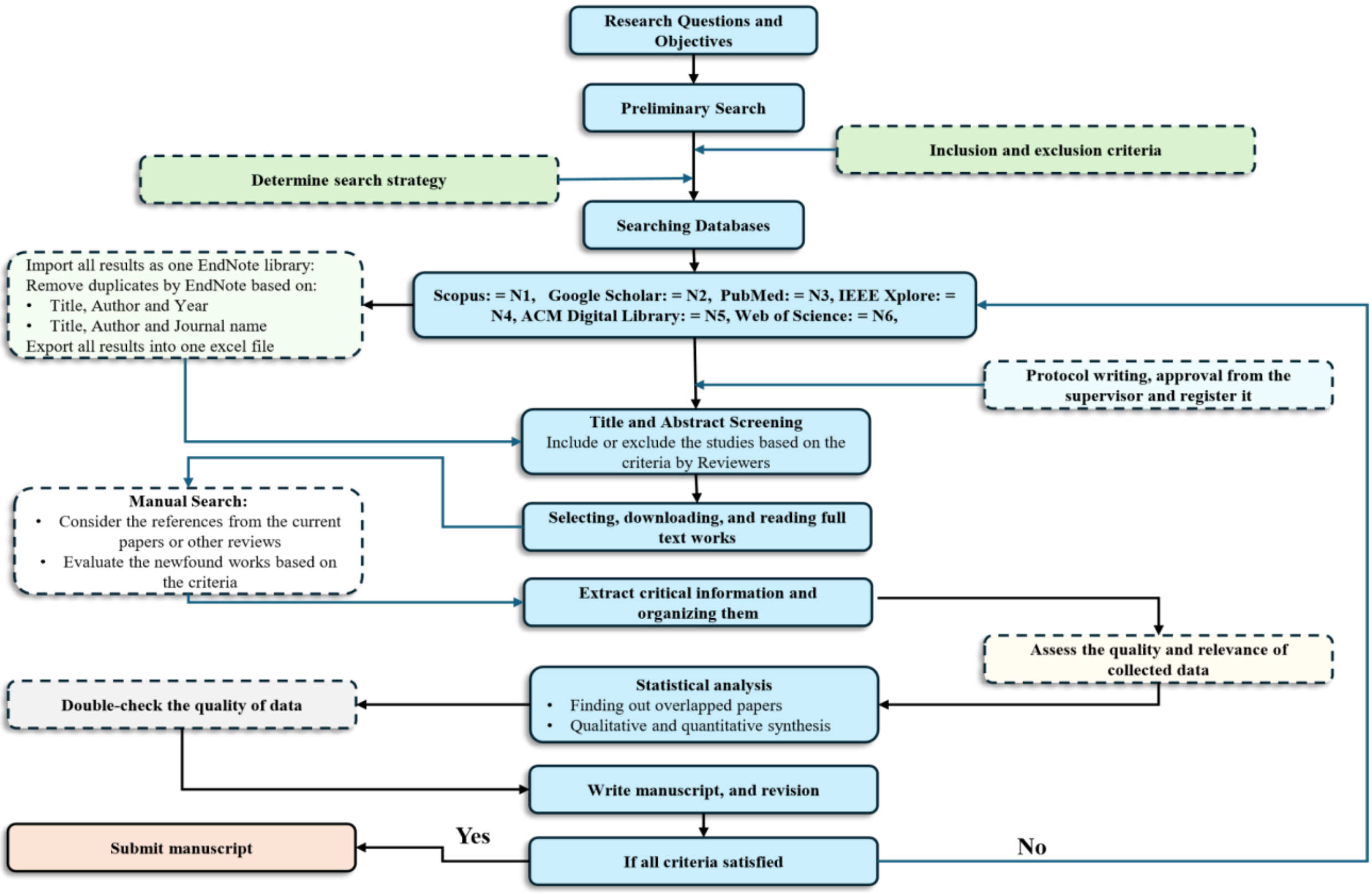

**Fig. 3**: Step by step developing the systematic review (SLR) of detecting and generating deepfake researches.

- In the subsequent stages, we were involved in the composition and revision of the review work. A straightforward, concise, and well-structured narrative is crafted in adherence to the conventions of academic writing. Multiple rounds of revision were undertaken to ensure that the review work was coherent, accurate, and readily comprehensible to the target audience.

Figure 3 shows the steps of conducting this systematic review research for deepfake detection and generation works.

## 2.2 Research Objectives

The objective of the review paper is to provide a comprehensive examination of the current state of deepfake generation and detection methods, focusing specifically on the role of deep learning approaches. The primary goals are to identify, distinguish, assess, and evaluate all relevant studies in the domain of deepfake technology. By categorizing and analyzing the most impactful techniques used for both deepfake generation and detection, the paper aims to establish a clear understanding of their advantages and drawbacks. Additionally, the review seeks to explore challenges, propose potential solutions, and offer insights into future directions for advancing the field of deepfake research. A few Research Questions (RQs) that have been defined are as follows: • *What are the main techniques used for deepfake generation and detection?*

Discussed in Section 3 (Deepfake generation techniques, challenges, and solutions) and Section 4 (Deepfake detection techniques, challenges, and solutions).

- *What are the challenges faced in deepfake generations, and how can they be overcome?*
  Discussed in Section 3.6 (Challenges and solutions in Deepfake generations), which includes details on challenges such as visual fidelity, temporal coherence, and the need for broad generalization, along with proposed solutions like data augmentation and advanced loss functions.
- *What ethical implications and concerns arise from the use of deepfake technologies?*
  Discussed in Section 1.2 (Malicious use of Deepfake Technology), which covers issues like misinformation, character assassination, and the decline of trust in media.
- *How can detection models be improved to generalize across various deepfake types and datasets?*
  Discussed in Section 4 (Deepfake detection techniques, challenges, and solutions). The paper emphasizes using domain adaptation, transfer learning, and enhancing the diversity of training data to improve generalizability.
- *What role do explainability techniques play in building trust in deepfake detection models?*
  Discussed in Section 5 (Deepfake detection techniques, challenges, and solutions). Techniques such as Layer-wise Relevance Propagation (LRP) and Local Interpretable Model-agnostic Explanations (LIME) are suggested to enhance transparency and trustworthiness.
- *What datasets and metrics are used in training and benchmarking deepfake generation and detection models?*
  Discussed in Section 6 (Analysis of Key Datasets and Their Characteristics) and throughout Section 7, with mentions of popular datasets and metrics such as those used in the Deepfake Detection Challenge (DFDC).
- *What are the advantages of using ensemble learning techniques for deepfake detection?*
  Discussed in Section 4.1 (Deep learning techniques and applications). Ensemble learning techniques, including the use of hybrid models, are highlighted for their ability to improve accuracy and robustness against different types of manipulations.
- *What are the laws and policies regarding Deepfake Technology in the world?*
  Discussion in Section 4 of the review paper. This section provides a comprehensive overview of the various laws and policies adopted by different countries to regulate deepfake technology, including notable legislation in the United States, United Kingdom, European Union, Singapore, India, and Australia

These sections provide comprehensive insights into the significant questions related to deepfake generation, detection techniques, challenges, and advancements.

## 2.3 Article Selection

This study's article search, and selection procedure are separated. Based on the keywords and terms used to search publications in the initial step, as shown in Figure 2, referenced by Table 1. This collection includes papers found using an electronic database search. The electronic databases used include Google Scholar, Scopus, IEEE, ACM, Springer, Elsevier, Emerald Insight, Taylor & Francis, Wiley, Peerj, and MDPI. Other discoveries include journals, conference articles, books, chapters, notes, technical studies, and special issues.

With the remarkable advancements in DL techniques over the past decade, there has been a significant surge in research studies dedicated to detecting, manipulating, and synthesizing media, including images, videos, and audio. Figure 5(a) illustrates a compelling upward trend in the proliferation of deepfake related research among scholars, showcasing an astounding growth rate of 471% from 2020 to 2023.

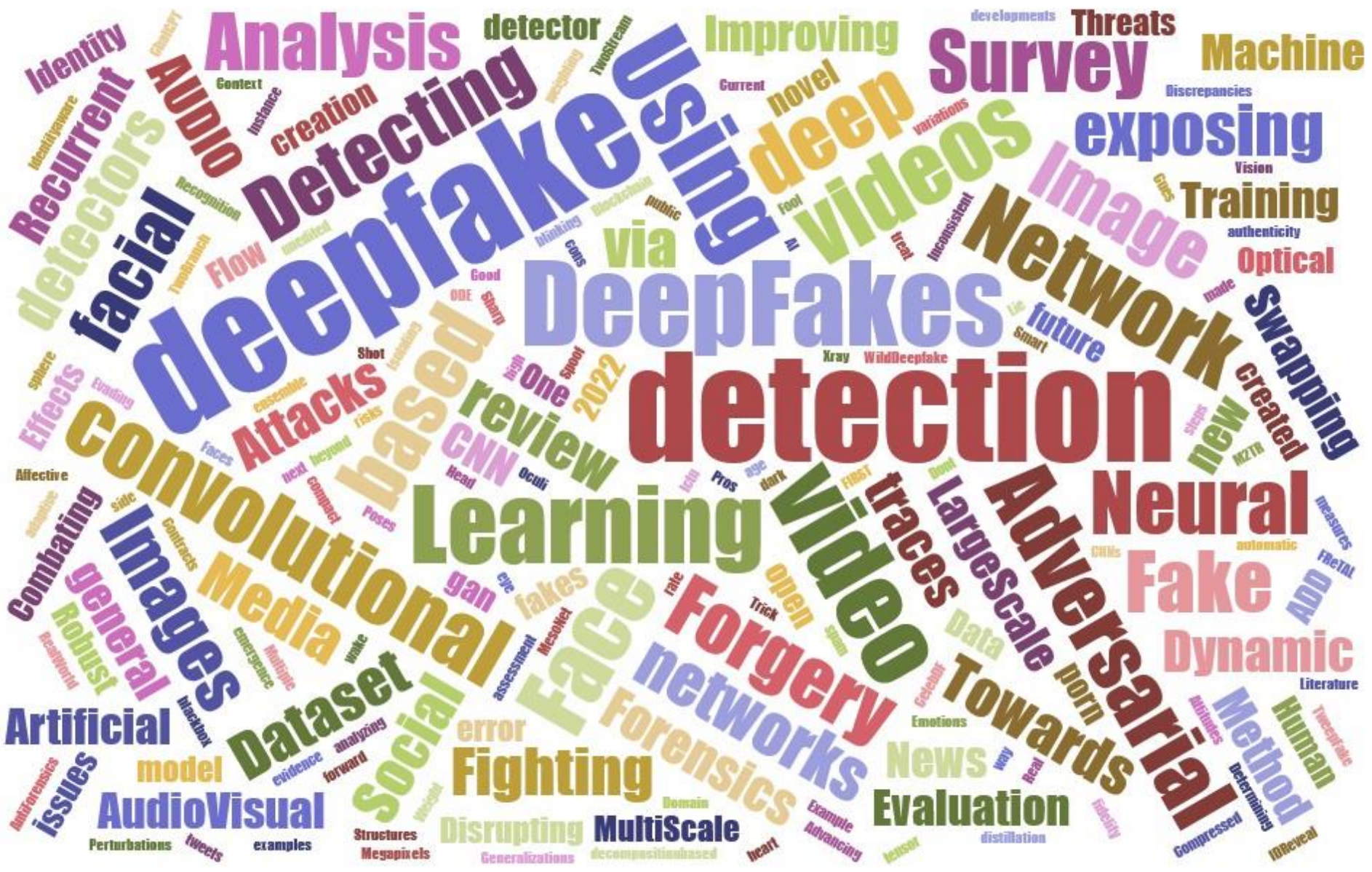

**Fig. 4**: Word Clouds analysis regarding Deepfake Detection and Generation

| |
|---|
| Generative Adversarial Networks, Deepfake Generation, Deepfake Detection 16 Deep learning and Deepfake Generation, Autoencoders, Machine learning and Deepfake detection, Face Swap, Deep learning and Image forensic, |
| Lip Syncing, Neural network and Deepfake detection, |
| Voice Cloning, AI methods deepfake detection or Artificial intelligence deepfake |
| , Transfer Learning, Deep transfer learning and Deepfake |
| , Forensic Analysis, Transfer learning and Deepfake detection |
| , Biometric Verification, Image forensic |
| , Liveness Detection, Neural network Deepfake detection |
| , Image and Video Analysis, Video Forensic |
| , Digital Watermarking, Capsule Networks |
| , Blockchain for Media Authenticity, Explainable AI |
| , Adversarial Attacks, Multimodal Analysis |

**Table 1**: Deepfake Detection and Generation Searching Keywords and Terms

Notably, conference papers and articles constituted the primary modes of publication for deepfake works, accounting for 55% and 28%, respectively, as depicted in Figure 5(b).

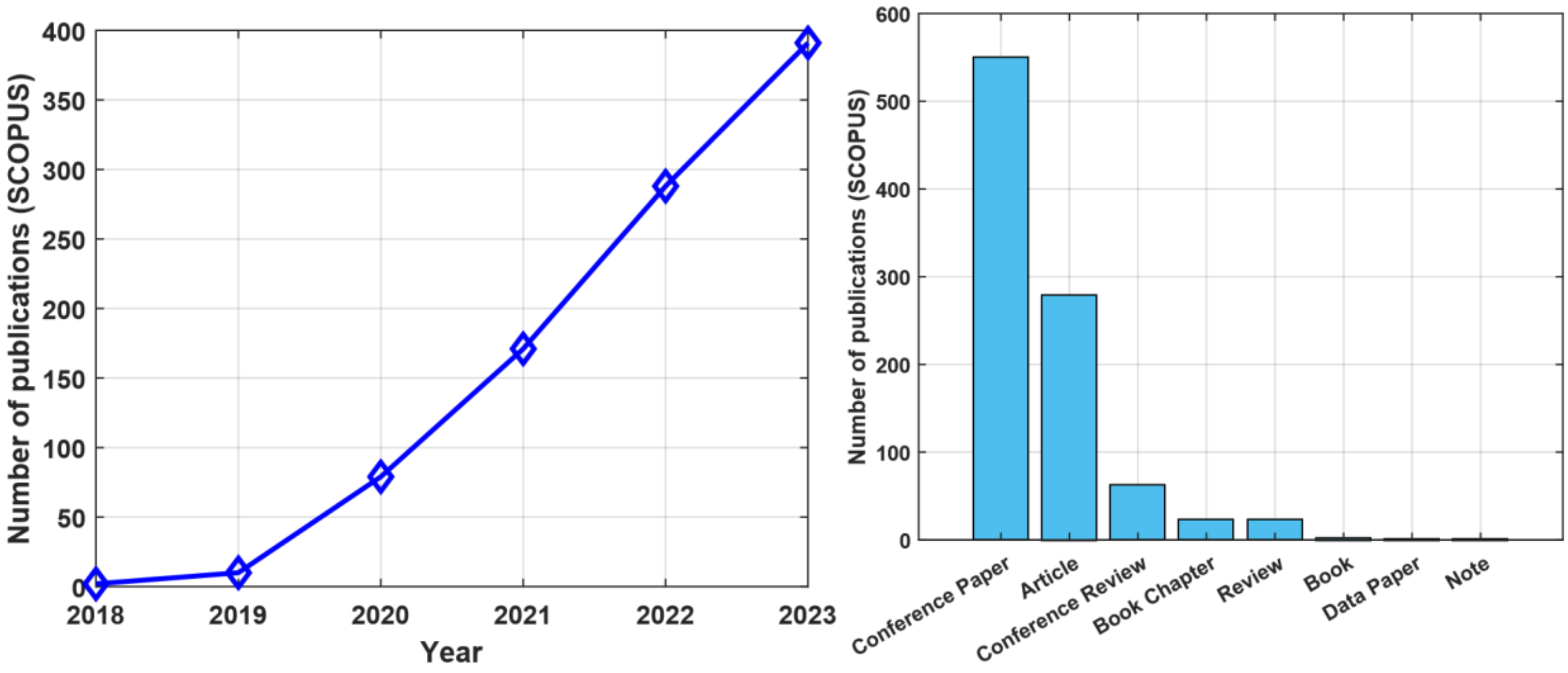

(a) (b)

**Fig. 5**: Distribution of studies related to deepfake detections based on listed on Scopus from 2018 to 2023, and (b) the type of publications

The analysis of deepfake studies reveals the presence of 20 pioneering research centres, as depicted in Figure 6(a). Notably, the Chinese Academy of Sciences stands out prominently, showcasing a distinct focus on deepfake research compared to other universities and institutes. Moreover, Figure 6(b) provides a comparative representation of the top 20 publications that have contributed to disseminating deepfake findings. It is noteworthy that Springer and ACM play a significant role in publishing and sharing advancements in this domain.

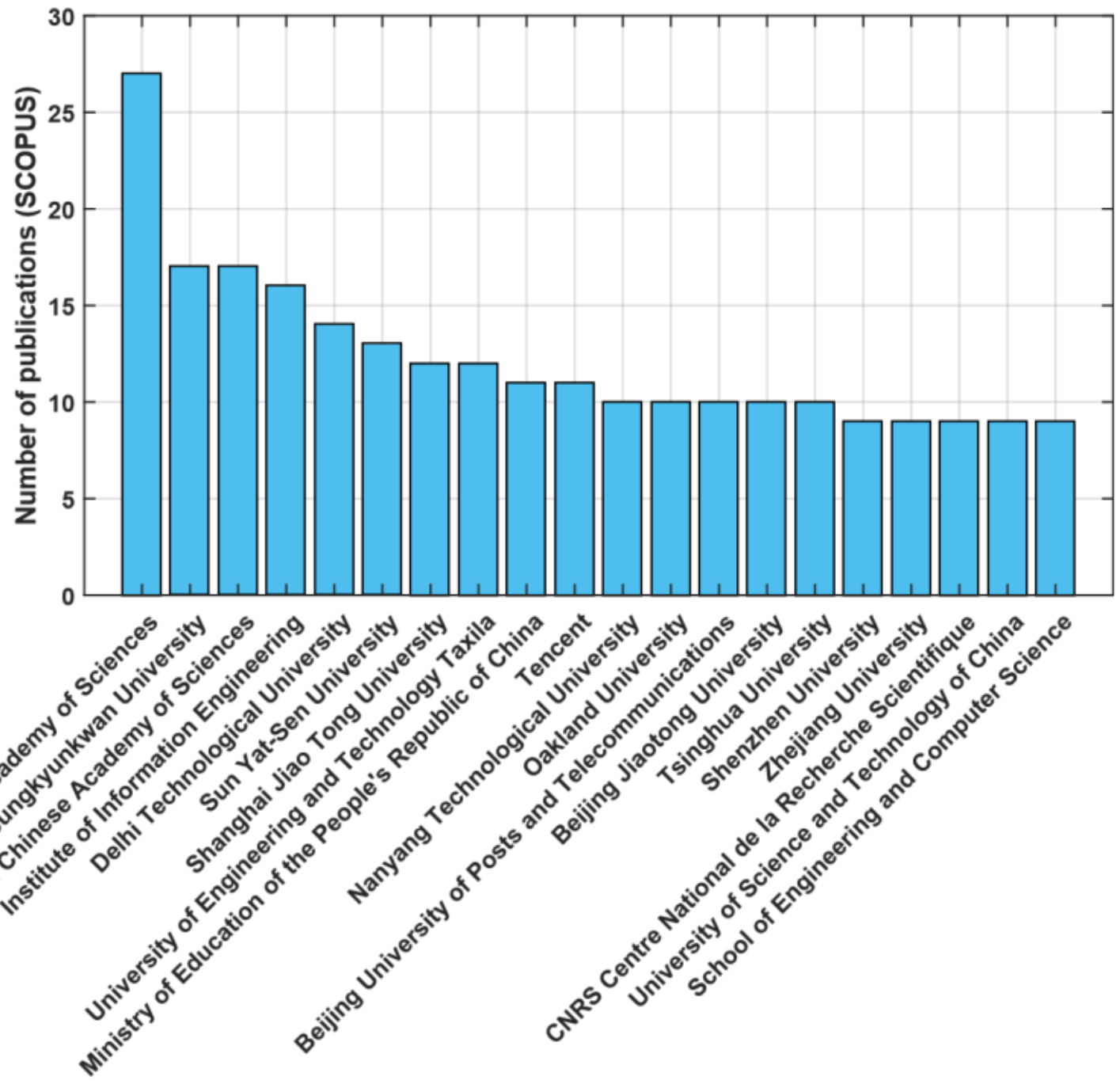

(a)

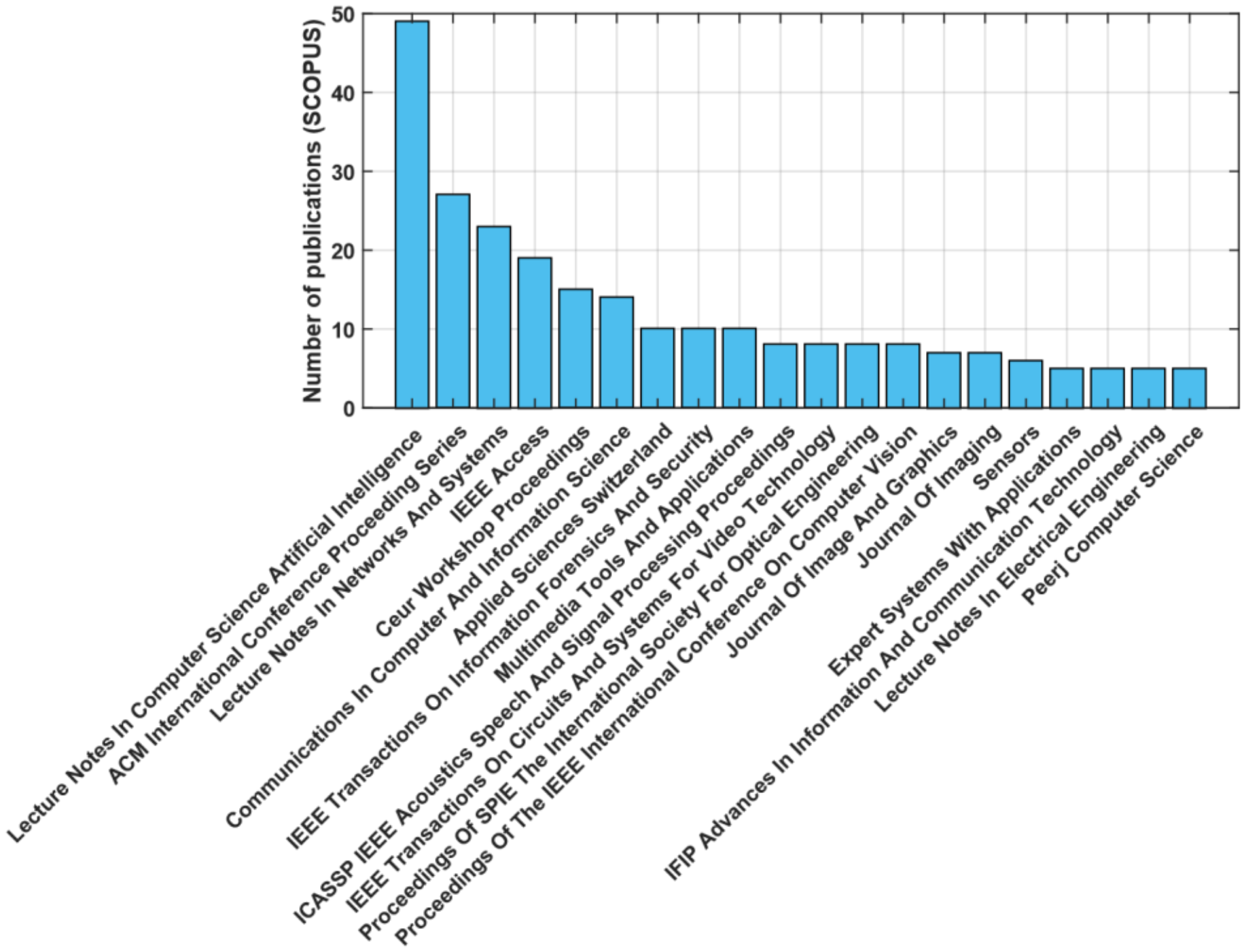

(b)

**Fig. 6**: The relevant publications number listed on SCOPUS based on the most popular (a) affiliations and (b) sources, including top journals, conferences, etc.

As anticipated, China, India, and the United States emerged as the top three countries pioneering the development and research in the field of deepfake. From 2018 to 2023, Scopus lists 210, 190, and 150 works in deepfake research for China, India, and the United States, respectively (refer to Figure 7(a)). Furthermore, various disciplines contribute to the advancement of deepfake research, but the three most significant ones are Computer Science, Engineering, and Mathematics. Figure 7(b) provides statistical insights into the distribution of subjects involved in driving research in deepfake.

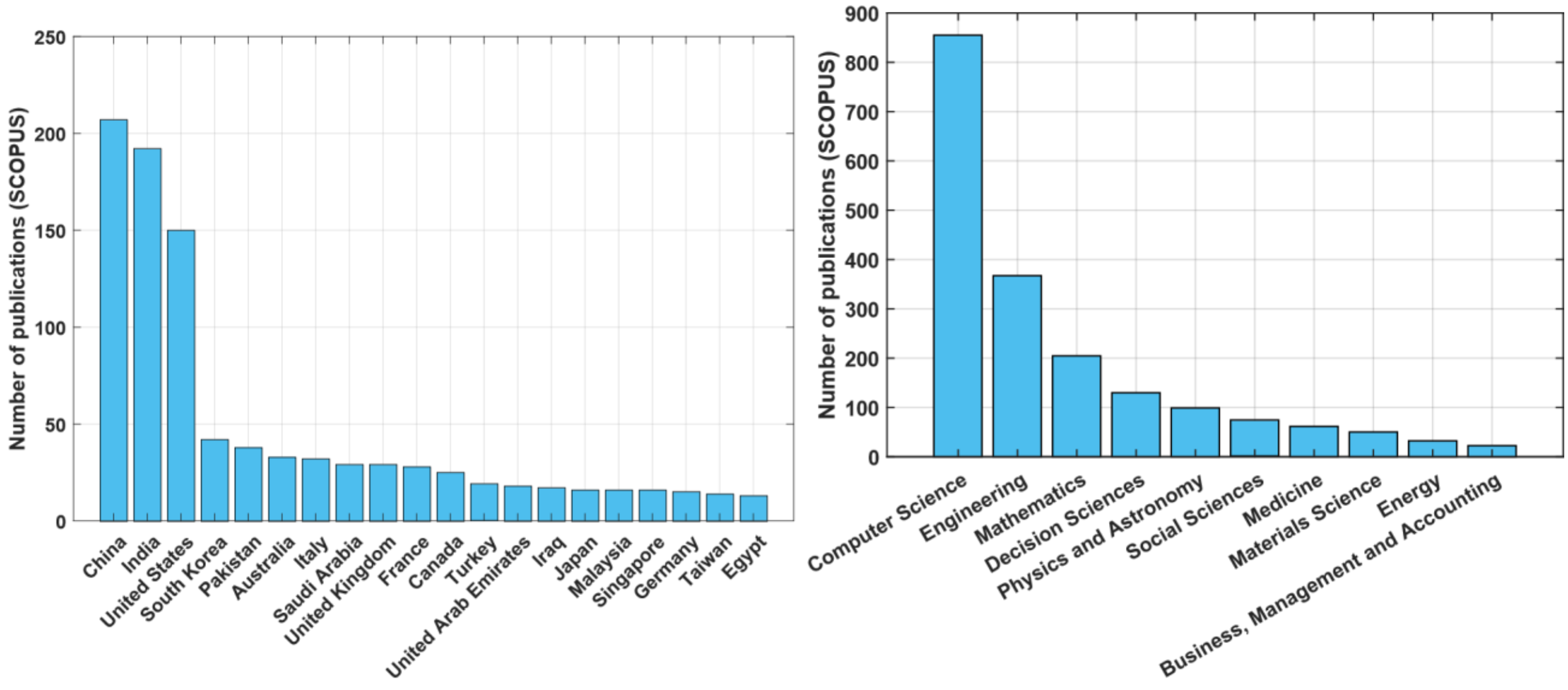

**Fig. 7**: The relevant publications of deepfack detection numbers listed on SCOPUS based on (a) countries and (b) subjects.

In Figure 8, a comprehensive network visualization depicts the extensive landscape of deepfake research relationships, categorizing them based on technique, detection accuracy, and authors. Colours convey distinct meanings or attributes of the nodes and edges in the network, facilitating visual grouping and identification. Specifically, nodes representing techniques or methodologies belonging to the same category are assigned the same colour, aiding in visual grouping and facilitating easy identification of different technique types. This meticulously designed colour scheme instils confidence by visually organizing the network and distinguishing between different categories of techniques. Additionally, the edges in the network are coloured based on the strength and intensity of the relationships between connected techniques or methodologies. The colour coding of the edges provides insights into the level of association or similarity between the connected nodes. More intense or vibrant colours can represent stronger relationships, while weaker relationships can be depicted using lighter or more subdued colours.

Figure 9 presents a visualization that captures the intricate relationships among authors in the field of deepfake research based on the number of publications over the past five years as documented in Scopus. The visualization highlights the strong connections between authors, indicating how much they cite each other's works. However, it also reveals the presence of specific authors who form clusters with minimal connections to the works of others. Further technical assessments are recommended to ensure a comprehensive evaluation of the disconnected research clusters. One possible explanation for these isolated clusters could be the availability of specific publications that do not provide the full text of the works. In such cases, the limited accessibility of certain research papers may hinder the extent of citation and collaboration between authors. Understanding and addressing the challenges in deepfake research is of utmost importance. To gain a deeper understanding of these disconnected clusters, it is essential to explore the reasons behind their limited interaction and investigate whether it is due to restricted access to relevant publications. Addressing these challenges and promoting open access to research findings may facilitate greater collaboration and knowledge sharing among authors in the field of deepfake research, potentially leading to significant advancements in the field.

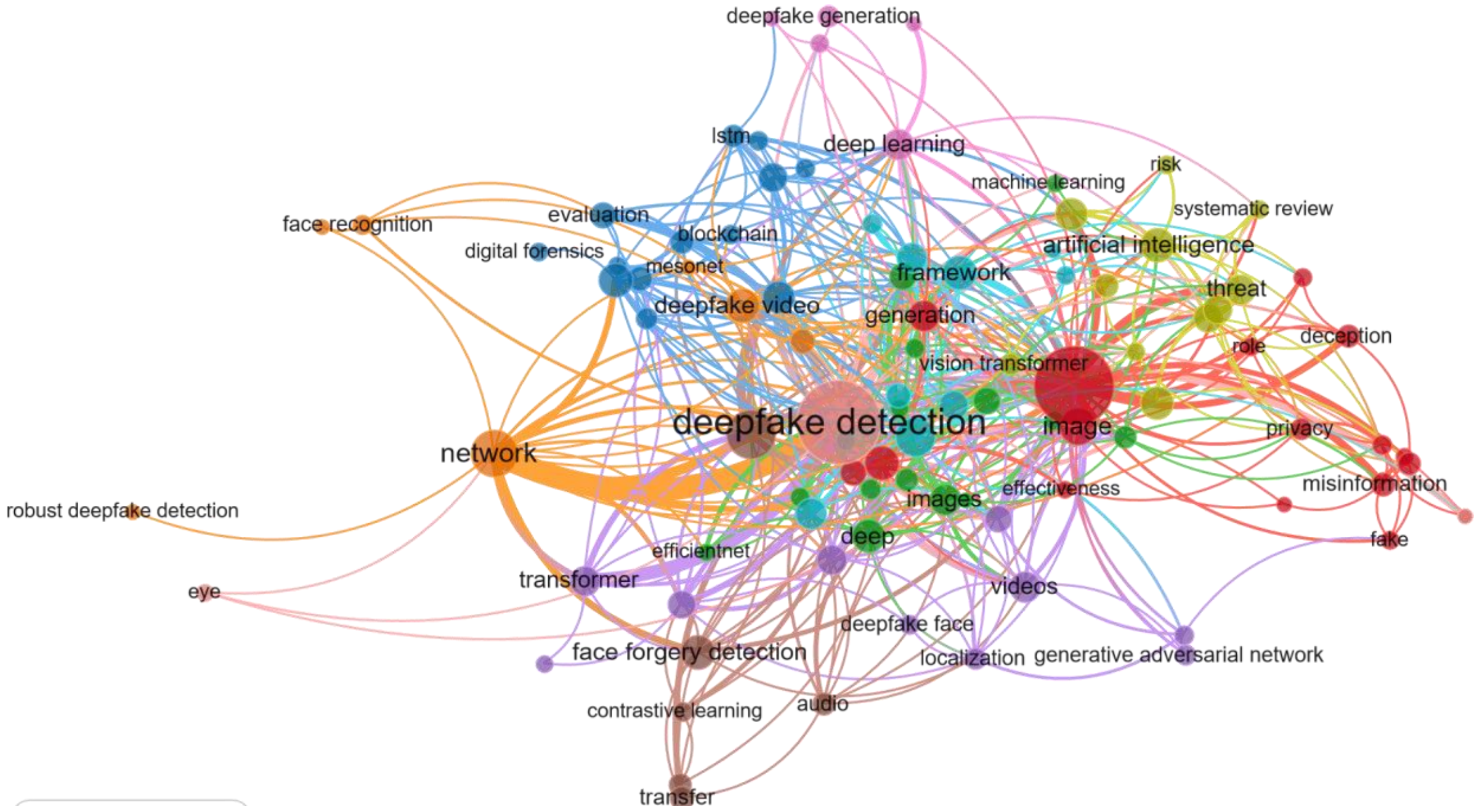

**Fig. 8**: Network visualization showcasing the relationships among various techniques and methodologies employed in the review of deepfake detection.

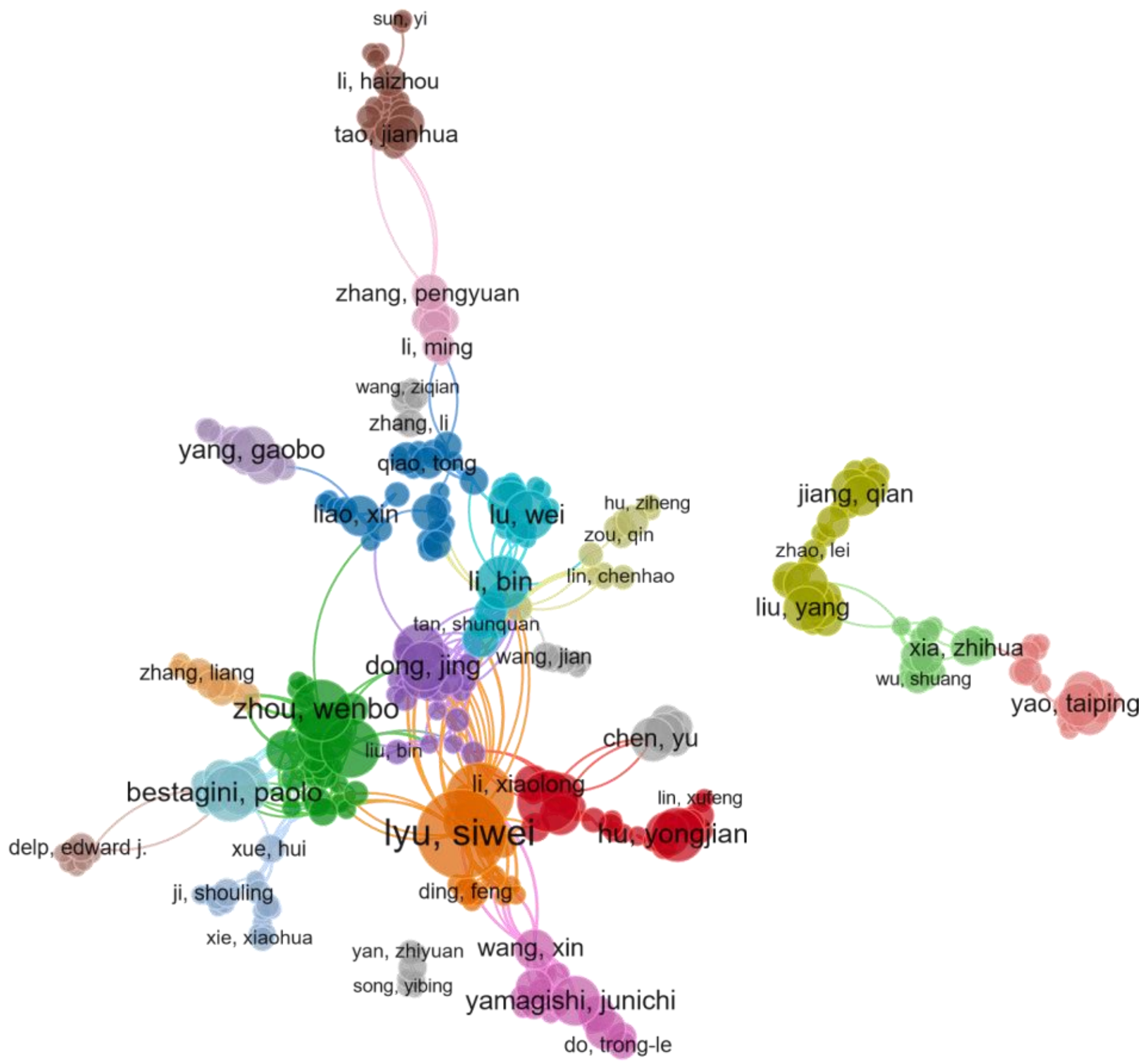

**Fig. 9**: Network visualization showcasing the relationships among authors and co-authors considering the number of publications in the field of deepfake detection.

In the context of citation impact and author relationships within the research background, Figure 9 highlights the significant clusters that emerge. It is observed that authors with a higher number of connections exhibit more robust clusters and edges, indicating their prominence and influence within the network. These authors tend to receive more citations compared to authors in isolated clusters. Visualization provides insights into the citation impact and interconnections among authors, confirming the expected pattern where well-connected authors, through their extensive networks, tend to accumulate more citations. The strength of the clusters and edges reflects the significance of relationships and the level of influence within the research community, highlighting the value of their work. By identifying these major clusters, researchers can gain a better understanding of the key contributors and influential figures in the field. This information not only enlightens but also guides further studies, collaborations, and knowledge dissemination within the deepfake research domain, providing a clear direction and purpose for future research.

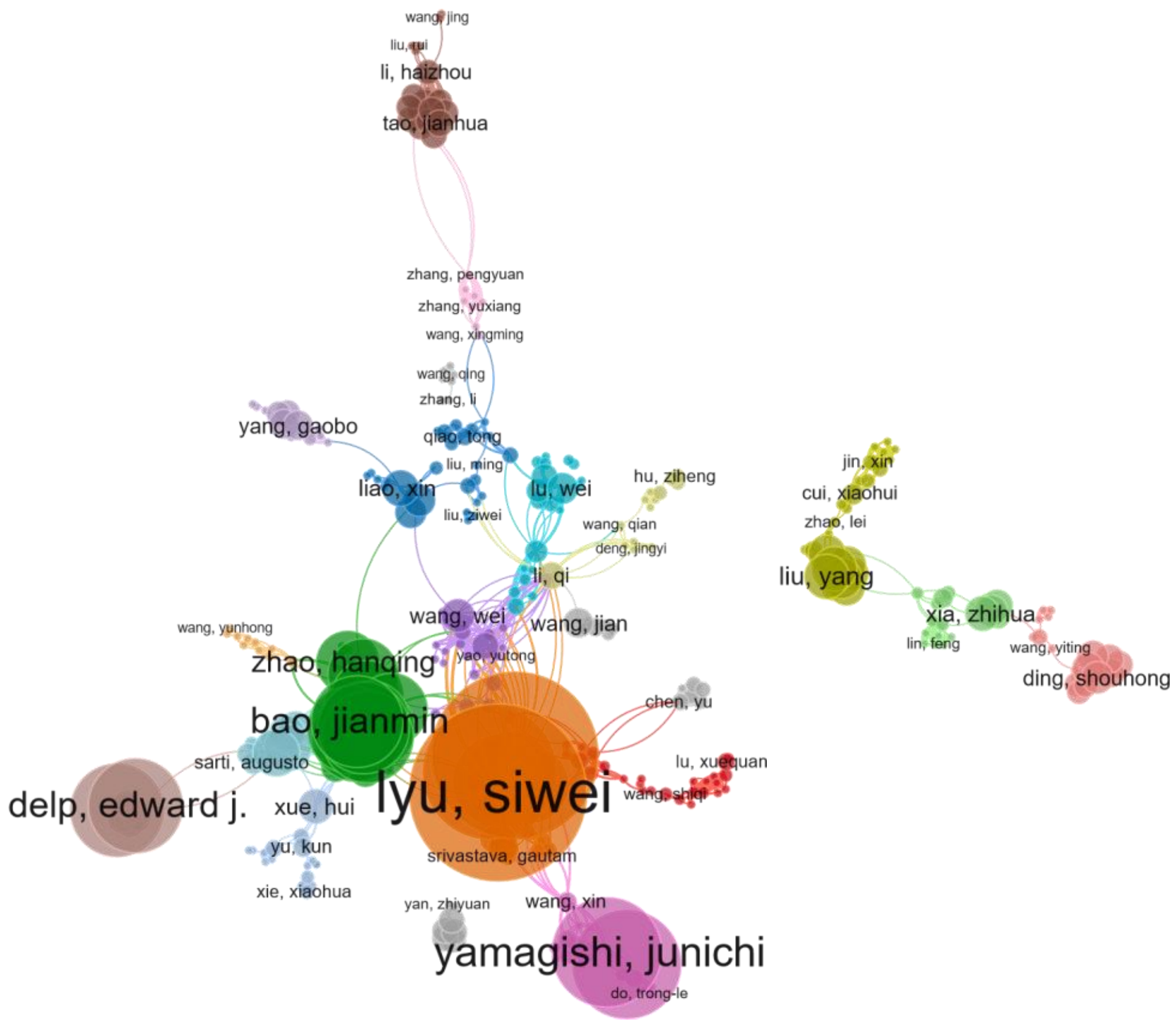

**Fig. 10**: Network visualization showcasing the relationships among authors and co-authors considering the citation number in the field of deepfake detection.

# 3 Deepfake generation techniques, challenges and solutions

Deepfake creation involves the utilization of advanced techniques rooted in AI and DL algorithms. Several methods have emerged as prominent approaches in this field, including GANs (Singh et al., 2020), autoencoders (Khalid and Woo, 2020), variational encoders (Zendran and Rusiecki, 2021), and deep neural networks (Seow et al., 2022). Figure 3 comprehensively depicts the general mechanisms employed by GANs, autoencoders, and variational encoders for deepfake generation. GANs, consisting of a generator and discriminator network, engage in an adversarial learning process to generate realistic deepfakes. Variational encoders adopt a probabilistic approach to learning facial feature distributions, facilitating the generation of plausible deepfakes through sampling. Deep neural networks, with their deep and complex architectures, play a fundamental role in learning intricate patterns and nuances, enabling the creation of highly realistic deepfakes (Rana et al., 2022). These methods enable the synthesis of highly realistic and convincing deepfakes, raising both opportunities and concerns. In this section, we delve into the specifics of each technique, highlighting their contributions to deepfake generation.

In recent years, there has been a proliferation of online tools that facilitate the generation of deepfakes with remarkable realism and ease of use. Among these tools, FaceSwap (Kohli and Gupta, 2021), DFaker++ (Shao et al., 2024), Faceswap-GAN (Lu, 2018), DeepFaceLab (Perov et al., 2020) and StarGAN (Choi et al., 2018) have gained popularity among users due to their ability to manipulate images and videos in impressive ways. These platforms employ sophisticated ML algorithms and deep neural networks to seamlessly swap faces, alter expressions, and transform identities, often producing highly convincing results. Users can upload their media to these platforms and apply various DL-based techniques to create deepfakes. One of the most popular tools is also ”faceswap-GAN” (See Figure 11), an innovative creation by Shaoan Lu (Lu, 2018) that harnesses the power of a denoising autoencoder in conjunction with adversarial losses and attention mechanisms to achieve face-swapping. This cutting-edge tool is crafted to produce lifelike face exchanges in images and videos, leveraging sophisticated methodologies from DL, including GANs and attention mechanisms. Figure 11 shows the detailed architecture of faceswap-GAN, encoder, decoder and discriminator component. Intricate facial expressions may encompass gestures with sharp angles, unique lighting conditions, obstructions, or poses that aren’t directly facing the viewer. These factors can create hurdles for face-swapping algorithms, potentially resulting in distortions or imprecisions in the swapped faces (Nirkin et al., 2019). Highly intricate facial expressions, particularly those showcasing multiple emotions at once or swift transitions between expressions, can be incredibly daunting for face-swapping technologies. Effectiveness may diminish in such circumstances, yielding less believable outcomes. Figure 12 shows various Faceswap-GAN performance samples, including easy, extreme facial expressions and hard models.

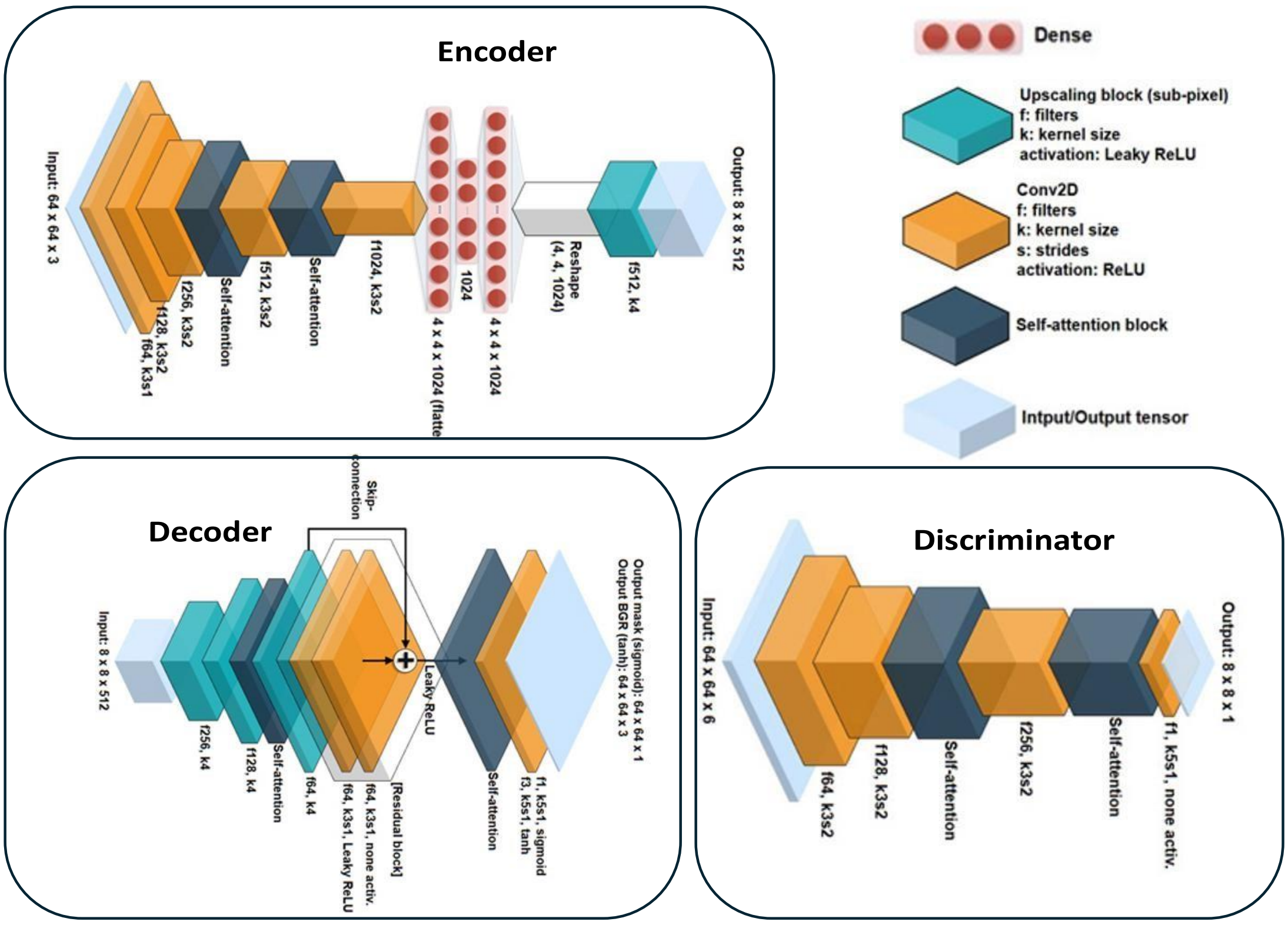

**Fig. 11**: Generative adversarial networks architecture of Faceswap-GAN for face swapping (Adapted from Lu (2018)) (Originally published on GitHub under a CC0 license to distribute)

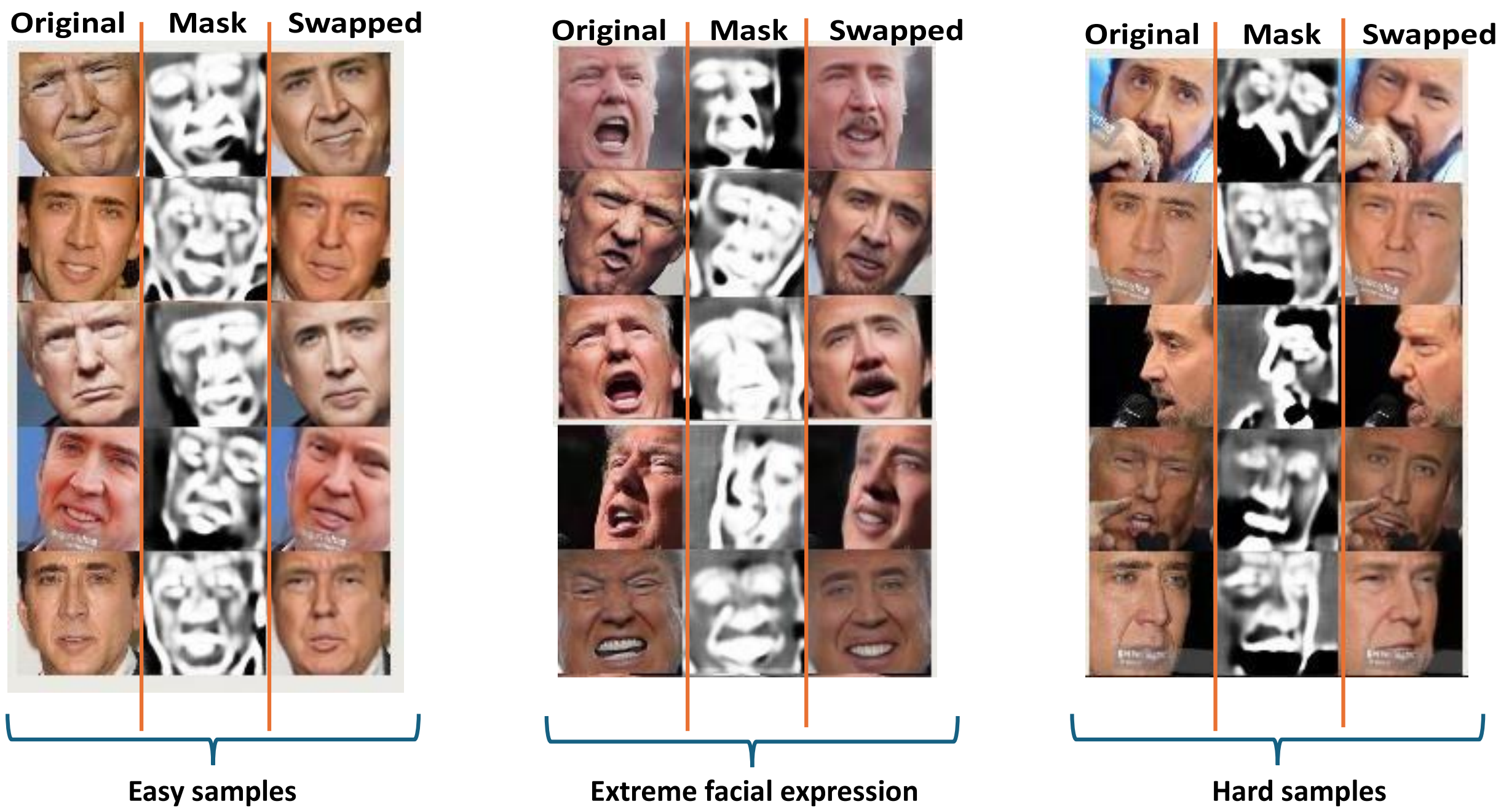

**Fig. 12**: The performance of Faceswap-GAN for face swapping in different levels of difficulties (Adapted from Lu (2018)) (Originally published on GitHub under a CC0 license to distribute)

## 3.1 Generative Adversarial Networks (GANs)

The creation of deepfakes has been greatly facilitated by the advent of GANs. GANs operate through a dual network structure consisting of a generator and a discriminator. The generator network generates synthetic content, while the discriminator network is responsible for distinguishing between real and fake data. Through an iterative and adversarial learning process, the generator strives to improve its ability to produce highly realistic deep fakes. At the same time, the discriminator becomes increasingly proficient at identifying any inconsistencies that may arise. Figure 13 depicts the components of a GAN model in developing Deepfake images (Remya Revi et al., 2021). Hence, both networks engage in a minimax competition, which can be mathematically represented by the following value function detailed in the text (Remya Revi et al., 2021).

$$\min_G \max_D V(D,G) = \mathbb{E}_{x \sim p_{data}(x)}[\log D(x)] + \mathbb{E}_{n \sim p_n(n)}[\log(1 - D(G(z)))] \quad (1)$$

where $p_{data}(x)$ represents the data distribution of authentic images $(x)$ and $p_n(n)$ stands for the distribution of noise signals $(n)$. Once the essential training is accomplished, the generator will gain the ability to produce synthetic images that look convincingly real by utilizing the noise elements $(n)$. At the same time, the discriminator's skill in differentiating between fake and authentic images will be sharpened. The next segment explores the various types of GANs employed in the creation of synthetic images.

In a relevant study, the authors provided valuable insight into applying GANs and their integration with other technologies in deepfake creation. This reference offers figures illustrating the architecture and combination of GANs with convolutional neural networks (CNNs), recurrent neural networks (RNNs), and other architectures. This integration of GANs with other advanced technologies has revolutionized the field of deepfake creation, enabling the generation of highly convincing and visually coherent synthetic media (Mirsky and Lee, 2021). The fabrication of videos using autoencoders and GANs is explored in (Khalil and Maged, 2021), where two autoencoders are employed to learn the features of the source and target images, followed by reconstruction using the source image's decoder. Additionally, the paper mentions using DFDNet, a DL image enhancement method, to enhance the quality of the generated deepfake.

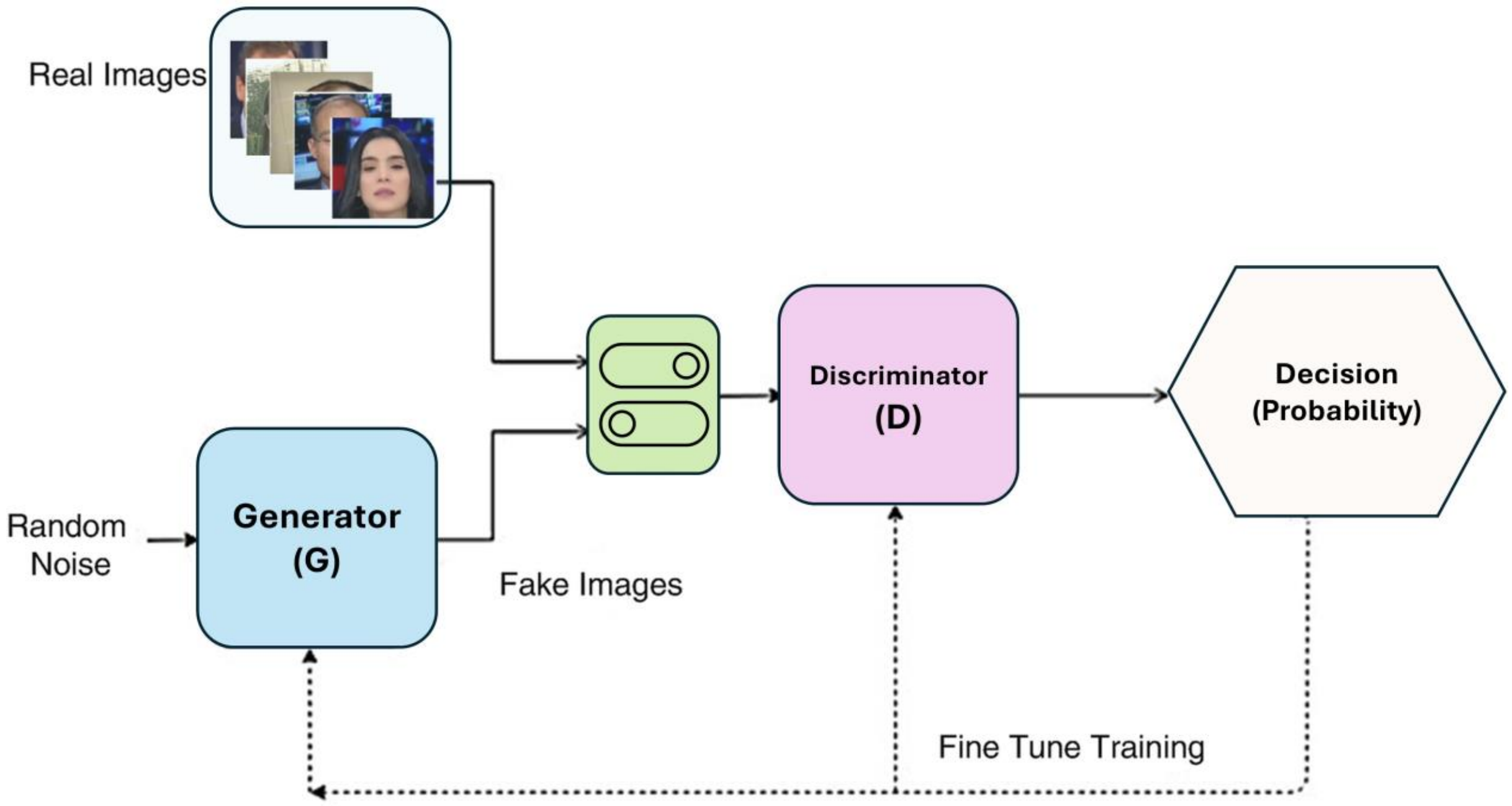

**Fig. 13**: General architecture of Generative Adversarial Networks (GANs) in Deepfake generation process.

### 3.1.1 Deep Convolutional GAN

The Deep Convolutional GANs (DCGANs) model was introduced by Radford et al. (Radford, 2015) as a promising approach for unsupervised representation learning in computer vision, providing stable training and useful image representations, especially in deepfake generation applications. GANs offer an alternative to traditional maximum likelihood techniques for representation learning. Furthermore, DCGANs utilize discriminators for image classification with competitive performance and visualize filters learned by GANs, demonstrating their ability to draw specific objects. DCGANs show that generators have properties enabling the manipulation of generated samples' semantic qualities. While offering significant benefits, training complexity and interpretability challenges remain important considerations in utilizing DCGANs for practical applications.

### 3.1.2 Progressive Growing GAN

The innovative framework known as Progressive Growing GAN (PGGANs) (Karras, 2017) was introduced as a remarkably swift and highly efficient training strategy specifically designed for the complex realm of GANs, which has revolutionized the way for deepfake generation. The pivotal contributions of this groundbreaking methodology can be succinctly outlined in the following key points. To begin with, the foundational concept revolves around the ingenious idea of incrementally expanding both the generator and the discriminator components, initiating the process from lower-resolution images and progressively incorporating new layers as the training advances, thereby enabling the model to capture increasingly intricate and refined details with each iteration. Figure 14 (a) shows the main architecture of the PGGAN model. The major benefits of the PGGANs model are as follows:

- This sophisticated approach not only facilitates the creation of exceptionally high-quality images but also extends to renowned datasets like CELEBA, allowing for the generation of images at astonishing resolutions of up to 1024x1024 pixels, showcasing a level of image quality that is truly impressive and a testament to its effectiveness.
- Moreover, the PGGANs model presented a novel technique to enhance the diversity of the generated images, resulting in an unprecedented inception score in the challenging context of unsupervised CIFAR10, showcasing its remarkable capability to produce varied outputs.
- In addition to this, the progressive growing method significantly contributes to the stabilization of the training process, making it remarkably more consistent and trustworthy, even when dealing with the complexities associated with higher resolutions in image generation.
- Finally, not only accelerated training times but also enhanced stability when operating at elevated resolutions, coupled with a marked improvement in the images' variance, thus successfully addressing the long-standing trade-off between image quality and the diversity of outputs generated.

### 3.1.3 Boundary Equilibrium Generative Adversarial Network

One of the most renowned and widely recognized models in the fascinating realm of deepfake generation is undoubtedly the Boundary Equilibrium GANs(BEGAN) Berthelot (2017). This innovative framework introduced a groundbreaking approach to the training of auto-encoder GANs by meticulously enforcing a state of equilibrium between the two pivotal components: the generator and the discriminator. The central concept behind this remarkable technique is utilising a loss function intricately derived from the Wasserstein distance. It meticulously balances the training processes of these two fundamental elements, ensuring that they develop in harmony with one another. This unique methodology not only results in a significantly faster and more stable training process but also culminates in producing images that exhibit exceptionally high visual quality while simultaneously providing a novel mechanism to expertly navigate and control the delicate trade-off between image diversity and visual fidelity. Furthermore, BEGAN introduces an approximate measure of convergence that stands apart from traditional methods, including those utilized in Wasserstein GANs, thereby offering a fresh perspective on achieving model stability. Figure 14 (b) shows the main architecture of BEGAN model. The primary advantages that distinguish the performance of BEGAN from its counterparts can be summarized as follows:

- A remarkably swift and stable training process that minimizes the typical complications associated with GAN training.
- The generation of images that possess an outstanding level of visual quality, making them indistinguishable from real-life counterparts.
- A streamlined and simplified neural network architecture contrasts sharply with the more complex structures often in conventional GAN methodologies.
- An enhanced degree of control over the intricate balance between image diversity and visual quality allows tailored outcomes that can be finely tuned based on specific requirements.

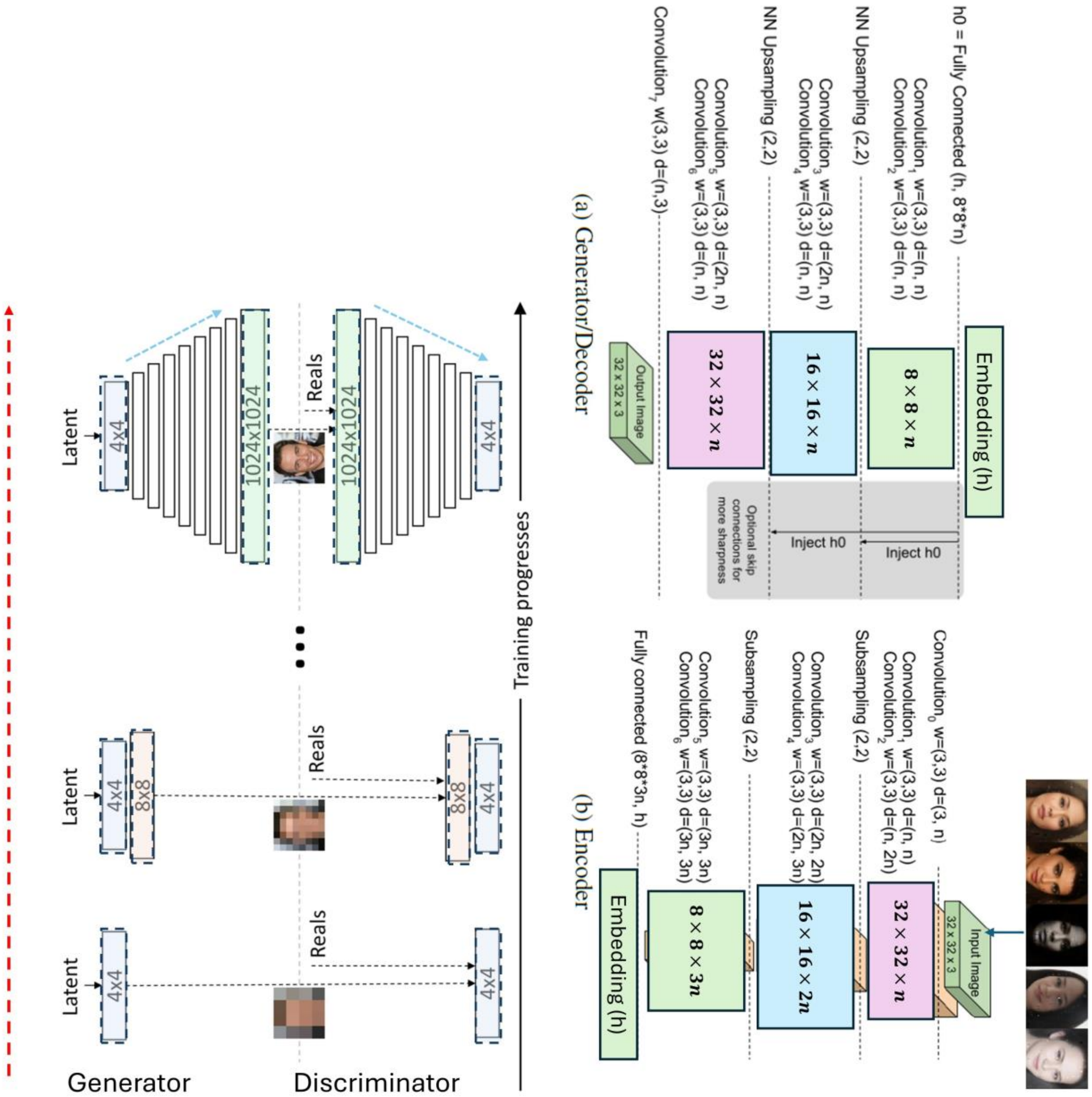

(a) (b)

**Fig. 14**: a) PGGANs architecture is incrementally expanding both the generator and the discriminator components, initiating the process from lower-resolution images and progressively incorporating new layers as the training advances (Adapted from Karras (2017)). b)BEGANs architecture for the generator and discriminator (Adapted from (Berthelot, 2017)) (originally published on arXiv under a Non-exclusive license to distribute).

### 3.1.4 Wasserstein Generative Adversarial Network

Wasserstein GANs (WGANs) (Gulrajani et al., 2017) concept introduced by Ishaan Gulrajani et al., is a significant advancement in the field of GANs. It focuses on improving the stability and performance of GANs, particularly through the WGAN framework. The authors identified issues with weight clipping in WGAN, which can lead to undesirable behaviour during training. They introduced a unique approach called WGAN-gradient penalty(GP) to overcome this. This method minimizes the need for hyperparameter tuning and enables stable training across various GAN architectures, including deep neural networks like 101-layer ResNets and language models. WGAN-GP has successfully achieved high-quality image generation on datasets such as CIFAR-10 and LSUN bedrooms. The potential of WGAN is immense, with the following major advantages.

- Enables stable training of GANs with diverse architectures. Improves the quality of generated samples compared to standard WGAN.
- Requires minimal hyperparameter tuning and can be applied to various GAN setups.
- Achieves high-quality image generation results on standard datasets.

Despite the numerous strengths of WGAN, implementing the gradient penalty method may introduce complexities to the GAN training process. This new approach may require additional computational resources compared to weight clipping. Furthermore, integrating the proposed method into existing GAN architectures may necessitate adjustments and careful tuning, posing potential challenges to its adoption.

### 3.1.5 Cycle GAN

Zhu and colleagues (Zhu et al., 2017) unveiled a groundbreaking technique for unpaired image-to-image translation utilizing Cycle-Consistent Adversarial Networks (CycleGANs). The core aim is to grasp the relationships between two distinct image realms without the need for paired instances by incorporating cycle consistency constraints. CycleGAN facilitates the conversion between image realms without the necessity of paired training datasets. Moreover, by embedding cycle consistency within the translations, the model guarantees that transforming an image from one domain to another and reverting it will yield the original image. Additionally, the framework utilizes adversarial losses to ensure that the translated visuals blend seamlessly with the images of the target domain. The key benefits of CycleGANs include: • Eliminating the restriction of requiring paired training data, thus broadening its applicability across diverse tasks.

- Suitable for a multitude of image translation applications, such as style transformation, object metamorphosis, attribute modification, and photo refinement.
- CycleGANs mark a remarkable advancement in the arena of image-to-image translation, surpassing earlier techniques that depended on manually defined separations of style and content or shared embedding functions. This exceptional capability is a clear indication of the model's progress and inventiveness, prompting further inquiry and growth in this field.

Even though all CycleGANs exhibit high efficiency, they encounter a few shortcomings, including conventional methods that may lead to mode collapse, where all input images converge to a singular output image, obstructing optimization. Additionally, training both mapping functions can demand considerable computational resources. Furthermore, the approach may lack clarity regarding how individual input and output pairs align in an unpaired context.

## 3.2 Autoencoders

Autoencoders function by encoding input data into a lower-dimensional representation and then decoding it to reconstruct the original input. In the context of deepfakes, autoencoders are trained on a dataset of the target person's images, capturing their essential facial features and variations (Katarya and Lal, 2020). By manipulating the encoded representation within certain boundaries, autoencoders allow for the generation of deepfakes with subtle alterations while retaining the overall likeness to the target person. This approach provides control over the level of modification in the generated deepfakes.

## 3.3 Variational Encoders

Variational encoders offer another approach to deepfake creation, focusing on the probabilistic modelling of facial features. By learning the underlying distribution of facial characteristics, variational encoders enable the generation of diverse deepfakes by sampling from the learned distribution. This technique ensures that the generated deepfakes closely resemble the target person's features, adding to their realism. Yadav and Salmani (Yadav and Salmani, 2019) conducted a survey on facial forgery techniques using GANs, shedding light on the application of variational encoders in deepfake generation.

## 3.4 Deep Neural Networks

Deep neural networks, with their complex architecture and depth, play a fundamental role in deepfake creation. These networks, such as Convolutional Neural Networks (CNNs), Recurrent Neural Networks (RNNs), and transformer networks, have the capacity to capture intricate patterns and fine-grained details in human faces. By training deep neural networks on large datasets, deepfakes can achieve remarkable levels of visual realism, capturing not only facial appearance but also subtle expressions and movements.

Furthermore, Brooks and Yuan provide a comprehensive examination of synthetic image animation, specifically concentrating on the captivating realm of full-body animation [38]. This paper explores various facets of the field, including critical topics such as pose transfer, motion transfer, and motion generation. As per the paper, the animation methods using Deep Neural Networks (DNNs), existing studies can be categorized into roughly three areas: Human Pose Transfer (MA et al.), Human Motion Transfer, Human Motion Generation (Tulyakov et al., 2018). The authors intended to develop a deeper understanding of the various animation methods that employ DNNs, offering insights into the advancements and challenges within the field of deepfake animation. Table 2 and 3 present a comprehensive overview of these different tools, summarizing their unique features and drawbacks.

## 3.5 Formulation of manipulation in Deepfake generation methods

Four widely recognized generations of deepfakes have been carefully chosen based on various perspectives that encompass the different modelling types, the sources of conditions that influence them, the effects of forgery that they produce, and the diverse functions they serve in digital manipulation. To formalize this concept, it is represented as the destination subject that is subject to alteration as $x_t$ ($i_t$,$a_t$,$b_t$), which possesses a myriad of attributes—namely, $i$, which signifies the unique identity of the person, $a$, which refers to the identity-agnostic content that is not tied to any specific individual, and $b$, which denotes the external characteristics that can distinctly identify the subject.

In contrast, the source $x_s$ ($i_s$,$a_s$,$b_s$) is seen as the conditional media ($c$) that acts as a catalyst, prompting the target to undergo transformations in either its identity, its attributes, or sometimes even both simultaneously. Within the literature, the notation $f_{sw}$,$f_{re}$,$f_{sy}$,$f_{ed}$ were applied to symbolize four distinct categories of deepfakes that the technical details can be seen in Algorithms 1,2, 3, 4, are analyzed: *FS*, *FR*, *EFS*, and *FE*, each representing a unique approach to the manipulation of digital content. The visual representations of these categories are illustrated in Figure 15.

### 3.5.1 Face-Swapping Approach

As depicted in Figure 15(a), and Algorithm 1, the Face-Swapping (FS) method endeavours to seamlessly substitute the content of the target $x_t$ with the content sourced from $x_s$, all while meticulously ensuring that the identity $i_s$ remains intact and unaffected. Formally, this transformation can be expressed as **Table 2**: Summary of deepfake generation tools (section 1)

| No. | Tool | Year | Brief Description | Advantage | Disadvantage | Manipulation type |
|---|---|---|---|---|---|---|
| 1. | Disco Face-GAN Deng et al. (2020) | 2020 | Disentangled and Controllable Face Image Generation via 3D Imitative-Contrastive Learning | User-friendly and flexible training | Limited features and community support | ImitativeContrastive Learning |
| 2. | DeepFaceLab Perov et al. (2020) | 2019 | The software is wellknown for its proficiency in manipulating lips, replacing faces, and altering heads. | High Per-formance, Feature-Rich | Steep Learning Curve, Resource Intensive. | FaceSwap |
| 3. | Zao Anto-niou (2019) | 2019 | Users can utilize a single image to overlay their faces onto movie footage using this tool. | Quick Results | Privacy Concerns, Limited Control | FaceSwap |
| 4. | FSGAN Nirkin et al. (2019) | 2019 | (RNN)-based approach for face reenactment, which adjusts for both pose and expression variations and can be applied to a single image or a video sequence. | Identity Preservation | Complex Setup | FaceSwap |
| 5. | FakeApp Nirkin et al. (2019) | 2018 | FakeApp software was initially developed by Reddit users. | User-Friendly | Outdated | FaceSwap |
| 6. | StarGAN Choi et al. (2018) | 2018 | This tool employs unified GANs to facilitate image-toimage translation across multiple domains. | MultiDomain Support | Complexity and parameter tuning | Face AIGC |
| 7. | DFaker Alheeti et al. (2021) | 2017 | Inputs are 64x64 images outputs are a pair of 128x128 images one RGB with the recon-structed face | Efficient Processing | Limited Features | DSSIM |
| 8. | FaceApp Neyaz et al. (2020) | 2016 | The smartphone application employs multiple AI filters and effects to generate deepfake images | User-Friendly | Privacy Concerns | Facial Attribute manipulation |

| 9. | Faceswap Neyaz et al. (2020) | 2015 | It is compatible with all operating systems and necessitates a powerful GPU for optimal performance. | Extensive Community | Steep Learning Curve | FaceSwap |
|---|---|---|---|---|---|---|

**Table 3**: Summary of deepfake generation tools, section 2

| No | Tool | Year | Brief Description | Advantage | Disadvantage | Manipulation type |
|---|---|---|---|---|---|---|
| 1 | Synthesia Kaate et al. (2023) | 2022 | create high-quality video presentations | Realistic Facial Manipulation, Potential for Creative Applications | Ethical Concerns, Misinformation and Manipulation, Legal Implications | digital facial manipulation |
| 2 | Heygen HeyGen (2024) | 2022 | Good in text-to-video deepfakes | Efficiency, Scalability, Customization | potential to spread misinformation, Detection is challenging | Facereenactment |
| 3 | BlendFace Shiohara et al. (2023) | 2023 | latest face-swapping method | Great in single-frame face swapping | Unstable and inconsistent face swap in video frames | Face-swapping (FS) |
| 4 | e4s Liu et al. (2023) | 2023 | considering both geometry and tex-ture details | Fine-grained Control, Preservation of Iden-tity | Temporal Inconsistency, Complexity | Face-swapping (FS) |
| 5 | FaceDancer Rosberg et al. (2023) | 2023 | single-stage approach for face swapping of unknown identities, | High-Quality Face Swaps, Identity Preservation, High-Quality Face Swaps | Complex Imple-mentation, Resource Intensive | Face-swapping (FS) |
| 6 | TPSMM Zhao and Zhang (2022) | 2022 | introduces a thinplate spline motion model | Thin-Plate Spline Motion Model, Unsupervised Method | Complex Imple-mentation, Potential for Over-fitting | Facereenactment (FR) |
| 7 | SadTalker Zhang et al. (2023) | 2023 | decomposes into:audio-motion mapping and 3D motion modeling | High-Quality Video Output, Realistic Facial Expressions | Training Data Requirements, Ethical Concerns | Facereenactment (FR) |
| 8 | Hyper Reenact Bounareli et al. (2023) | 2023 | another recent reenactment method | One-shot Reenactment, Artifact Minimization, Cross-Subject Reenactment | Potential for Misuse, Legal Implications | Facereenactment (FR) |
| 9 | StyleGANXL Sauer et a l. (2022) | 2022 | another version of the original Style-GAN | Scalability to large and diverse datasets, Stateof-the-art synthesis | Potential misuse, Resource-intensive | Entire Face Synthesis (EFS) |
| 10 | *PixArt–α* Chen et al. (2023) | 2024 | a Transformer-based text-to-image synthesis model | Competitive Image Quality, Reduced Training Cost | Limited Extensive Testing, Overfitting Risks | Entire Face Synthesis (EFS) |
| 11 | SiT-XL/2 Atito et al. (2021) | 2021 | Scalable Interpolant Transformers | Scalable Interpolant Transformers, Modular Design Exploration | Potential Overfitting, Ethical Considerations | Entire Face Synthesis (EFS) |
| 12 | CollabDiff Dolhansky et al. (2020) | 2023 | multi-modal face generation and editing | Multi-Modal Control, High-Quality Image Synthesis | Dependency on Pre-Trained Models, Security Concerns | Face Editing (FE) |
| 13 | UniFace Xu et al. (2022a) | 2022 | another recent faceswapping technique | Disentangled Representations, Feature Disentanglement | Visible Rectangular Frame, Blending and Transition Challenges | Face-swapping (FS) |
| 14 | RDDM Liu et al. (2024) | 2024 | residual Denoising Diffusion Model | Robust Image Generation, Directed Degradation Modeling | Model Inter-pretability, Resource Intensive | Entire Face Synthesis (EFS) |
| 15 | DiT-XL/2 Peebles and Xie (2023) | 2023 | termed Diffusion Transformers | Improved Scalability, Generative Modeling Capabilities | Complexity and Computational Cost, Ethical Concerns | Entire Face Synthesis (EFS) |

$\tilde{x}_t\left(\tilde{i}_s, a_t, b_t\right)$, which indicates that the only change made during this process is the replacement of the identity $i$ from the source $x_s$ to the target $x_t$, while the identity-agnostic content $a$ remains preserved in its original form. The intricate procedure of this swapping mechanism can be articulated as follows:

$$f_{sw}\left(x_t\left(i_t, a_t, b_t\right), x_s\left(i_s, a_s, b_s\right) \mid \text{swap}\right) = \tilde{x}_t\left(\tilde{i}_s, a_t, b_t\right) \quad (2)$$

**Algorithm 1** *Face − Swapping (FS)*

1: **procedure** FS ( Original image)
2: **Initialisation**
3: $x_t\left(i_t, a_t, b_t\right)$ ▷ Destination image
4: $x_s\left(i_s, a_s, b_s\right)$ ▷ Original image
5: Formulate $x_t$ =destination image with properties $i_t, a_t, b_t$
6: Formulate $x_s$ =original image with properties $i_s, a_s, b_s$
7: Execute Face-Swapping technique to exchange identity $i$ from the origin $x_s$ to the destination $x_t$, conserving the identity-neutral content $a$.
8: $\tilde{x}_t\left(\tilde{i}_s, a_t, b_t\right) \leftarrow f_{sw}\left(x_t\left(i_t, a_t, b_t\right), x_s\left(i_s, a_s, b_s\right) \mid \text{swap}\right)$
9: **return** $\tilde{x}_t$ ▷ Manipulated image
10: **end procedure**

FS techniques shine in crafting stunningly realistic modifications in both videos and images. This innovative technology goes beyond simple visual illusions for entertainment (Zhu et al., 2021), allowing for the production of deepfake videos that embody humour or artistic creativity. In the film industry, FS emerges as a crucial tool for creating intricate special effects and effortlessly altering actors' looks without relying on heavy makeup or prosthetics. Moreover, these techniques open up a vibrant landscape for exploration and innovation in computer vision, ML, and graphics. However, alongside these benefits, the potential for misuse of FS technology raises significant alarms, as it can be exploited for malicious purposes like generating fake news, spreading false information, or committing identity fraud (Li et al., 2024). This technology brings forth considerable privacy concerns, as it facilitates altering or replacing people's faces without informed consent. The ethical dilemmas associated with FS are substantial, particularly when modifying an individual's likeness without their permission. The legal framework governing the creation and distribution of deepfake content is still evolving, with potential consequences tied to copyright violations and defamation. The rampant spread of deepfake material threatens the credibility of visual media, breeding doubt about the legitimacy of photographs and videos.

### 3.5.2 Face Reenactment Approach

As illustrated in Figure 15(b), and Algorithm 2, the Face Reenactment(FR) method applied to the variable $x_t$ $(i_t, a_t, b_t)$ adeptly maintains the essence of its identity, denoted as $i_t$, while simultaneously allowing for the intricate intrinsic attributes represented by $a_t$, such as the pose, mouth shape, and overall expression of the face, to be deftly manipulated through the influence of a dynamic variable labelled $c_a$. This manipulation results in a transformed representation, which it denotes as $\tilde{x}_t\,(i_t, \tilde{a}_s, b_t)$, suggesting that the identity remains intact even as its features undergo significant modifications. In mathematical terms, we can elegantly derive the subsequent Equation 3 that encapsulates this transformative process.

$$f_{re}\,(x_t\,(i_t, a_t, b_t) \mid c_a) = \tilde{x}_t\,(i_t, \tilde{a}_s, b_t) \tag{3}$$

**Algorithm 2** *Face − Reenactment (FR)*

**procedure** FR ( ORIGINAL IMAGE)
**Initialisation**
  $x_t\,(i_t, a_t, b_t)$ ▷ Destination image
  $c_a$ ▷ Conditional source
  Specify $x_t$=destination image with characteristics $i_t, a_t, b_t$.
  Define $c_a$ as the dependent origin to modify the inherent attributes $a_t$.
  Employ the Face-Reenactment technique to alter the inherent attributes $a_t$ of the destination image $x_t$ while safeguarding its identity $i_t$.
  $\tilde{x}_t\,(i_t, \tilde{a}_s, b_t) \leftarrow f_{re}\,(x_t\,(i_t, a_t, b_t) \mid c_a)$
  **return** $\tilde{x}_t$ ▷ Manipulated image
**end procedure**

FR methodologies thrive in amplifying the realism of deepfake videos by meticulously capturing and recreating facial expressions and movements (Thies et al., 2015). This innovative technology also enhances lip synchronization in dubbed films or animations, perfectly aligning facial motions with corresponding audio signals. FR is also an essential asset for translating foreign language content or localizing videos, enabling the seamless synchronization of lip movements with adapted dialogues, thus providing a flexible solution for global content transformation (Tripathy et al., 2021). FR techniques are pivotal in producing lifelike animations, groundbreaking special effects, and captivating character portrayals in cinema. Nevertheless, the potential for misusing FR technology raises alarms about the spread of misinformation, video alterations, and the fabrication of events, which could erode trust and credibility (Thies et al., 2018). Privacy concerns emerge as FR can mimic individuals' facial subtleties without explicit approval, invoking ethical dilemmas regarding unauthorized representation. Moreover, manoeuvring through the intricate legal terrain surrounding FR-generated material requires addressing challenges related to copyright violations and intellectual property issues (Yao et al., 2020). The extensive integration of FR-generated media blurs the lines between reality and fiction, posing significant questions about the authenticity and dependability of visual content.

### 3.5.3 Entire Face Synthesis

Entire Face Synthesis (EFS) technique is capable of generating a wholly synthesized facial image, represented as $\tilde{x}_t\left(\tilde{i}_t, \tilde{a}_t, \tilde{b}_t\right)$, which showcases an entirely new visual representation. To effectively address and narrow down the disparity in personal identity that often exists between generative models and other prevalent Deepfake techniques, it is meticulously fine-tuning the generative model $G$ using the same authentic dataset that other forgery methods, such as e4s [50], utilize. This fine-tuning process allows the technique to create a robust face-synthesis model, denoted as $f_{sy}$, specifically designed to generate highly realistic facial images. Subsequently, it can synthesize new facial representations from random noise $n$, leading to a diverse array of facial outputs, as illustrated in the general generation process depicted in Figure 15 (c) and Algorithm 3.

$$f_{sy}(n) = \tilde{x}_t\left(\tilde{i}_t, \tilde{a}_t, \tilde{b}_t\right) \tag{4}$$

**Algorithm 3** *Entire Face Synthesis (EFS)*

1: **procedure** EFS ( ORIGINAL IMAGE)
2: **Initialisation**
3: $n, D, G,$ ▷ Noise, Real Data, Generative Model
4: Introduces $n$ as the disturbance employed in creating a wholly artificial face.
5: Adjust the generative model $G$ with the authentic data $D$ to acquire a facial synthesis model $f_{sy}$.
6: Implement the EFS technique to produce a fully synthetic face from the disturbance $n$ utilizing the refined model $f_{sy}$.
7: $\tilde{x}_t\left(\tilde{i}_t, \tilde{a}_t, \tilde{b}_t\right) \leftarrow f_{sy}(n)$
8: **return** $\tilde{x}_t$ ▷ Manipulated image
9: **end procedure**

EFS approach in deepfake generation offers a multitude of benefits and drawbacks. EFS facilitates the holistic reconstruction of a visage, enabling elaborate and meticulous alterations to facial features, emotions, and gestures. EFS techniques can produce astonishingly realistic deepfake visuals by fabricating complete faces, leading to captivating aesthetic results (Zhang et al., 2024). This method acts as a channel for creativity, allowing individuals to explore artistic endeavours, filmmaking, and entertainment through cutting-edge face fabrication. EFS emerges as an invaluable tool in the film and media industries for creating spectacular effects, altering actors' looks, and devising visually striking scenes (Yu et al., 2023). Nevertheless, the potential abuse of EFS technology for deceptive activities, such as generating false content, disseminating misinformation, or committing identity theft, poses a substantial threat. EFS also brings forth significant privacy concerns as it enables the alteration of entire faces, which could infringe upon personal privacy rights. Ethical dilemmas may surface when employing EFS, significantly when an individual's facial identity is changed without permission or for harmful purposes. The legal framework surrounding EFS-produced content is complex, involving matters related to intellectual property laws, privacy regulations, and the authenticity of digitally altered visuals (Hsu et al., 2024). In the end, the rampant spread of EFS-generated deepfakes could erode trust in visual media, cultivating scepticism about the genuineness and integrity of images and videos.

### 3.5.4 Face Editing approach

Face Editing (FE) approach: in reference to Figure 15 (d) and Algortihm 4, the FE method applied to the variable $x_t$ ($i_t$,$a_t$,$b_t$) facilitates alterations to its external attributes, represented as $b_t$, which include various characteristics such as facial hair, age, gender, and ethnicity, all of which are controlled by a conditional source variable $c_b$. This process enables us to achieve a modified version of the facial representation, denoted as $\tilde{x}_t\left(i_t, a_t, \tilde{b}_s\right)$, signifying that external features can be flexibly altered. Furthermore, the capability for multiple attribute manipulations through two distinct editing approaches is incorporated, allowing for simultaneous adjustments, such as cohesively altering hair and eyebrows. This transformational capability can be formalized using the following equation: the notation ∼ indicates that the particular element has been manipulated through sophisticated forgery algorithms, enhancing the overall versatility and realism of the synthesized faces.

$$f_{ed}\left(x_t\left(i_t, a_t, b_t\right) \mid c_b\right) = \tilde{x}_t\left(i_t, a_t, \tilde{b}_s\right) \tag{5}$$

**Algorithm 4** *Face Editing (FE)*

1: **procedure** FE ( ORIGINAL IMAGE)
2: **Initialisation**
3: $x_t\left(i_t, a_t, b_t\right)$, ▷ Destination image
4: $c_b$ ▷ Conditional source
5: Specify $x_t$=destination image with characteristics $i_t, a_t, b_t$.
6: Define $c_b$ as the dependent origin to adjust the exterior attributes $b_t$.
7: Utilize the Face Editing (FE) technique to modify the exterior attributes $b_t$ of the destination image $x_t$, while retaining its identity $i_t$ and inherent attributes $a_t$.
8: $\tilde{x}_t\left(i_t, a_t, b_s\right) \leftarrow f_{ed}\left(x_t\left(i_t, a_t, b_t\right) \mid c_b\right)$
9: **return** $\tilde{x}_t$ ▷ Manipulated image
10: **end procedure**

The FE method in deepfake generation presents a range of advantages and disadvantages. By utilizing FE techniques, precise facial features, expressions, and attribute adjustments can be achieved with remarkable accuracy (Nickabadi et al., 2022). This approach not only enhances visual effects in videos, films, and images by skillfully editing and transforming facial appearances for desired outcomes but also serves as a conduit for creative expression (Rana et al., 2022). Innovative face editing methods allow Users to explore artistic pursuits, storytelling, and digital content creation. Moreover, FE facilitates facial reenactment, enabling the realistic replication of facial movements and expressions, which is a valuable asset for the entertainment, special effects, and animation industries. In essence, while Face Editing techniques provide sophisticated tools for facial modification and creative expression, the risks associated with misuse, privacy breaches, ethical dilemmas, legal intricacies, and implications for trust and authenticity underscore the necessity for conscientious development, regulation, and ethical considerations in their utilization (Pernuˇs et al., 2023). In Figure 16, a comparison of eight effective face-swapping generated datasets within the FF domain reveals that e4s and FSGAN fall short in producing high-quality synthetic faces that accurately mimic real facial features. These methods, e4s and FSGAN, exhibit limitations in generating synthetic faces that convincingly replicate the natural effects and details observed on real faces.

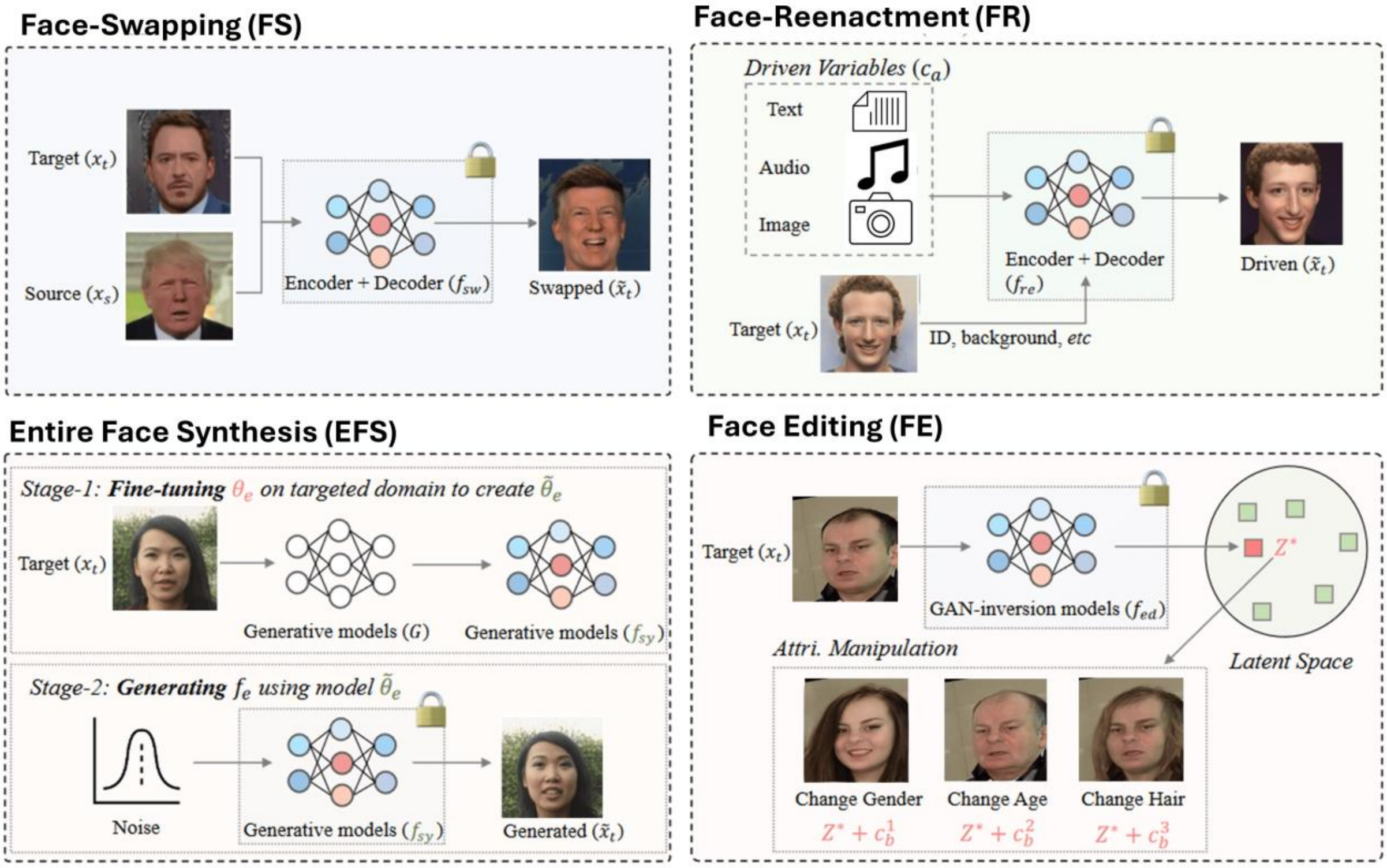

**Fig. 15**: Overview of four common Deepfake data generation pipelines, featuring Face-Swapping (FS), Face-Reenactment (FR), Entire Face Synthesis (EFS), and Face Editing (FE) techniques (Adapted from Yan et al. (2024).(Originally published on arXiv under a CC0 license to distribute)

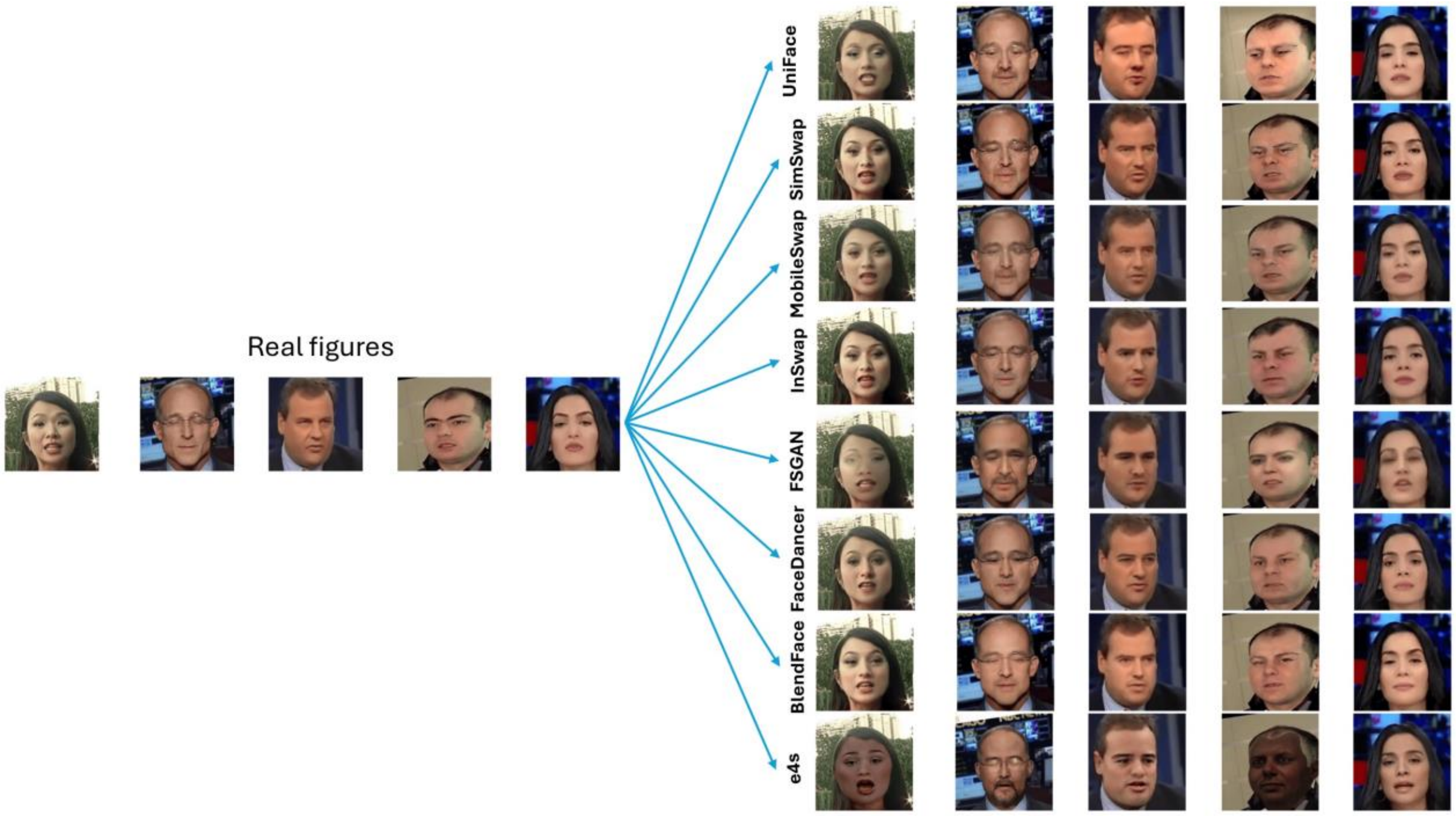

**Fig. 16**: Visual comparisons for eight popular face-swapping generated data performance within the FF domain Adapted from Yan et al. (2024). Originally published on arXiv under a CC0 license to distribute

Figure 17 illustrates a notable trend in deepfake detection techniques, showcasing the significant growth in the applications and theoretical developments of forgery detection methods compared to other approaches. The prominence of forgery detection methods in deepfake detection can be ascribed to several key factors. Firstly, these methods boast a solid foundation built upon years of research in digital forensics and media authentication (Kaur et al., 2023). Additionally, their interdisciplinary nature allows for the utilization of a diverse set of tools and methodologies from fields such as computer vision, ML, and cryptography. Furthermore, the adaptability and transferability of knowledge from forgery detection to deepfake detection provide a robust framework for identifying synthetic media. The practical relevance of forgery detection methods (Mehrjardi et al., 2023) in domains like law enforcement, cybersecurity, and content verification further underscores their importance in addressing the escalating threat posed by deepfake technology in today's digital landscape.

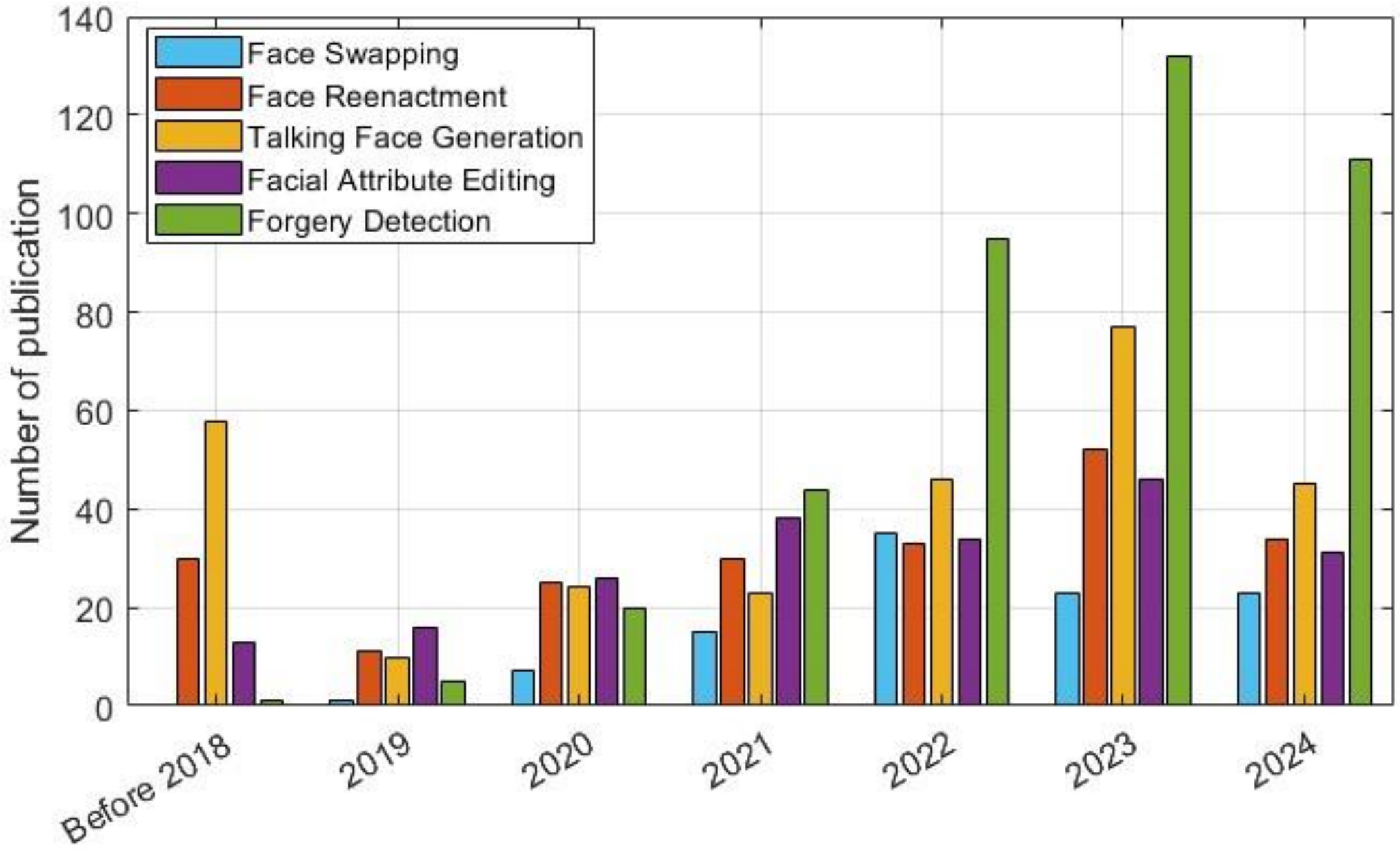

**Fig. 17**: Number of publications of five common Deepfake data generation approaches, including Face Swapping, Face-Reenactment, Talking face generation, Facial attribute editing, and Forgery detection techniques based on Scopus (Accessed at 09/2024)

## 3.6 Challenges and solutions in Deepfake generations

In the intricate and rapidly evolving realm of deepfake generation tools, the pursuit of responsible and ethical applications faces a multitude of significant and daunting challenges that require careful consideration and innovative solutions (Yan et al., 2024). The accuracy and realism of deepfake algorithms encounter substantial obstacles that prevent them from producing flawless, realistic images free from any unsightly artefacts or distortions that may detract from their authenticity. In addition to the paramount concern of visual fidelity, ensuring temporal coherence throughout lengthy video sequences presents an immensely challenging endeavour, as it necessitates the implementation of highly sophisticated mechanisms designed to facilitate seamless transitions that maintain viewer engagement (Ju et al., 2023). The preservation of the integrity of the identity and facial subtleties of the individual being targeted emerges as a critical issue, one that is essential for establishing and maintaining the credibility and trustworthiness of the content that has been generated (Liu et al., 2023). The overwhelming need for broad generalization across a wide array of identities, poses, varying lighting conditions, and emotional expressions remains a significant hurdle for deepfake models, which demands the development of robust and effective strategies aimed at enhancing their adaptability in diverse scenarios. Furthermore, the inherently resource-intensive characteristics of deepfake generation models highlight the urgent necessity for optimizing computational efficiency and memory utilization (Narayan et al., 2023), which are vital for streamlining the production process in an effective and sustainable manner. These myriad challenges serve to underscore the multifaceted and complex nature of advancing deepfake technology, all while striving to uphold and adhere to ethical standards in its deployment and application within society.

In the elaborate and multifaceted realm of tools designed for the generation of deepfakes, a thorough and insightful examination uncovers a set of five key technical hurdles intricately intertwined with various groundbreaking solutions that are both creative and effective. Among these, the enhancement of the quality of training data stands out as a fundamental aspect, alongside the adoption of cutting-edge image processing techniques, such as super-resolution, which significantly elevate the visual quality of the output while also incorporating adversarial training methodologies that play a crucial role in minimizing unwanted artefacts that can detract from the realism of the generated images (Sauer et al., 2022). As we delve even deeper into the complexities of this field, it becomes abundantly clear that the integration of temporal coherence mechanisms, which include sophisticated processes like optical flow estimation and frame interpolation, in addition to custom-designed loss functions, is vital for achieving not only smooth transitions between frames but also for significantly improving the overall quality of the video content produced (Song et al., 2023). Furthermore, the delicate task of maintaining the unique identity and emotional expressions of the individual being represented requires the implementation of advanced techniques, such as identity preservation loss functions, landmark-guided generation methods, and intricate attention mechanisms that ensure the fidelity of the portrayal. In addition to these considerations, the challenge of diversifying the training data necessitates the exploration of various data augmentation strategies and a deep dive into domain adaptation techniques, which are essential for enhancing the model's ability to generalize effectively while simultaneously increasing the diversity of the generated content (Sun et al., 2023).

Ultimately, optimizing model architectures for improved efficiency emerges as a critical focus, with strategies like knowledge distillation, quantization, and model pruning being indispensable for achieving a streamlined computational footprint without sacrificing the high-performance standards expected in this fast-evolving domain (Wang et al., 2022). These intricate technical challenges not only highlight the rapidly changing landscape of deepfake technology but also emphasize how innovative solutions are driving significant advancements in the field, all while ensuring that ethical considerations and responsible deployment practices remain at the forefront of this transformative journey. As we continue to navigate this complex environment, we must remain attentive to the implications of these technologies, fostering a culture of responsibility and vigilance that prioritizes the ethical use of our creative capabilities. The confluence of technical prowess and ethical accountability will ultimately shape the future of deepfake technology, ensuring that it serves as a tool for creativity and innovation rather than deception and harm.

# 4 Deepfake detection techniques, challenges and solutions

Detecting deepfakes poses a significant challenge as these sophisticated manipulations have become highly convincing (Yan et al., 2024). However, with the rapid advancement of technologies, substantial research has been dedicated to developing effective deepfake detection techniques. Baisc's steps for deepfake detection are given in Figure 18, along with detection techniques in Figure 19.

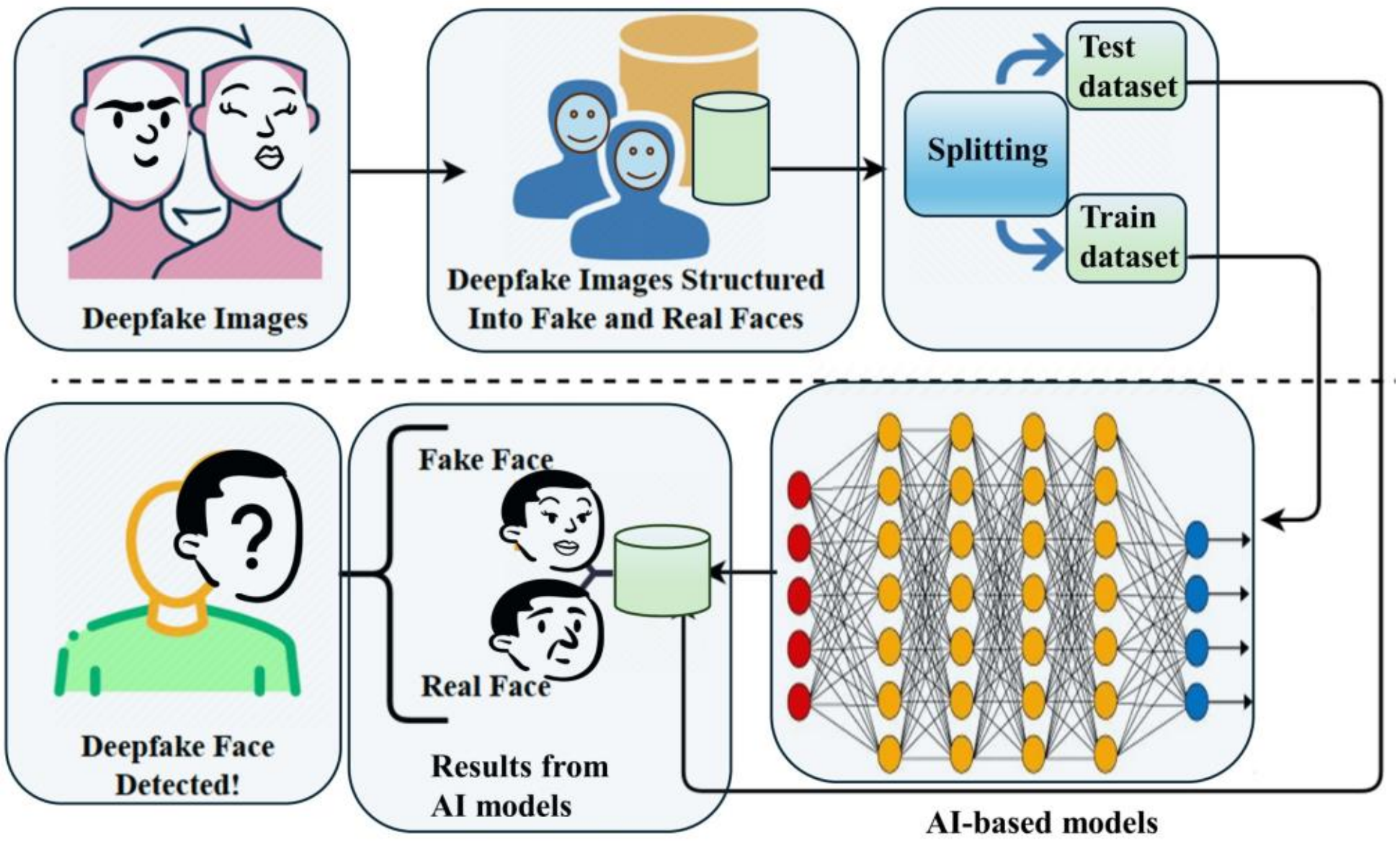

**Fig. 18**: Steps of deepfake detection (redesigned from (Raza et al., 2022)). Originally published on open access under a CC BY 4.0 license to distribute

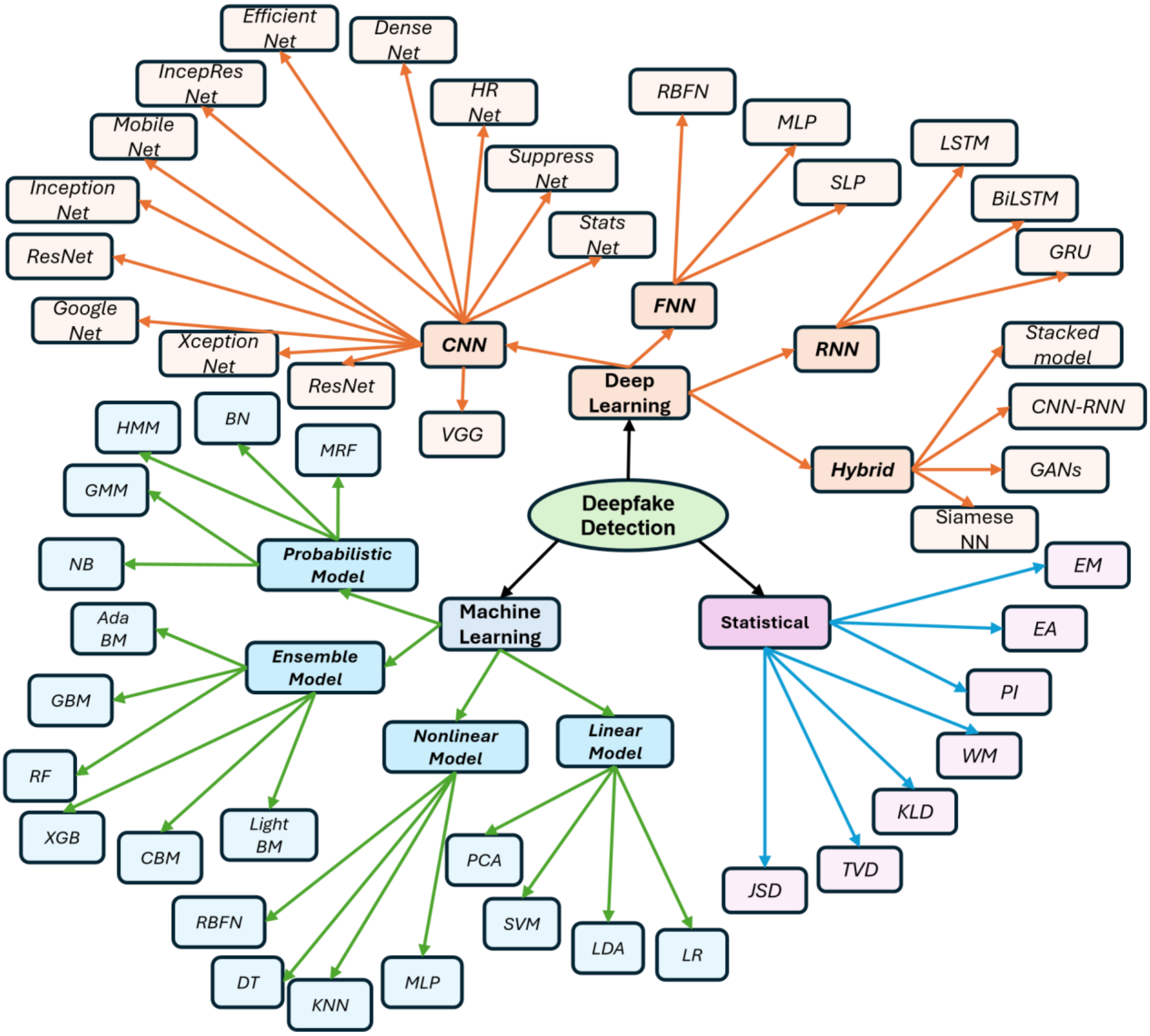

**Fig. 19**: The major classification of Deepfake detection techniques

In response to this pressing need, Facebook organized the Deepfake Detection Challenge (DFDC) (Juefei-Xu et al., 2022), a landmark competition aimed at tackling the deepfake problem head-on. The primary objective was to foster innovation and collaboration in developing state-of-the-art deepfake detection algorithms (Yu et al., 2021). Facebook curated a diverse and comprehensive dataset comprising original and manipulated video clips to facilitate this endeavour. This dataset (Montserrat et al., 2020), consisting of approximately 100,000 video clips, provided a rich and varied training set for participants to develop and refine their detection algorithms. The results of the DFDC were both enlightening and sobering, and the winning algorithms demonstrated significant progress in detecting deepfakes; they also revealed the complexity and evolving nature of the problem. Surprisingly, 1/3rd portion of video clips classified as deepfakes by the algorithms were, in fact, authentic recordings (Dolhansky et al., 2020). Since the DFDC, the research community has made significant strides in enhancing deepfake detection techniques.

Advancements in ML, computer vision, and forensic analysis have fueled the development of more sophisticated algorithms that identify subtle traces of manipulation. Researchers have explored various approaches, including multimodal analysis, neural network architectures, and forensic watermarking, to improve the accuracy and reliability of deepfake detection.

Since the emergence of deepfakes in 2017, there has been a significant surge in research and development efforts focused on deepfake application of DL methods. These techniques leverage the power of Artificial Neural Networks (ANN) to analyze and identify anomalies in visual and audio cues, enabling the detection of manipulated content. However, it is worth noting that DL is not the sole solution in the deepfake detection landscape. Other approaches, such as forensic analysis, image and video metadata analysis, and data-driven statistical methods, have also been explored to augment the detection capabilities and provide a more comprehensive defence against the proliferation of deepfakes. The collective efforts of researchers and experts in multiple disciplines are crucial in advancing the state of the art and ensuring a robust and multifaceted approach to deepfake detection. In a study, the authors categorize deepfake detection algorithms into four groups: ML-based, deep learning-based, statistical approaches, and blockchain-based methods (Rana et al., 2022).In the following sections, we will explore each of the categories mentioned above of deepfake detection algorithms in detail.

## 4.1 Deep learning based models

Deep learning-based models have emerged as a dominant force in Deepfake detection. Due to their ability to capture complex patterns and correlations, DL-based models have shown remarkable accuracy in detecting deepfakes. Their widespread adoption and impressive performance make them a prominent choice for researchers and practitioners seeking robust and reliable deepfake detection solutions. By training on large-scale datasets of real and manipulated videos, DL models can capture subtle visual cues and inconsistencies indicative of deepfake generation. The ability of DL algorithms to automatically extract and analyze high-dimensional features makes them well-suited for detecting deepfakes across various modalities, including images, videos, and audio. With continuous advancements in DL techniques and the availability of labelled deepfake datasets, DL-based methods offer promising avenues for improving the detection and mitigation of deepfake threats.

Deep learning-based algorithms for deepfake detection can be categorized into several broad categories, including CNNs, RNNs, Hierarchical Multi-Scale Networks (HMNs), and Multi-Task Cascaded Convolutional Networks (MTCNNs). These categories represent architectural designs and techniques that leverage deep neural networks to identify and distinguish between authentic and manipulated media. The category of DL methods can be seen in Figure 19.

### 4.1.1 Convolutional Neural Networks:

CNNs have undeniably transformed the landscape of DL, showcasing exceptional prowess in the meticulous analysis of spatially organized data, with a particular emphasis on images rich in visual information (Soudy et al., 2024). At the heart of these innovative networks lies a crucial component known as the convolutional layer, where the neurons are designed with fixed local receptive fields and implement a technique called weight sharing, thereby constructing a lattice-like framework that perfectly corresponds to the dimensions of the input data. In this intricate arrangement, these neurons act as local filters or kernels that engage in the process of convolution with the input data, which is represented in a two-dimensional format, to produce activations that serve as a foundation for the network to unearth complex spatial features that are embedded within the images. In a typical convolutional layer, one can find an array of multiple filters, each meticulously crafted to perform a unique feature extraction task, thereby augmenting the network's capability to accurately recognize various patterns and objects present within the input data. The formulation of a 2D convolution layer is as follows.

$$(f * g)(i, j) = \sum_{m=-\frac{s}{2}}^{\frac{s}{2}} \sum_{n=-\frac{s}{2}}^{\frac{s}{2}} f(m, n) g(i + m, j + n) \tag{6}$$

In this process, the mathematical representation where *f* signifies a filter of size $s \times s$ that is applied to every $s \times s$ patch centred at the coordinates (*i,j*) on the input image denoted as *g* becomes highly significant. Generally, an element-wise nonlinear function is subsequently applied to the outcomes generated by the convolution operation, resulting in a structured lattice of neurons, each equipped with identical weighted-sum-and-nonlinearity characteristics, with each neuron focusing its attention on a distinct $s \times s$ segment of the input image in order to derive meaningful insights.

The implementation of convolutions within CNNs bestows upon them a series of distinctive advantages when compared with traditional MultiLayer Perceptrons, commonly known as MLPs as follows: • As the overall number of unique trainable weights is significantly diminished, leading to an inherently more robust model against the pitfalls of overfitting, it becomes a more reliable tool for various applications (Min-Jen and Cheng-Tao, 2024).

- Furthermore, this reduction in complexity results in a model that is not only considerably smaller in terms of storage requirements but also often exhibits enhanced speed during both the training phase and the subsequent execution phase.
- Additionally, the architecture of the convolutional layer, along with the weights learned during training, remains invariant to the size of the input image, thereby facilitating their easier reuse across different tasks and datasets.
- Moreover, the property of convolutions grants a remarkable translation invariance, meaning that the features are detected with the same efficiency and accuracy, irrespective of their positional location within the image.
- This translation invariance, or more precisely referred to as equivariance when discussing a singular convolutional layer, is of paramount importance for achieving effective generalization, particularly in scenarios where the objects that are being detected may appear in various random configurations within the images, which is a common occurrence in real-world applications. This capability allows the network to proficiently detect each object or feature with equal effectiveness, regardless of its location within the frame, thereby significantly enhancing the overall robustness and performance of the model in practical implementations (Ishrak et al., 2024).

Figure 20 shows a CNN architecture, including the input layer, convolution layers, and pooling layers.

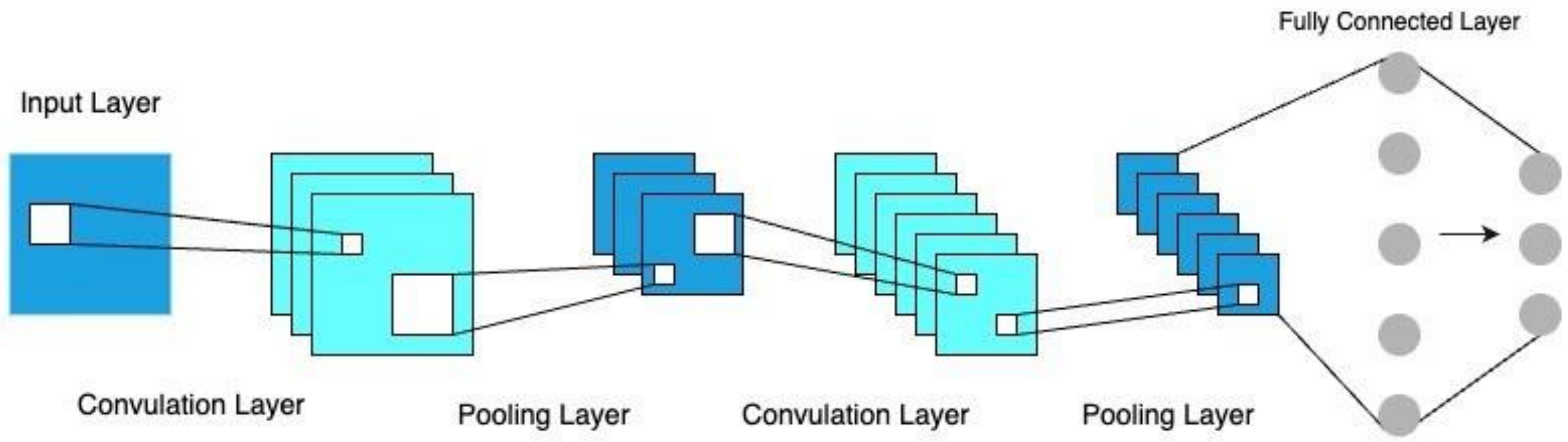

**Fig. 20**: Mechanism of Convolution Neural Network.

In the field of Deepfake detection, researchers have proposed various DL-based methods. One notable approach is the YOLO-CNN-XGBoost method (Ismail et al., 2021), (Wajid et al., 2024), which utilized the YOLO detector trained to detect tight bounding boxes of faces. To enhance deepfake detection, the size of the detected bounding box is increased by 22% to capture more area around the face. The face photos are then resized to 224×224 and passed through a pre-trained CNN model, specifically InceptionResNetV2, to extract discriminant spatial features. These spatial-visual features are subsequently fed into an XGBoost recognizer to distinguish between real and deepfake videos. Another novel technique the authors Shin je et al. (Raza et al., 2022) presented is the deepfake predictor (DFP) approach, which combines VGG16 and CNN architecture. This approach replaced the top layers of the VGG16 (Jiang et al., 2021) model with a multi-layer perceptron (MLP) block. By hyper-parameterizing the DFP approach, the researchers achieved high accuracy in deepfake detection. This study marks the first instance of employing a hybrid of transfer learning and DL-based neural network architecture for effective deepfake detection.

In the realm of facial video forgery detection, another interesting study presented MesoNet, a compact network consisting of two variants: Meso-4 and MesoInception-4. The Meso-4 network comprises four layers of successive convolutions and pooling, followed by a dense network with one hidden layer. To improve generalization and prevent the vanishing gradient effect, ReLU activation functions and Batch Normalization are incorporated in the convolutional layers (Afchar et al., 2018). MesoInception-4, an alternative structure, replaces the first two convolutional layers of Meso-4 with a variant of the inception module (Afchar et al., 2018). Remarkably, both networks achieve commendable scores of approximately 90% when considering each frame independently, showcasing their efficacy in deepfake detection.

Study (Malolan et al., 2020) contributed to the field of Deepfake detection by developing interpretable and explainable models using DL techniques. Their approach involved training a CNN on a face database and employing two explainable AI techniques, namely Layer-Wise Relevance Propagation (LRP) and Local Interpretable Model-Agnostic Explanations (LIME), to visualize the protruding regions in the images. The authors presented a comprehensive set of explainable results, including heat maps, image slices, and input perturbation, demonstrating the model's rotational invariance and robustness in detecting deepfake videos. Their work adds to the growing body of research focused on building transparent and interpretable models for accurate deepfake detection. In paper (Amerini et al., 2019), the authors proposed an innovative approach to distinguish between deepfake videos and genuine ones by leveraging dissimilarities in the optical flow field. This technique focuses on detecting anomalies in the temporal dimension of video sequences. The initial experiments involve representing motion vectors as a 3-channel image, which serves as input for a neural network. The preliminary results obtained on the FaceForensics++ dataset using different types of networks show promising outcomes. These findings highlight the potential of optical flow fields as a feature to identify dis-homogeneities between deepfake and authentic videos.

In a recent study, a binary classification model proposed for deepfake detection, D-CNN, utilized CNN to extract deep features from input images. Also, a deep layer is added to the CNN model to enhance the depth of feature extraction. By performing convolution operations on the images at earlier stages, the D-CNN model could extract deeper and more informative features that can effectively classify the input images into either genuine (Ir) or deepfake (Idf) classes. The D-CNN approach leveraged the capabilities of CNNs to distinguish between authentic and manipulated images accurately (Wajid et al., 2023). To explore further in the application of CNNs in Deepfake detection, Ha et al. (Ha et al., 2023) introduced a robust DeepFake detection method proposed by combining vision transformer (ViT) and CNN models. The experiments demonstrate the ViT model's effectiveness in handling side faces and low-quality videos. The proposed method combined the ResNeSt269 and DeiT models, incorporating a weighted majority voting ensemble (WMVE) approach. This ensemble technique improved the overall detection performance by leveraging the strengths of both models. By integrating ViT and CNN models, the proposed method offers a promising approach for detecting DeepFakes with improved accuracy and robustness. Similarly, a hybrid method was proposed in a study (Heo et al., 2023) that combines vector-concatenated CNN features with patch-based positioning to interact with all positions and identify artefact regions effectively. The model incorporated a distillation token trained through binary cross entropy using the sigmoid function, which enhances its generalization and performance. Experimental results demonstrated that the proposed model achieved superior performance compared to the state-of-the-art model, with an improvement of 0.006 in AUC and 0.013 in f1 score on the DFDC test dataset. This highlights the effectiveness of the proposed approach in detecting DeepFakes and outperforming existing methods.

Furthermore, an effective approach was proposed to detect Deepfakes generated using the GAN model. The proposed algorithm, DeepVision (Jung et al., 2020), focused on analyzing video blinking patterns, a natural and involuntary action. By fusing the Fast-HyperFace and EAR (Eye Aspect Ratio) algorithms, DeepVision tracks the changes in eye blinking patterns. When the eyes are closed, the EAR value decreases below a predefined threshold in consecutive video frames, indicating potential manipulation. Conversely, the EAR value returns to its normal range when the eyes are open. This analysis allows DeepVision to detect abnormal patterns in eye blinks that may arise from randomly generated blinks or specific algorithms. This approach provides a valuable tool for identifying Deepfakes and enhancing the detection capabilities in the fight against digital misinformation.

In conclusion, CNN-based deep learning techniques have proven to be powerful tools for deepfake detection. These methods leverage the strength of CNNs in extracting spatial features from images and videos, enabling effective discrimination between real and manipulated content. Various approaches, such as YOLO-CNN-XGBoost, VGG16 hybrid models, MesoNet, and combining CNN with other architectures, have shown promising results in detecting deepfakes. Additionally, interpretability and explainability have been addressed through techniques like LRP and LIME, enhancing the trustworthiness of deepfake detection models. Further research and advancements in CNN-based models hold great potential for improving the accuracy and robustness of deepfake detection systems, contributing to mitigating the harmful impacts of deepfake technology on society.

### 4.1.2 Modern CNNs with transfer learning

Harnessing the incredible power of deep CNNs, particularly those equipped with transfer learning functionalities, which include remarkable models such as VGG19 (Agarwal and Ratha, 2024), MobileNet (Arivazhagan et al., 2024), ResNet (Min-Jen and Cheng-Tao, 2024), InceptionNet (Uppal et al., 2024), XceptionNet (Al-Qazzaz et al., 2024), and GoogleNet (Abhineswari et al., 2024), among a plethora of others, presents a multitude of advantages that can effectively tackle the formidable challenges associated with deepfake detection in today's digital landscape as follows.

- These sophisticated pre-trained CNN models exhibit exceptional proficiency in the extraction of hierarchical and intricate features from images, meticulously analyzing various layers to unveil the underlying structure. By strategically leveraging these well-honed models for the task of deepfake detection, the identification of subtle inconsistencies and artefacts are effectively done that serve as indicative signs of manipulated or altered content, thus enhancing the detection capabilities significantly (Suratkar and Kazi, 2023).
- Through the innovative transfer learning approach, these pre-trained CNN architectures demonstrate their adaptability by harnessing and building upon the vast knowledge they have acquired from extensive and diverse datasets. By fine-tuning these models specifically on deepfake datasets, researchers facilitate their adaptation to the unique nuances and characteristics inherent to deepfakes, consequently boosting detection accuracy while simultaneously minimizing the need for large volumes of training data, which can often be a significant hurdle.

- Notable architectures such as ResNet, XceptionNet, and InceptionNet are widely celebrated for their remarkable resilience to variations in data inputs and their exceptional capacity to discern complex patterns within that data. This inherent robustness is crucial when detecting deepfakes, which frequently involve elaborate and sophisticated manipulations and forgeries that would otherwise go unnoticed by less resilient models.
- Models like MobileNet and GoogleNet have been meticulously designed to be lightweight and efficient, making them particularly well-suited for applications that require real-time deepfake detection. Their remarkable ability to deliver rapid inference speeds without sacrificing accuracy proves to be invaluable in the urgent task of swiftly identifying and flagging deepfake content as it emerges in our increasingly digitalized environments, instilling confidence in their real-time capabilities.
- Deep CNN models with transfer learning capabilities are adept at generalizing well to previously unseen data, ensuring reliable and consistent performance when detecting a diverse range of deepfake techniques. This essential capability is crucial for effectively addressing the constantly evolving landscape of deepfake technology, which continues to grow and adapt at an alarming rate.
- Ensemble techniques, which strategically combine multiple CNN models, are crucial in enhancing detection accuracy. By leveraging the unique strengths of each architecture, researchers can significantly improve the robustness and reliability of deepfake detection systems, thereby creating a more comprehensive detection framework (Thing, 2023).
- CNN models excel at learning high-level abstract features, which provides a deeper understanding of the detection results. By thoroughly analyzing the activations and responses from various layers within these sophisticated networks, it can be possible to gain valuable insights into the discriminative features used for deepfake detection, thereby enhancing our comprehension of these advanced models.

By leveraging the extraordinary capabilities of deep CNN models equipped with transfer learning functionalities, such as VGG19, MobileNet, ResNet, InceptionNet, XceptionNet, GoogleNet, and a host of others, dramatically enhance the efficiency, accuracy, and robustness of deepfake detection systems can be achieved, thereby providing crucial support in mitigating the myriad challenges posed by the increasingly sophisticated technologies that underpin deepfake creation and dissemination. Table 4 shows the advanced CNN models with transfer learning and their advantages and drawbacks.

Table 4: Summary of Convolutional Neural Networks, Recurrent Neural Networks, and Generative Adversarial Network in deepfake detection

| Models | Methods | Benefits | Drawbacks | Refs |
|---|---|---|---|---|
| CNN | SCNN | interpretable features, scalability, Good in learning hierarchical features | capturing global context and long-range dependencies | Ahmed et al. (2022) Liu et al. (2021) Kolagati et al. (2022) Soudy et al. (2024) |
| | VGG | Transfer Learning, Robustness, interpretable Features | Overfitting, Computationally expensive, Feature redundancy | Chang et al. (2020) Boongasame et al. (2024) Agarwal and Ratha (2024) Dincer et al. (2024) |
| | ResNet | Prevent vanishing gradient problem, reuse of learned features | Increased complexity, Fine-tuning challenges, capturing global context | Nawaz et al. (2024b) Min-Jen and Cheng-Tao (2024) Reis and Ribeiro (2024) Romeo et al. (2024) |
| | Xception Net | Better interpretability, improved performance, depthwise separable convolutions | Computationally expensive, Fine-tuning complexity, Transfer learning variability | Al-Qazzaz et al. (2024) Ciamarra et al. (2024) Deng et al. (2024) Parmar and Mala (2024) Tânțaru et al. (2024) Wang et al. (2024) |
| | Google Net | Parallel feature extraction, Strong performance in image classification | Computational complexity, Long training time | Li et al. (2020) Usmani et al. (2024) Gao et al. (2024) Zou et al. (2024) |
| | Inception Net | Good in extract complex patterns and enhancing generalization | Resource-intensive and potentially unsuitable for real-time deepfake detection | Al-Qazzaz et al. (2024) Uppal et al. (2024) Tiwari et al. (2024) Agarwal and Ratha (2024) Ramadhani et al. (2024) Liu et al. (2024) |
| | Mobile Net | lightweight and computationally efficient, Scalability | requiring careful adjustments to prevent overfitting or underfitting. | Leporoni et al. (2024) Arivazhagan et al. (2024) Kumar and Kundu (2024) Thakur and Rohilla (2024) |
| | IncepRes Net | Enhance deepfake detection capabilities, Parallel feature extraction | Computational complexity, challenging to interpret | Kothandaraman et al. (2024) Nawaz et al. (2024b) Uppal et al. (2024) Jayashre and Amsaprabhaa (2024) |
| | EfficientNet | Computationally efficient, real-time applications, high performance | Demand significant computational resources, Interpretability | Cunha et al. (2024) Ghasemzadeh et al. (2024) Albahli and Nawaz (2024) Thakur and Rohilla (2024) |
| | Dense Net | Feature reuse, mitigates vanishing gradient problem | Memory Consumption, computational complexity, Compatibility | Muthukumar et al. (2024) Khalid et al. (2024) El-Gayar et al. (2024) Kaddar et al. (2024) |
| | HR Net | High-resolution feature representations, Improved Accuracy | computational intensity, Memory usage, Training complexity | Li et al. (2024) Heidari et al. (2024) Zhang et al. (2024) Tsai et al. (2024) |
| RNN | LSTM | Capturing sequential patterns, learn complex relationships and dependencies | Vanishing gradient, Model interpretability, overfitting | Nawaz et al. (2024a) Kosarkar and Sakarkar (2024) Nawaz et al. (2024b) Sekar et al. (2024) Wu et al. (2024) Deng et al. (2024) |
| | BiLSTM | comprehensive understanding of temporal relationships, Adaptability and Flexibility | Increased Computational Complexity, Training and Parameter Tuning | Zhang et al. (2024) Merryton and Gethsiyal Augasta (2024) Rabhi et al. (2024) Liu et al. (2024) Ayetiran and Özgöbek (2024) Kumari and Prasad (2024) |
| | GRU | Excel at capturing sequential patterns, Efficient Training, Gradient Flow | Low generalization, overfitting, challenging to interpret | Kosarkar and Sakarkar (2024) Asha et al. (2024) Amerini et al. (2024) KASIM (2024) |
| GANs | SGANs | Learn discriminative features, generating synthetic examples, Baseline Comparison | Computationally intensive and challenging to optimize, Limited Generalization | Gowrisankar and Thing (2024) Ben Aissa et al. (2024) Tan et al. (2024b) Guarnera et al. (2024) Albahli and Nawaz (2024) Lu and Ebrahimi (2024) |
| | StackGAN | Generating high-resolution and realistic images, Multi-Stage Generation, Variability and Diversity: | Complexity and Resource Intensiveness, Training Instability, interpretability | Kesarwani and Rai (2024) Jiang et al. (2024) Guo and Gu (2024) Luo et al. (2024) |
| | Self-Attention GAN | Global Context Understanding, Selective Focus, adaptability and flexibility | Generalization to New Manipulations, instability during optimization or convergence issues | Wang et al. (2024) Omar et al. (2024) Zhang et al. (2024) Zhang et al. (2024) Kaddar et al. (2024) |

Table 5: Summary of Deepfake Detection Linear and Nonlinear Machine Learning Methods.

| Models | Methods | Benefits | Drawbacks | Refs |
|---|---|---|---|---|
| Linear | LR | Computationally efficient, low complexity, good baseline model, easily interpretable results | Poor ability in complex non-linear models, assumption of linearity, overfitting, poor in feature engineering | Abdali et al. (2021) |
| | LoR | Feature importance, probabilities and easily interpretable coefficients, computationally efficient | linear decision boundary, limiting to capture intricate and non-linear patterns, sensitive to class imbalances | de Weever et al. (2020) Matern et al. (2019) |
| | LDA | Dimensionality reduction, supervised learning, extract features that maximize class separability | Sensitive to Outliers, struggle with complex multi-class classification tasks, curse of dimensionality, Limited non-linearity | Korshunov and Marcel (2018) Korshunov and Marcel (2019) Chakravarty and Dua (2024) Guarnera et al. (2020) |
| | SVM | Robust in overfitting, well in high-dimensional spaces, effective in non-linear data, generalization performance | computationally expensive, sensitive to the choice of the kernel, difficult to interpret, sensitive to noisy data, outliers, and mislabeled instances | Korshunov and Marcel (2019) Chakravarty and Dua (2024) Rafique et al. (2023) Yang et al. (2019) Guarnera et al. (2020) Sandotra and Arora (2024) Saha et al. (2024) |
| | PCA | Reduce the dimensionality of the data, filtering out noise and irrelevant features, visualizing high-dimensional data | Loss of Interpretability, sensitive to outliers, limited to capturing linear relationships | Korshunov and Marcel (2019) Al_Dulaimi and Ibrahim (2023) Huang et al. (2020) Chen et al. (2022) Korshunov and Marcel (2018) |
| Nonlinear | MLP | Good in capturing complex non-linear relationships, Scalability, High generalization abilities | Extensive hyperparameter tuning, overfitting, challenging to interpret, significant computational resources | Essa (2024) Rana et al. (2021) Mallet et al. (2023) Zhao et al. (2022) Raza et al. (2022) Aghasanli et al. (2023) |
| | FFBP | Model complex non-linear relationships, Scalability, Generalization, Adaptability | Complexity, Overfitting, difficult to interpret, Data requirements | Shad et al. (2021) Sahla Habeeba et al. (2021) RATHGEB et al. Abualigah et al. (2024) Sadiq et al. (2023) |
| | KNN | Non-parametric algorithm, No training phase, Robust to outliers, transparency in decision-making | Computationally expensive, Curse of dimensionality, Sensitive to noise, Need for optimal k | Chakravarty and Dua (2024) Rafique et al. (2023) Rafique et al. (2021) Sebyakin et al. (2021) Sridhar et al. (2023) Cocchi et al. (2023) Sandotra and Arora (2024) |
| | DT | Easy to interpret and visualize, Handling non-numeric data, Robust to Outliers | Biased trees towards classes with more instances, may not capture complex relationships | Rana et al. (2021) Solaiyappan and Wen (2022) Patel et al. (2020) Siegel et al. (2021) Hamza et al. (2022) Sandotra and Arora (2024) Rana and Sung (2024) |
| | RBFN | capture complex non-linear relationships, universal approximators, Scalability | Overfitting, Computational Cost, Limited interpretability | Hamza et al. (2022) Yang (2021) Saber et al. (2022) Javed et al. (2022) Wang et al. (2022) Saha et al. (2024) Guhagarkar et al. (2022) |

Table 6: Summary of Deepfake Detection Ensemble and Probabilistic Machine Learning Methods

| Models | Methods | Benefits | Drawbacks | Refs |
|---|---|---|---|---|
| Probabilistic | NB | Easy-to-understand, Efficiency, Handling missing data, Scalability | Assumption of feature independence, sensitive to imbalanced datasets, Limited decision boundary | Chakravarty and Dua (2024) Hassan et al. (2021) Aghasanli et al. (2023) Hamza et al. (2022) |
| | GMM | Clustering capabilities, Scalability, used for semi-supervised learning | Computationally intensive, Sensitive to initialization, Model complexity | Frank and Schönherr (2021) Raza and Malik (2023) Giudice et al. (2021) |
| | HMM | well-suited for modelling sequential data, Unsupervised learning abilities, Interpretability | limited memory capacity, Difficulty with high-dimensional data, computationally demanding | Fagni et al. (2021) Khochare et al. (2021) Borrelli et al. (2021) Sridhar et al. (2023) Bilika et al. (2023) |
| | BN | uncertainty to be explicitly represented, Interpretability, Causal inference, Handling incomplete data | Computationally intensive, Data requirements, Assumptions of conditional independence | Joshi (2022) Zhou et al. (2023) Lin et al. (2021) |
| | MRF | Incorporating prior knowledge, Handling noisy data, efficient inference algorithms | Computationally demanding, Model tuning, Boundary effects, Poor interpretability | Nguyen et al. (2019) Fagni et al. (2021) Khochare et al. (2021) |
| Ensemble | GBM | High accuracy, Feature importance, Robustness to overfitting | computationally intensive, Need hyperparameter tuning, sensitive to noisy data or outliers | Ugale and Midhunchakkaravarthy (2023) Ramkissoon et al. (2022) Kaliyar et al. (2021) Xu et al. (2024) Sadiq et al. (2023) |
| | LightGBM | Handling large datasets, high predictive accuracy and generalization capability | Sensitive to hyperparameter, Limited interpretability, Potential overfitting | Rajkumar et al. (2024) Chen et al. (2022) Ugale and Midhunchakkaravarthy (2023) Sun et al. (2022) Corcoran and Kougianos (2022) Dhiman et al. (2024) Deb et al. (2023) Guo et al. (2023) |
| | XGBoost | High predictive accuracy, Scalability, Flexibility, Regularization | Tuning of hyperparameters, computationally intensive, sensitive to imbalanced datasets, Limited Explainability | Ismail et al. (2021) Hamza et al. (2022) Kaliyar et al. (2021) Saadi and Al-Jawher (2023) Huda et al. (2024) Naskar et al. (2024) |
| | AdaBoost | Robust to noisy data and outliers, simple and intuitive to implement, Robustness to overfitting | Vulnerability to data imbalance, computationally expensive, Limited flexibility | Kumar et al. (2024) Remya Revi et al. (2021) Khalid et al. (2024) Holla and Kavitha (2024) Kulangareth et al. (2024) Oulad-Kaddour et al. (2023) |
| | RF | Feature importance, effectively capture non-linear relationships, Scalability | Computationally intensive in large datasets, Sensitive to noisy data, Hyperparameter tuning | Chakravarty and Dua (2024) Patel et al. (2020) Sandotra and Arora (2024) Hamza et al. (2022) Chakravarty and Dua (2024) Rana and Sung (2024) Sridhar et al. (2023) Yu et al. (2021) |

### 4.1.3 Recurrent Neural Network:

In the ever-evolving landscape of ML algorithms, particularly in the past few years, RNNs have not only risen to prominence but have also garnered significant acclaim for their exceptional capabilities, leading to their widespread application in the intricate domain of time series forecasting (Weerakody et al., 2021), natural language processing, deepfake detection which encompasses various tasks such as speech recognition, forecasting, image classification and machine translation, among others, all due to their extraordinary effectiveness in managing time series data that exhibits long-term dependencies (Das et al., 2023). The fundamental concept that underpins RNNs lies in their ingenious built-in loop structure (See Figure 21), where this unique looping mechanism facilitates the seamless transfer of information from one layer of the network to the subsequent layer, thereby creating a continuous flow of data. This intrinsic chain-like nature of RNNs illustrates how these cyclic neural networks are intricately linked to sequence-related challenges, which include but are not limited to, speech recognition and language modeling, both of which are critical for advancing human-computer interaction. Moreover, RNNs possess the remarkable ability to allow each value from a particular time step to utilize shared parameters. This means that the statistical significance of different time steps can be collectively harnessed at adjacent time steps, thereby significantly amplifying their capacity for feature extraction. All of these inherent characteristics and advantages contribute to the impressive performance of RNNs, enabling them to surpass the capabilities of simpler multilayer perceptron's and various other DL architectures that lack sophisticated frameworks.

Nevertheless, despite their remarkable advantages, the practical implementation of RNNs can present certain challenges as follows.

- Especially when there is a substantial gap between the previously received information and the current forecasting position, which can hinder the RNN's ability to effectively retain and recall previous data, leading to notable difficulties such as the gradient disappearance or gradient explosion phenomena (Weerakody et al., 2021).
- Specifically, this issue arises when the gradient signal interacts with the weight matrix of neurons traversing the hidden layers; if this weight matrix is either excessively small or excessively large, it ultimately results in the gradient signal diminishing to a point where learning effectively ceases, or conversely, expanding to such an extent that the learning process diverges uncontrollably.
- Traditional RNNs, designed for sequential data, face a significant challenge in effectively capturing long-range dependencies due to the vanishing gradient problem. This problem, akin to a silent predator, gradually erodes the model's ability to learn from distant past inputs, thereby critically hampering their overall performance. This is especially evident in complex tasks that require a deep understanding of context across a vast span of the input sequence, often leading to suboptimal results.
- In addition to their struggles with long-range dependencies, traditional RNNs need to improve in incorporating attention mechanisms, which are essential for homing in on the most relevant segments of the input sequence that require emphasis and consideration (Hewamalage et al., 2021). By integrating attention mechanisms, these models gain the ability to selectively weigh various parts of the input data with a newfound precision, significantly enhancing their performance across multiple tasks such as machine translation, where every word's context is crucial, and image captioning, where describing an image accurately depends on focusing on the correct visual elements.
- Moreover, RNNs are alarmingly susceptible to overfitting, particularly when faced with intricate datasets or when the model's architecture is not suitably regularized, which can lead to disastrous consequences (Rithani et al., 2023). Overfitting occurs when a model learns the noise and fluctuations in the training data rather than the underlying patterns, causing it to perform exceedingly well on the training set but fail to generalize effectively on unseen data, which is the ultimate test of any robust ML model.

Consequently, these challenges create a scenario where the outcome of the forecasting task is far from ideal, leaving room for further enhancement and exploration in the realm of RNN applications and improvements.

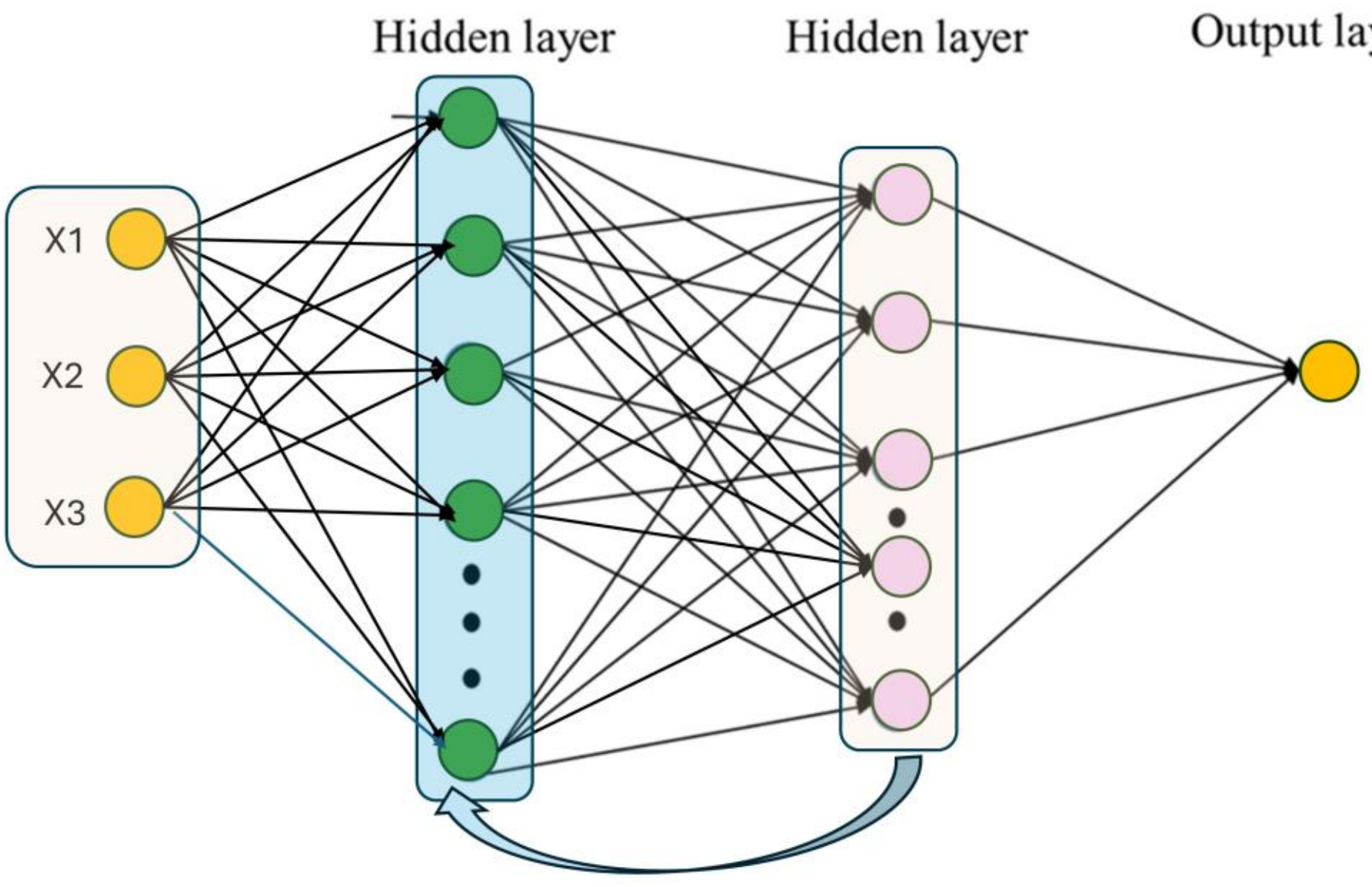

**Fig. 21**: Mechanism of Recurrent Neural Network

Deepfake detection has been a critical research area due to the growing concerns surrounding the proliferation of manipulated videos. In recent years, RNNs have emerged as a powerful tool for analyzing sequential data and capturing temporal dependencies. In this context, this paper introduces an innovative deepfake detection technique based on RNNs (Wajid et al., 2024). Figure 21 shows the general working mechanism of RNNs. Specifically, in paper (Zhao et al., 2023), the authors propose an Interpretable Spatial-Temporal Video Transformer (ISTVT), which leverages a decomposed spatial-temporal self-attention mechanism and a self-subtract mechanism to capture spatial artefacts and temporal inconsistencies in deepfake videos effectively. Moreover, using a relevance propagation algorithm, the ISTVT model allows for interpretability by visualizing the discriminative regions in both the spatial and temporal dimensions.

The detection of deepfake media has predominantly relied on CNN based discriminators, which focus primarily on the spatial characteristics of individual video frames. However, these approaches often overlook the temporal information in the inter-frame relations, limiting their ability to identify deepfakes accurately. A novel approach is proposed that addresses this limitation by leveraging optical flow-based feature extraction to capture temporal dynamics. These extracted temporal features are then utilized in a hybrid model combining CNN and RNN architectures. This hybrid model demonstrates promising performance on popular open-source datasets such as DFDC, FF++, and Celeb-DF. By integrating spatial and temporal information, the hybrid CNN-RNN model offers a more comprehensive and effective solution for deepfake detection (Saikia et al., 2022).

A similar approach that combines both CNN and RNN architectures to integrate spatial and temporal features has been demonstrated in several papers, including (Heo et al., 2021; Al-Dhabi and Zhang, 2021; Jiwtode et al., 2022). These studies recognize the importance of capturing both the spatial attributes of individual frames and the temporal dynamics between frames for effective deepfake detection. By incorporating RNN components into the detection models alongside CNNs, these approaches aim to leverage the strengths of both architectures to enhance the overall detection performance. Integrating spatial and temporal information through the combination of CNN and RNN architecture represents a promising direction in deepfake detection research, offering improved accuracy and robustness in identifying manipulated media. In study (Li et al., 2018), a method is described for identifying videos containing fake faces generated by deep neural network models. This approach focuses on detecting the blink of an eye, which is typically not replicated convincingly in fake videos. The method combines CNNs and Long-term Recurrent CNNs (LRCN) to differentiate between open and closed-eye states. By leveraging the temporal information captured by LRCN in conjunction with the spatial features extracted by CNNs, the method aims to uncover discrepancies in eye movements that can be indicative of deepfake videos.

In summary, DL techniques, including CNNs, RNNs, and hybrid architectures, have shown promise in detecting deepfake videos. By leveraging spatial, temporal, and unique features, these models strive to identify manipulated content and preserve the integrity of digital media. Continued research in this area is essential to develop robust and effective methods for combating the challenges posed by deepfakes and ensuring trust in the digital landscape.

### 4.1.4 Multi-Task Cascaded Convolutional Networks:

Multi-task cascaded Convolutional Networks (MTCNNs) (Patel et al., 2020) have emerged as a popular deep-learning technique for deepfake detection. In a study, a proposed comprehensive system based on MTCNNs consists of three phases: face detection and extraction, feature extraction, and classification. The first phase involves detecting and extracting faces from the video using the MTCNN model, known for its efficiency and effectiveness in face and facial landmark detection. The second phase focuses on feature extraction, where the face images are transformed into meaningful features using various feature extractors such as VGG16, ResNet50, InceptionV3, MobileNet, and DenseNet. These extracted features play a crucial role in distinguishing between real and fake images. Finally, in the last step, the generated features are classified as either real or fake, enabling the system to make accurate predictions. The MTCNN-based approach offered a robust and efficient solution for deepfake detection by leveraging the power of CNNS and multitasking capabilities.

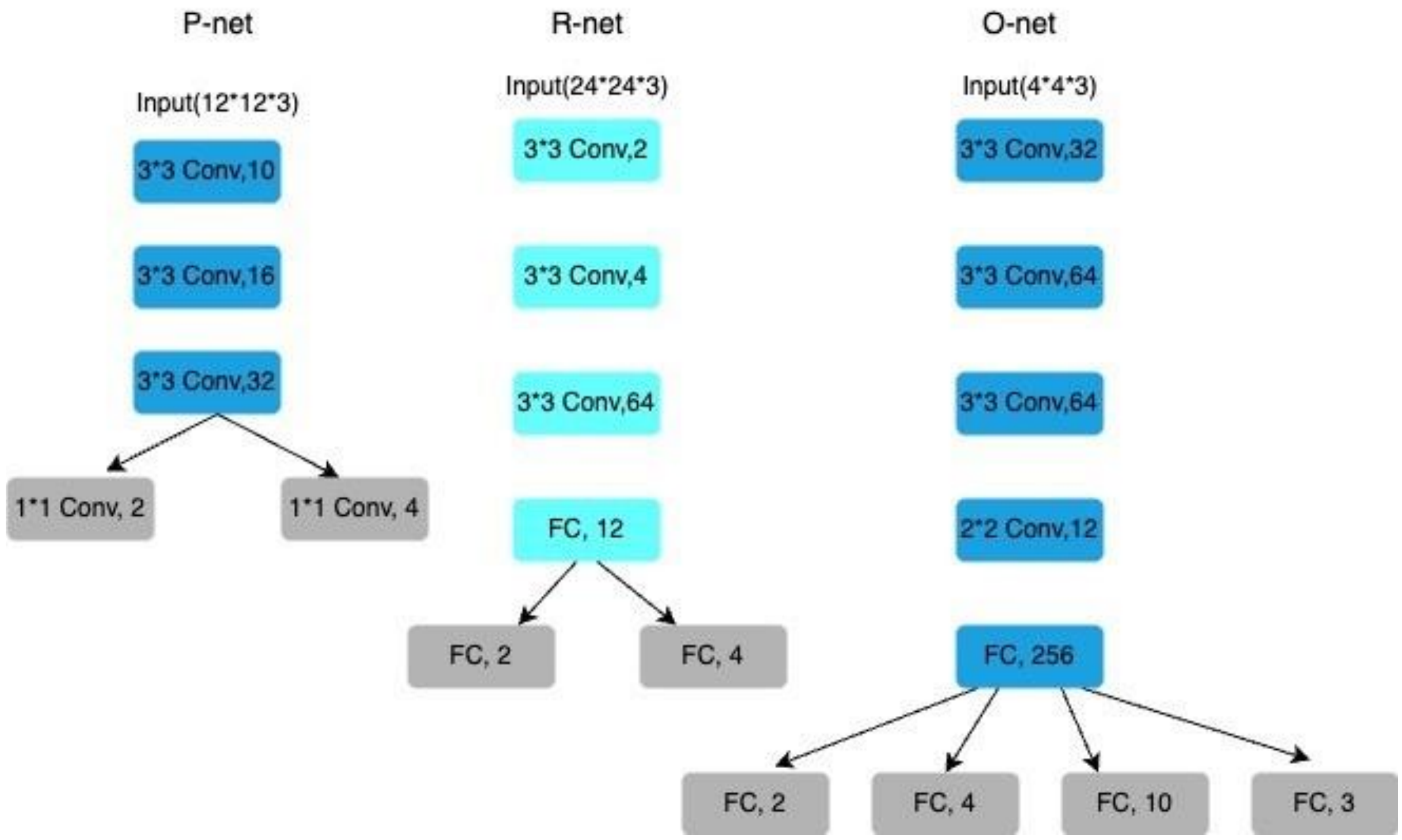

**Fig. 22**: Mechanism of Multi-Task Cascaded Convolutional Networks

Jonsons et al. (Johnson et al., 2022) explored the effectiveness of utilizing the eye region as an area of interest for distinguishing between original and deepfake videos. They employed the MTCNN (Figure 22), a powerful stacked neural network designed for face detection and alignment. By leveraging MTCNN's high accuracy in face detection, the authors can significantly reduce false positive images in the dataset, ensuring that the subsequent analysis focuses on relevant facial regions. Specifically, a region around both eyes was cropped, with additional padding, to serve as input for training a CNN layer. The coordinates returned by MTCNN for the eye regions were used to extract the necessary visual information precisely. By examining the eye region, the research aimed to evaluate its potential as a valuable feature for discriminating between original and deepfake videos.

In the proposed scheme (Qureshi et al., 2021), detecting deepfake videos was approached by splitting the video stream into audio and video frames while maintaining synchronization information. The feature extraction process was then performed on both the speech and video frames using commonly used techniques such as Mel-frequency cepstral coefficients (MFCC) for speech signals and MTCNN for video frames. To further enhance the security and authentication process, secret keys were employed to generate a fragile watermark. This watermark was designed to interlink the speech and video components, creating a cross-reference between them. Additionally, the scheme incorporated the use of a few bits from the video ID to generate a robust watermark that contains both copyright information and related metadata of the video. The detection of deepfake videos involved the application of watermark extraction algorithms and the retrieval of relevant video metadata stored in a blockchain, ensuring the integrity and authenticity of the content. Also, this approach (Qureshi et al., 2021) presented a comprehensive framework for detecting deepfake videos while leveraging multiple modalities and incorporating secure watermarking techniques.

In summary, MTCNNs have proven to be effective DL techniques for deepfake detection. By efficiently detecting and extracting faces from videos, MTCNN enables accurate identification of regions of interest, such as the eye area, which is crucial in distinguishing between genuine and manipulated videos. With its cascaded architecture and integration into feature extraction and classification stages, MTCNN provides a reliable and efficient solution for combating deepfake manipulation. Its continued development and adaptation will be instrumental in staying ahead of evolving deepfake technologies.

### 4.1.5 Hierarchical Multi-Scale Networks:

Deepfake detection is an evolving field that requires innovative techniques to identify artefacts of manipulation effectively. In the study presented in paper Wang et al. (2022), the authors tackle this challenge by focusing on capturing subtle inconsistencies at various scales using the power of transformer models. The proposed approach, known as Hierarchical Multi-Scale Networks (HMNs), introduces a Multi-modal Multi-scale TRansformer (M2TR) framework. This framework operates on patches of different sizes, enabling the detection of local manipulation artefacts across different spatial levels. Moreover, M2TR incorporates the analysis of forgery artefacts in the frequency domain, complementing the RGB information, by leveraging a carefully designed cross-modality fusion block. By employing this approach, the authors aim to enhance the accuracy and robustness of deepfake detection, paving the way for more reliable identification of manipulated content.

Detecting deepfake images requires comprehensive analysis and understanding of the forgery process. In the relevant research work Yang et al. (2021), the authors approached the problem from the perspective of image generation and focused on tracing the potential texture artefacts left behind during the forgery process. To accomplish this, they proposed a multi-scale self-texture attention Generative Network (MSTA-Net) that aimed to track and identify texture traces in the image generation process while mitigating the impact of post-processing techniques employed in deep forgery. The MSTA-Net consisted of a generator with an encoder-decoder structure that disassembled the images and generated trace information. The generated trace image was then merged with the original map and fed into a classifier utilizing Resnet as the backbone. A self-texture attention mechanism (STA) is introduced as a skip connection between the encoder and decoder to enhance the texture characteristics during the disassembly process. A novel loss function called Prob-tuple loss, guided by the classification probability, was also proposed to refine the generation of forgery trace. To validate the effectiveness of the MSTA-Net, the authors conducted various experiments to demonstrate the feasibility and advancements of their proposed method.

In conclusion, the development of HMNs has shown great promise in deepfake detection. By leveraging transformer models and incorporating multi-modal and multi-scale approaches, such as the Multi-modal Multi-scale Transformer (M2TR), HMNs aim to capture subtle manipulation artefacts at different scales. These techniques enable the detection of local inconsistencies in images at various spatial levels and the identification of forgery artefacts in the frequency domain. Additionally, the introduction of self-texture attention mechanisms and loss functions specific to forgery trace generation, as demonstrated in the MSTA-Net, further enhances the detection capabilities of HMNs. The advancements in HMNs provide a robust and comprehensive solution for deepfake detection, paving the way for improved security and trust in digital media content. Continued research and development in this area will undoubtedly contribute to the ongoing battle against deepfake manipulation and its potential misuse.

### 4.1.6 Diffusion-based Models

Diffusion models constitute a fascinating and intricate subset of generative models that are firmly rooted in the expansive realm of DL (Yang et al., 2023), specifically engineered to create synthetic data that is not only imaginative but also strikingly authentic, such as computer-generated artwork resonating with the distinctive style of the legendary artist Picasso, all accomplished by utilizing meticulously specified input parameters that guide the creative process. Historically speaking, the landscape of generative tasks has mainly been dominated by the formidable GANs, which boast a sophisticated structural design composed of a data generator whose role is to fabricate synthetic data and a discriminator tasked with the critical function of discerning between genuine data and its generated counterparts, ensuring the authenticity of the output. However, in recent times, the rise of diffusion-based generative models has captured considerable attention and enthusiasm across a multitude of domains thanks to several pivotal advantages that these innovative models offer, making them an attractive alternative. These diffusion models exhibit an exceptional capability to smoothly and effectively learn complex data distributions, adeptly managing high-dimensional data while generating a remarkable variety of diverse and highquality data outputs that stand out in comparison to their predecessors (Guo et al., 2023). The origins of diffusion models can be traced back to the groundbreaking contributions made by Sohl-Dickstein (SohlDickstein et al., 2015) and his colleagues, whose pioneering work laid the groundwork for subsequent advancements, including the introduction of Denoising Diffusion Probabilistic Models, or DDPM (Ho et al., 2020), which compellingly demonstrated that diffusion models possess the potential to compete with other state-of-the-art generative models in various image generation tasks.

Despite the diverse and extensive range of applications that diffusion models are capable of addressing, it is important to note that they typically rely on three principal formulations, as illustrated in Figure 23, which include (1) Denoising Diffusion Probabilistic Models (DDPMs), (2) Noise-Conditioned Score Networks (NCSNs), and (3) Stochastic Differential Equations (SDEs) (Croitoru et al., 2023). These foundational formulations play a crucial role in elucidating the intricate mechanisms through which diffusion models operate, enabling them to tackle various tasks and challenges across a multitude of domains, thereby solidifying their position as a cutting-edge technology in the ever-evolving landscape of generative modelling.

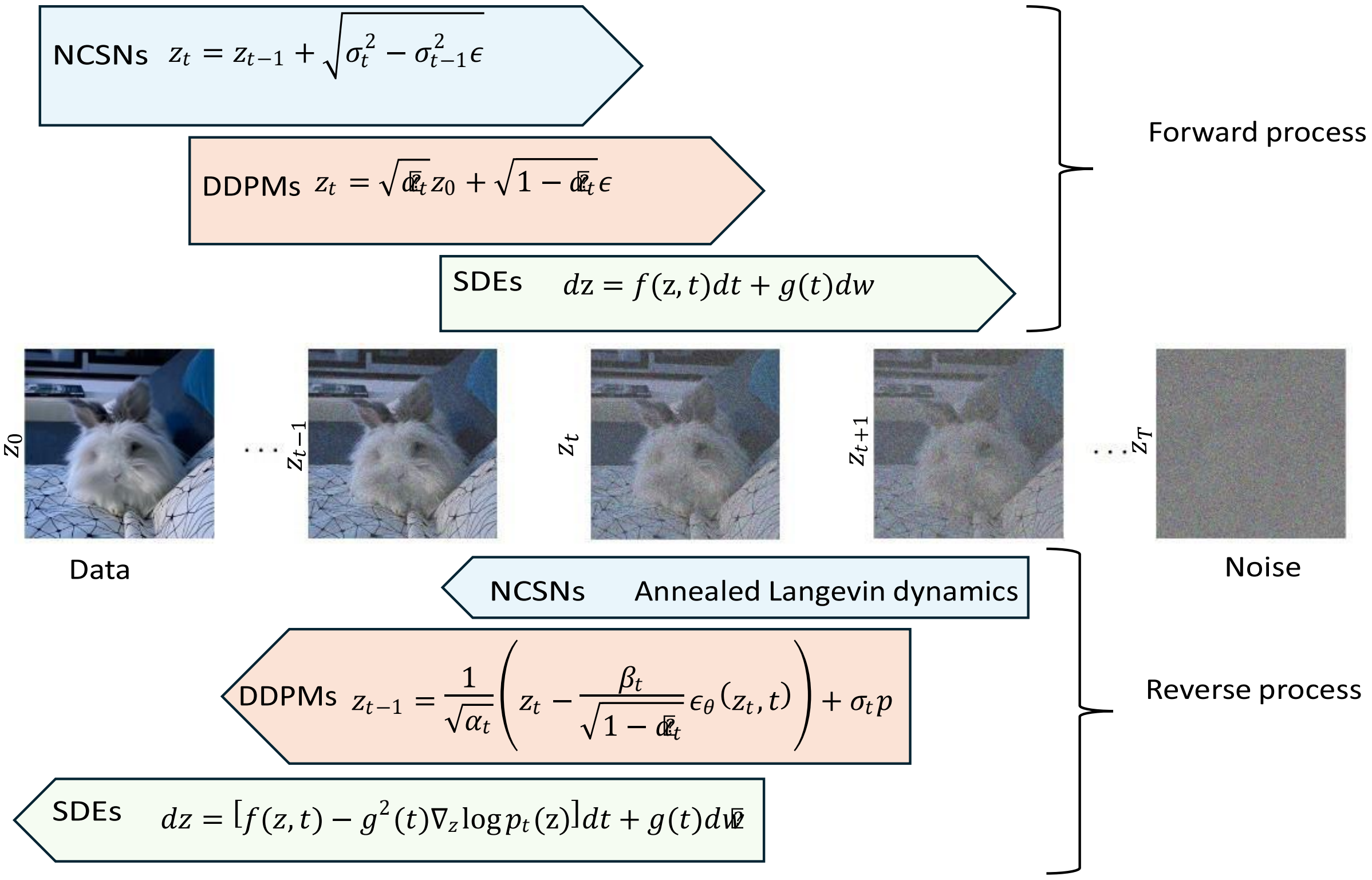

**Fig. 23**: The framework and formulation of three components of diffusion models.

Recent explorations within the fascinating realm of Diffusion-based models have unveiled a plethora of promising breakthroughs that significantly enhance generation capabilities, a fact that is clearly illustrated by the most recent scholarly studies (Ding et al., 2023; Han et al., 2023; Kim et al., 2022; Liu et al., 2024; Zhao et al., 2023). By skillfully leveraging the robust capabilities of the pre-trained Stable Diffusion (Blattmann et al., 2023), researchers have remarkably improved not only the quality of the outputs that are generated but also the overall efficiency of the training of the model itself, thereby highlighting the significant progress that has been made in the realm of advancing facial manipulation technologies. In a particularly noteworthy development, Diff-Swaps (Zhao et al., 2023) has redefined the complex face swapping challenge by creatively framing it as a conditional inpainting task, thereby providing a novel and refreshing lens through which to approach the intricate art of facial manipulation tasks. In a effective contribution to this field, Liu et al. (Liu et al., 2024) have introduced a sophisticated multi-modal face generation framework that skillfully integrates well-balanced identity and expression encoders into the overarching structure of the conditional diffusion model. This thoughtfully produced integration is strategically designed to achieve a satisfying harmony between the crucial aspects of identity substitution and the preservation of various attributes throughout the entire generation process, ultimately resulting in an impressive enhancement of both the fidelity and realism of the faces that are generated. Furthermore, the arrival of a groundbreaking facial generalist model, known as FaceX (Han et al., 2023), vividly illustrates a remarkable and versatile capacity to handle a wide array of diverse facial tasks, which includes a multitude of functionalities such as face swapping and meticulous editing.

## 4.2 Machine Learning based models

ML-based models for deepfake detection offer an alternative approach that can be more efficient than deep learning-based methods. These models require less extensive training data and computing power, making them suitable for scenarios with limited resources. However, it is essential to note that they may not achieve the same level of accuracy in detecting sophisticated deepfakes compared to DL approaches. The trade-off between efficiency and accuracy is crucial when choosing the appropriate detection method. Nonetheless, ML-based models provide a practical solution and have demonstrated their effectiveness in addressing deepfake detection challenges, albeit with varying degrees of success depending on the complexity of the manipulated content.

Various research studies proposed different models for deepfake detection, and one notable study (Rana et al., 2021) challenges the notion that DL-based approaches were the only practical solution. The authors presented their findings, indicating that traditional ML techniques alone can better detect deep fakes. Their ML-based approach followed standard methods of feature development, feature selection, training, tuning, and testing an ML classifier. The advantage of the ML approach lies in its better understandability and interpretability, coupled with reduced computational costs. The authors also reported impressive results on multiple deepfake datasets, with accuracies of 99.84% on FaceForecics++, 99.38% on DFDC, 99.66% on VDFD, and 99.43% on Celeb-DF datasets. These results suggested an effective deepfake detection system can be built using traditional ML methods, offering comparable or superior performance to state-of-the-art DL-based approaches. In alignment with the idea of utilizing classical methods for deepfake detection, another work (Durall et al., 2019) presented a straightforward approach to identifying fake face images. The method primarily relies on frequency domain analysis and a primary classifier. During the pre-processing phase, the images undergo a frequency domain analysis using the Discrete Fourier transform (DFT). A robust 1D representation of the power spectrum was obtained by applying DFT and performing azimuthal averaging. This representation served as the basis for subsequent classification using SVM and k-means clustering (Yazdinejad et al., 2020). The study highlighted the effectiveness of combining classical frequency domain analysis with simple classifiers, demonstrating their potential in detecting deepfake images.

In another example ML-based application was introduced by Mitra et al. (Mitra et al., 2020). They proposed a novel model for deepfake detection, employing a combination of a CNN and a classifier network, and explored the performance of three popular CNN architectures, namely XceptionNet, InceptionV3, and Resnet50, and conducted a comprehensive comparative analysis. Through rigorous evaluation, XceptionNet emerged as the most promising choice among the three, exhibiting superior capabilities for deepfake detection. A specially designed classifier was integrated into the framework to enhance the model's accuracy further. This study sheds light on the significance of selecting the appropriate CNN architecture and highlights the effectiveness of the proposed model in detecting deepfakes. In developing the application of ML models, the authors proposed a method based on the observation that Deepfakes introduced errors in 3D head pose estimation due to the splicing of synthesized face regions into original images (Yang et al., 2019; Wajid et al.). They conducted experiments to demonstrate this phenomenon and develop a classification method using features derived from this cue. The effectiveness of their approach was evaluated by employing an SVM classifier on a dataset consisting of both real face images and deepfakes. In conclusion, ML-based deepfake detection techniques offer a promising approach in the battle against manipulated media. By leveraging the power of advanced algorithms and feature extraction, these ML methods enable efficient and reliable identification of deepfakes. The application of various ML-based approaches, such as linear, nonlinear, probabilistic and ensemble models in improving the deepfake detection problem are listed in Tables 5 and 6. Furthermore, the ML-based models are compared regarding advantages and drawbacks in deepfake detection, briefly discussed in Table 5 and 6.

Recent research has shown that Ensemble ML-based models excel in detecting deepfakes compared to their standalone ML counterparts (Al-Qazzaz et al., 2024; Alghamdi et al., 2024). This advantage arises primarily as follows.

- From the ability of Ensemble models to mitigate overfitting by balancing out the biases and variances of individual models. By pooling predictions from a variety of models, ensemble approaches can better adapt to unfamiliar data, which is essential for identifying intricate deepfake alterations (Tien and Nam, 2024).
- Additionally, as Deepfake techniques consistently advance, spotting manipulated content becomes increasingly difficult. Ensemble models enhance resilience against adversarial threats and innovative deepfake creation methods by synthesizing the outputs of several models (Chakravarty and Dua, 2024). • Beyond the aforementioned benefits, Ensemble models frequently achieve superior accuracy rates than individual models by harnessing the collective wisdom of multiple models (Sekar et al., 2024). This can result in more dependable deepfake detection outcomes.
- Moreover, even if the individual base models are weak classifiers or regressors, amalgamation can yield a formidable model (Soudy et al., 2024). This concept is the foundation of boosting algorithms like AdaBoost and XGBoost (Amerini et al., 2024), which progressively enhance performance by targeting challenging classification cases.
- Ultimately, Deepfake detection datasets frequently experience class imbalance (Sharma et al., 2024), with authentic videos significantly outnumbering deepfake videos. Ensemble models can tackle this challenge by utilizing strategies such as balanced sampling and weighted voting to alleviate the effects of class disparity.

## 4.3 Blockchain Based Methods

In the realm of deepfake detection, current solutions often face challenges when providing comprehensive tracking of the provenance and history of digital media. Traditional methods struggle to trace the origin and maintain the integrity of digital content, especially in scenarios where the content has been copied and distributed multiple times. Blockchain-based solutions offer a promising approach to address the limitations of current deepfake detection methods, as mentioned in Figure 24. These solutions leverage blockchain technology's inherent properties, such as distributed ledger and cryptographic hashing, to track the provenance and history of digital media. By establishing a robust framework for authenticity and traceability, blockchain-based approaches aim to enhance deepfake detection by verifying the origin of content and detecting any subsequent alterations. Blockchain-based solutions for deepfake detection have gained significant attention due to their ability to track the provenance and history of digital content, providing a reliable means of verifying authenticity (Hasan and Salah, 2019). One notable company in this space is Truepic (Wajid et al., 2023), a US-based startup that has developed a robust system comprising mobile apps and server infrastructure. Users, including freelancers, can capture and save images securely, preserving their integrity. By comparing any potentially forged images (Mayer and Stamm, 2020) with the originals stored on their servers, Truepic's system enables easy detection of tampering attempts.

In another paper (Chan et al., 2020), authors presented a decentralized Proof of Authenticity (PoA) system that utilizes blockchain technology to address the challenge of verifying the authenticity of digital videos. The system employs metadata stored in the EXIF format, which includes key attributes such as the Ethereum address of the artist and the smart contract address. The video files, along with their associated metadata, were stored on a decentralized, peer-to-peer file system like IPFS (Benet, 2014). To ensure the integrity of the videos, a unique IPFS hash was generated and linked to a smart contract created by the original artist on the Ethereum blockchain. In the event of any edits or modifications to the video by a secondary artist, a separate smart contract is created, establishing a clear connection to the original video.

This blockchain-based approach provides a transparent and traceable mechanism for verifying the authenticity and provenance of digital videos.

In addition to the aforementioned system, other studies have also highlighted the significance of blockchain technology in the context of deepfake detection. The focus shifts from detecting fake content to providing irrefutable evidence of genuine content (Barclay, 2018). The objective is to establish tamperproof mechanisms that ensure the authenticity of digital media. An effective study Mao et al. (2022) proposed a deep forgery face video detection method known as FFS (fusion of frequency and spatial domain features), incorporating blockchain technology to enhance the detection process. The method involved two phases: model training and updating. The dataset images were processed during the model training phase, and their RGB and HSV representations are obtained (Dahea and Fadewar, 2018).

In response to the escalating threat posed by deepfake videos, a recent study (Heidari et al., 2024) introduced an effective solution that leverages blockchain-based federated learning (FL) to safeguard data source anonymity. The methodology addresses concerns of data heterogeneity from diverse global sources by integrating SegCaps and CNN techniques for enhanced image feature extraction coupled with capsule network (CN) training to improve generalization. Noteworthy components include: i. Introducing a novel data normalization process. ii. Utilizing transfer learning (TL) and preprocessing methods to enhance DL performance. iii. Establishing collaborative global model training facilitated by blockchain and FL technologies while upholding data source confidentiality. Rigorous experimentation validates the approach, showcasing a remarkable average accuracy boost of 6.6% compared to benchmark models and a 5.1% improvement in the area under the curve (AUC), surpassing existing detection techniques. Figure 24 shows a landscape of the BFLDL model with its components.

Frequency domain features are extracted using the Fourier transform, while spatial domain features are extracted using the wavelet transform. These features are fused to create discriminative features, which are then used to train a support vector machine. The trained model is deployed on a smart contract within the blockchain network in the model updating phase. The model is collaboratively updated and maintained through the public blockchain, ensuring transparency and integrity. By leveraging blockchain technology, the proposed method improves the generalization ability of the detection model and addresses the vulnerability to malicious attacks. Additionally, it provides an incentive mechanism for continuous model updates, ensuring a reliable and up-to-date deepfake detection system. In the study (Maurer, 2000), the authors presented a comprehensive examination of information-theoretic lower bounds concerning the probability of cheating in one-way message authentication. The study focused on the authentication of a sequence of messages using a shared secret key between the sender and receiver. The adversary's objective is to deceive the receiver by forging one of the messages in the sequence. The paper effectively explored two types of cheating scenarios: impersonation and substitution attacks. It derived lower bounds on the probability of cheating for any authentication system and investigated three goals the adversary may seek to achieve. The analysis contributed valuable insights into the security and limitations of authentication systems, enhancing our understanding of potential cheating probabilities and associated risks.

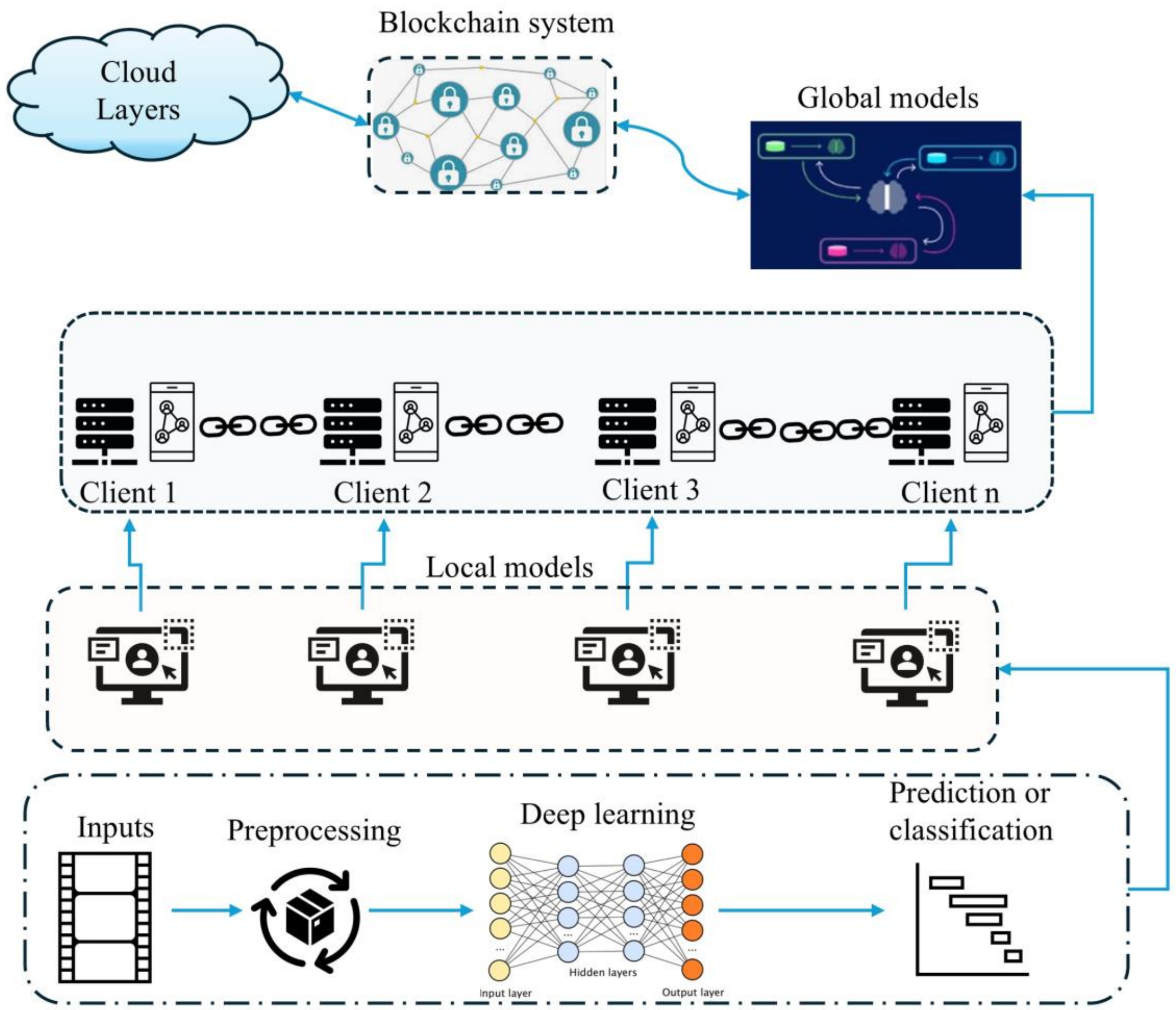

**Fig. 24**: Illustration depicting the BFLDL model, a blockchain-based Federated Learning (FL) approach integrated with DL techniques, designed for the recognition of deepfake videos.

By leveraging blockchain technology, these solutions address a critical limitation of current deepfake detection methods, which cannot often track the history and provenance of digital media. Blockchain enables a transparent and immutable record of content, facilitating the verification of authenticity even if the content is copied multiple times.

## 4.4 Statistical Approaches

Statistical approaches have emerged as significant tools in the realm of deepfake detection. These methods utilize sophisticated statistical analysis techniques to discern patterns and deviations within data, enabling the identification of manipulated content. By examining a range of statistical features, including image quality metrics, noise patterns, facial landmarks, and temporal inconsistencies, these approaches strive to differentiate between genuine and falsified media. Statistical models and algorithms are employed to establish robust statistical distributions and thresholds that effectively classify deepfakes. These approaches offer valuable insights into the statistical characteristics inherent to deepfake content, fostering the development of reliable and accurate detection techniques. Frequently utilized statistical models in deepfake detection include Expectation-Maximization (EM), Total Variational (TV) distance (Guarnera et al., 2020), Kullback-Leibler (KL) divergence, Jensen-Shannon (JS) divergence (Hao et al., 2022), and others. These models are commonly applied to analyze and measure the differences between real and fake data, providing valuable insights for detecting deepfake content.

Statistical approaches have gained significant attention in deepfake detection, and several studies highlight their importance. Varshney, L.R. et al. (Agarwal and Varshney, 2019), the authors present a video-based deepfake detection approach that leverages statistical techniques. The workflow begins with CNN-based face detection, followed by CNN-based facial feature extraction to identify key facial indicators for manipulation detection. An Automatic Face Weighting (AFW) mechanism and a Gated Recurrent Unit (GRU) network are then employed to analyze the facial features and extract meaningful information to verify the authenticity of videos. A Boosting Network is utilised alongside the backbone network to enhance the discrimination between authentic and manipulated videos. This comprehensive approach demonstrates the value of statistical methods in detecting and distinguishing deepfake content with improved accuracy and reliability. Moreover, the paper also proposed a complementary technique called Deepfake Detection via Audio Spectrogram Analysis, which focuses on detecting fake audio. By combining these advancements in visual and audio analysis, the proposed method offers a comprehensive approach to deepfake detection, showcasing the effectiveness of statistical techniques in addressing the challenges posed by deepfake media. In paper (Wang et al., 2022), a quantitative study was conducted to assess the reliability of deepfake detection models using a new evaluation metric and statistical techniques. The study aimed to address the challenges related to the reliability of current deepfake detection research, specifically focusing on transferability, interpretability, and robustness. While previous studies have addressed these challenges individually, the overall reliability of detection models has been overlooked, resulting in a lack of reliable evidence for real-life applications and legal proceedings. To fill this gap, the authors introduced a model reliability study metric that utilizes statistical random sampling and publicly available benchmark datasets. By constructing a population that mimics real-life deepfake distribution and implementing a scientific random sampling scheme, the study analyzed and computed confidence intervals for accuracy and AUC score metrics. This approach enables the derivation of numerical ranges indicating reliable model performance at 90% and 95% confidence levels. These findings contribute to a better understanding of the reliability of existing deepfake detection models when applied to arbitrary deepfake candidates.

A study (Saxena and Cao, 2021) explored the error probability of different GAN implementations from a robust statistics viewpoint. The study establishes bounds on error probability and investigates the performance of GANs. By employing a robust statistics framework, the paper simplifies these bounds using an Euclidean approximation for low-error scenarios. Moreover, it establishes connections between error probability and epidemic thresholds in network spreading processes. This analysis enhances our understanding of GAN behaviour, performance, and potential applications. Overall, these statistical approaches contribute to the ongoing efforts in combating the proliferation of deepfake content and ensuring the authenticity and integrity of multimedia data.

## 4.5 Adversarial Perturbations

Adversarial Perturbations on Deepfake Detectors is a field of research that focuses on the existence of universal perturbations capable of causing misclassification in deepfake detection systems. These perturbations are image-agnostic and extremely small, yet they have the ability to deceive DL models and lead to incorrect classification results (Matern et al., 2019). The discovery of such perturbations has raised concerns about the robustness and reliability of deepfake detection methods, as even subtle modifications in images can potentially bypass the detection mechanisms. In this context, researchers have been exploring techniques to understand and mitigate the vulnerability of deepfake detectors to adversarial perturbations.

In the study (Moosavi-Dezfooli et al., 2017), the authors investigate the use of adversarial perturbations to deceive deepfake detectors. Adversarial perturbations are small modifications applied to images to mislead classifiers. The results show that unperturbed deep fakes achieved over 95% accuracy on detectors, while perturbed deep fakes resulted in less than 27% accuracy. To counter these perturbations, the researchers explore two improvements: Lipschitz regularization and Deep Image Prior. They test the effectiveness of adversarial attacks using the Fast Gradient Sign Method (FGSM) and the Carlini and Wagner L2 Norm attack (CW-L2). The results indicate significant drops in accuracy for both whitebox and blackbox attacks on the VGG and ResNet models. Lipschitz

Regularization, which constrains the gradient of the detector, improves the detection of perturbed images. Additionally, the Deep image prior defence pre-processes the input to remove perturbations before classification. It achieves high accuracy on perturbed deepfakes that fooled the original detector while maintaining good accuracy in other cases. However, the DIP defence has longer processing times, which may limit its practicality in real-time scenarios with resource constraints. In the paper (Gandhi and Jain, 2020), the authors propose an adversarial face generation method to protect individuals' faces by incorporating random differentiable image transformations during the training of Deepfake models. This approach intentionally introduces more artefacts in the synthesised faces, which in turn makes it easier to detect and recognize manipulated images and videos (Yang et al., 2021). The study identifies the increases in adversarial and edge losses as key factors leading to significant degradation in the quality of the generated faces. Extensive experiments are conducted using multiple pairs of faces with varying resolutions under white-box, grey-box, and black-box settings, and the effectiveness and robustness of the defence method are demonstrated using various metrics. The results showcase the potential of the proposed approach in enhancing the detection and mitigation of deepfake manipulation, providing an important step towards safeguarding individuals' identities and privacy (Wajid et al., 2022).

In conclusion, the exploration of adversarial perturbations in the context of deepfake detection has shed light on the vulnerabilities and challenges faced by current detection systems. The ability to deceive deepfake detectors using small perturbations highlights the need for more robust and resilient defence mechanisms. Researchers have proposed various techniques, including regularization methods and deep image priors, to mitigate the impact of adversarial perturbations and improve the accuracy of detection. While these defences have shown promising results, there are still limitations to consider, such as increased processing time and the need for further evaluation under different attack scenarios. As the arms race between deepfake generation and detection continues, ongoing research and advancements in adversarial perturbation techniques are crucial for developing more effective countermeasures and safeguarding the integrity of digital media.

# 5 Analysis of Effective Approaches

Table 7 provides an overview of the methods discussed in this section, offering a concise summary of their key characteristics, findings, and distribution by the publisher. Figure 25 shows the percentage contribution of various methods in Deepfake detection. As can be seen, the deep learning methods participation is around 60% of the total approaches.

**Table 7**: Details of applied modern deep learning methods in Deepfake Detection

| No. | Author Name | Methodology | Dataset | Result |
|---|---|---|---|---|
| 1. | Patel et al. (2023) | Deep Learning (D-CNN Deepfake Detection) | DFDC | 97% |
| 2. | Ha et al. (2023) | Deep Learning (Robust DeepFake Detection, ResNeSt269 and DeiT model) | DFDC | 96.78% |
| 3. | Zhao et al. (2023) | Deep Learning (Interpretable Spatial-Temporal Video Transformer (ISTVT)) | FACEFORENSICS++ (FF++) with HQ and LQ, CELEB-DF, and DFDC | 99.6%, 98.9%, 99.8%, and 92.1% |
| 4. | Wang et al. (2022) | Reliability Perspective | FF++, DFDC, Celeb-DF, DF1.0, and FaceShifter | Best-performed MAT= 97.40%, 66.63%, 71.81%, 41.74%, 18.71% |
| 5. | Raza et al. (2022) | Deep learning (Deepfake predictor (DFP), VG16) | CelebDF and FaceForen-cics++ | 95% precision |
| 6. | Saikia et al. (2022) | Deep Learning (Hybrid CNN-LSTM model) | DFDC, FF++ and Celeb-DF | 91.21% |
| 7. | Jiwtode et al. (2022) | Deep Learning (LSTM) | HOHA | 97% |
| 8. | Johnson et al. (2022) | Deep Learning (SIFT) | FaceForensics++, NeuralTextures dataset | 98.3% |
| 9. | Wang et al. (2022) | Deep Learning (Multi-modal Multi-scale Transformers, M2TR) | SR-DF | 95.31 AUC |
| 10. | Rana et al. (2021) | Machine Learning (MLP, SGB) | FaceForecics++, DFDC, VDFD | 99.84% |
| 11. | Mao et al. (2022) | Blockchain (FFS) | DeepfakeDetection and Celeb-DF datasets | 90.17 AUC |
| 12. | Hao et al. (2022) | Multiple Data Modalities (GRU and ReLU) | ASVspoof2019 | 89.75 AUC |
| 13. | Ismail et al. (2021) | Deep learning (YOLO-CNN-XGBoost, Inception V3) | CelebDF-FaceForencics++ (c23) | 90.62% AUC |
| 14. | Heo et al. (2021) | Deep Learning (CNN and ViT) | DFDC | 99.3 AUC |
| 15. | Al-Dhabi and Zhang (2021) | Deep Learning (Resnext50) | DFDC, FaceForensics++, Celeb-DF | 95.5% |
| 16. | Qureshi et al. (2021) | Deep Learning (Digital Watermarking, Wav2Lip algorithm) | Hybrid dataset | PSNR=29.41 |
| 17. | Yang et al. (2021) | Deep Learning (Forgery Detection, MSTA-Net) | FaceForensics++ c23, FaceForensics++ c40, Deeperforensics, Celeb-DF, and DFDC | 98.96%, 95.31%, 99.86%, AUC=0.99, AUC=0.95 |
| 18. | Mitra et al. (2020) | Machine Learning (XceptionNet, InceptionV3 and Resnet50) | FaceForensics++ | 96% |
| 19. | Malolan et al. (2020) | Deep learning (Explainable Deep-Fake Detection) | Face Forensics Deep Fake Detection | 94.33% |
| 20. | Jung et al. (2020) | Deep Learning (Fast-HyperFace and EAR) | Eye Blinking Prediction dataset (Kaggle) | 87.5% |
| 21. | Patel et al. (2020) | Deep Learning (Trans-DF, VGG16, ResNet50) | FF++, DFDC, Celeb-DF | 90.2% |
| 22. | Yazdinejad et al. (2020) | Machine Learning (Unmasking DeepFakes, SVM) | CelebA ,FaceForensics++ | 91% |
| 23. | Durall et al. (2019) | Machine Learning (Unmasking DeepFakes, SVM) | CelebA ,FaceForensics++ | 91% |
| 24. | Amerini et al. (2019) | Deep learning (Optical Flow based CNN, PWC-Net) | FaceForensics++ | 81.61% |
| 25. | Li et al. (2018) | Deep Learning (In Ictu Oculi, LRCN VG16) | CEW Dataset | 0.99 AUC |
| 26. | Yang et al. (2019) | Machine Learning (SVM) | UADFV, DARPA GAN Chal-lenge dataset | AUC=0.89, 0.84 |
| 27. | Hasan and Salah (2019) | Ethereum Blockchain | PROVER | Cost: 0.095USDper transaction |
| 28. | Afchar et al. (2018) | Deep learning (MesoNet,56 Misconception-4) | Face Forensics, Deepfake Dataset, | 98% |

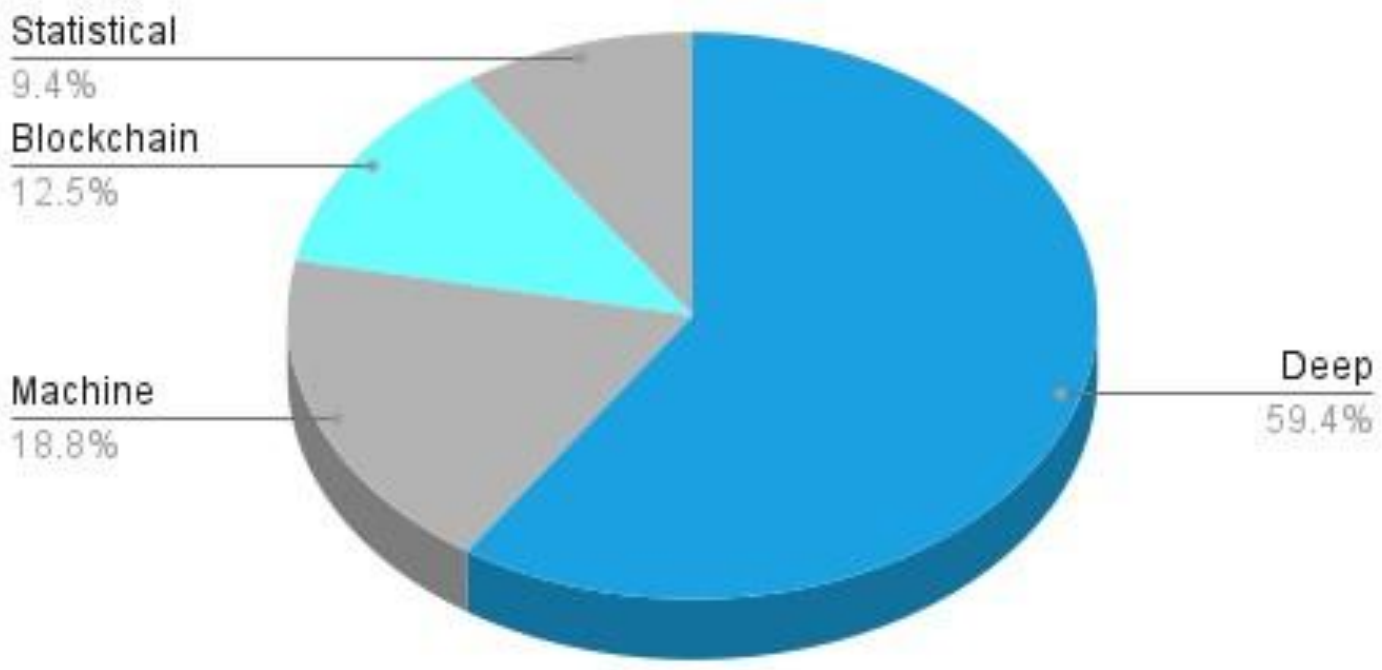

**Fig. 25**: The contribution percentage of various methods in Deepfake detection techniques.

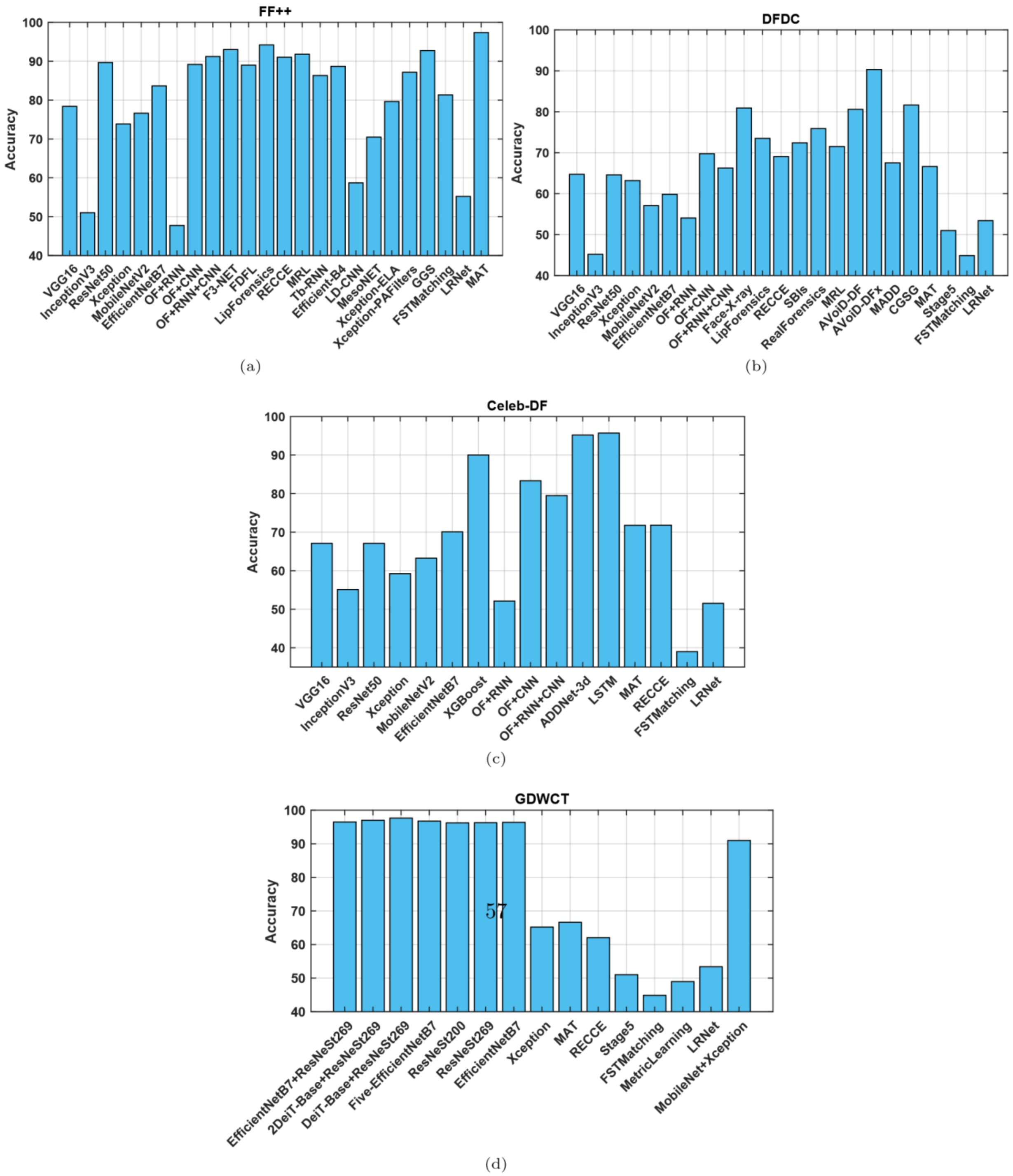

**Fig. 26**: Comprehensive comparison of heuristic and deep learning methods in Deepfake detection based on various datasets, a) FF++, b) DFDC, c) Celeb-DF, d) GDWCT

As illustrated in Figure 26, the leading models for detecting deepfakes within the FF++ dataset comprise MAT (Multi-Attention Transformer (Waseem et al., 2023; Li et al., 2023)), GGS (Generative Guided Supervision), LipForensics, and F3-NET. These innovative models utilize cutting-edge attention strategies, tailored feature extraction methods, and analyses specific to the domain (including frequency and temporal assessments) to attain remarkable effectiveness in uncovering deepfakes. Their proficiency in addressing various manipulation artefacts has led to consistent and dependable performance across the diverse types and qualities of manipulations found in the FF++ dataset. However, the performance of methods mostly depends on the characteristics of the datasets.

# 6 Analysis of Key Datasets and Their Characteristics

This section highlights the most recent and widely used datasets that have been generated specifically for deepfake detection using DL techniques. These datasets play a crucial role in training and evaluating deepfake detection models, enabling researchers to develop more robust and effective algorithms. Tables 8 and 9 offered a comprehensive overview of the diverse Deepfake datasets employed in the analysis conducted during the research. The table includes details on the specific datasets used, the number of real and fake samples, and other relevant information. Figures 27(a) and 27(b) show some image samples of five popular Deepfake datasets and the number of studies which were done based on the 11 most used datasets.

**Table 8**: Overview of popularly used datasets in Deepfake detection (section 1)

| No. | Author | Dataset | Links | Media | Sample size | Resolution |
|---|---|---|---|---|---|---|
| 1 | Yan et al. (2024) | DF40 | Hyper-link | Video, Image | 10M, 200K | × 256 |
| 2 | Lin et al. (2024) | VoxBlink2 | Hyper-link | Video | 10M | NA |
| 3 | Xie et al. (2024) | Codecfake | Hyper-link | Video | 1,058,216 | NA |
| 4 | Felouat et al. (2024) | eKYC-DF | Hyper-link | Video | 228K | NA |
| 5 | Zang et al. (2024) | CtrSVDD | Hyper-link | Video | 220,798 | kHz |
| 6 | Hou et al. (2024) | PolyGlot Fake | Hyper-link | Video | 15,238 | × 720 |
| 7 | Mittal et al. (2024) | GOTCHA | Hyper-link | Image | 673k | × 1024 |
| 8 | Kuckreja et al. (2024) | INDIFACE | Hyper-link | Video | 2072 | NA |
| 9 | Narayan et al. (2023) | DF-Platter | Hyper-link | Video | 133,260 | 360p, 720p |
| 10 | Khalid et al. (2021) | FakeAVCeleb | Hyper-link | Video | 20K | × 300 |
| 11 | Xie et al. (2022) | VFHQ | Hyper-link | Vedio | 16K | × 512 |
| 12 | Peng et al. (2021) | DFGC | Hyper-link | Image | 2228 | × 224<br>× 288 |
| 13 | He et al. (2021) | ForgeryNet | Hyper-link | Image & Video | 2.9M and 221,247 | × 256 |
| 14 | Li et al. (2020) | Celeb-DF | Hyper-link | Video | &<br>5639 | × 256 |
| 15 | Zi et al. (2020) | WildDeepfake | Hyper-link | Video | 7,314 | × 256 |
| 16 | Dong et al. (2020) | Vox-celeb1 | Hyper-link | Video | 100K | × 200 |
| 17 | Wu and Loy (2020) | DeeperForensics | Hyper-link | Video | 50K & 10K | × 1080 |
| 19 | Wang et al. (2020) | MEAD | Hyper-link | Video | 281,400 | × 1080 |
| 20 | Li et al. (2020) | Celeb-DFv2 | Hyper-link | Video | 6K | × 256 |

**Table 9**: Overview of popularly used datasets in Deepfake detection (section 2)

| No. | Author | Dataset | Links | Media | Sample size | Resolution |
|---|---|---|---|---|---|---|
| 21 | Rossler et al. (2019) | FaceForensics++ | Hyper-link | Image | 1.8M | 480p, 720p, 1080p |
| 22 | Dolhansky et al. (2020) | Deepfake Detection Challenge (DFDC) | Hyper-link | Video | 23,654; 104,500 | 1080 × 1920 |
| 23 | Lin et al. (2020) | Faceswap-GAN | Hyper-link | Video | 640 | 64 × 64 & 128 × 128 |
| 24 | Siarohin et al. (2019) | Tai-Chi-HD Dataset | Hyper-link | Video | 280 | 256 × 256 |
| 25 | Wen et al. (2019) | iPER | Hyper-link | Video | 241,564 | 256 × 256 |
| 26 | Dang et al. (2020) | DFFD: Diverse Fake Face Dataset | Hyper-link | Video | 1.8M | NA |
| 27 | Guarnera et al. (2022) | FFHQ | Hyper-link | Video | 70K | 1024 × 1024 |
| 28 | Jafar et al. (2020) | UADFV | Hyper-link | Image | 49 | 294 × 500 |
| 29 | Zi et al. (2020) | Deepfake-TIMIT | Hyper-link | Video | 620 | 64×64, 128× 128 |
| 30 | Zhou et al. (2021) | FaceForensics | Hyper-link | Video | 500K | 480p-1080p |
| 31 | Durall et al. (2019) | CelebA-HQ | Hyper-link | Image | 3K | 1024 × 1024 |
| 32 | Zi et al. (2020) | Fake Face in the Wild | Hyper-link | Image | 53K | 480p |
| 33 | Gu¨era and Delp (2018) | HOHA-based | Hyper-link | Video | 300 | 299 × 299 |
| 34 | Cao et al. (2018) | VGGFace2 | Hyper-link | Image | 3.31M | 256 × 256 |
| 35 | Chung et al. (2018) | VoxCeleb2 | Hyper-link | Video | 1.0M | 512 × 300 |
| 36 | Kaur et al. (2020) | SwapMe & FaceSwap | Hyper-link | Image | 3,305 | 1024 × 1024 |
| 37 | Parkhi et al. (2015) | VGGFace | Hyper-link | Image | 2.6M | 224 × 224 |

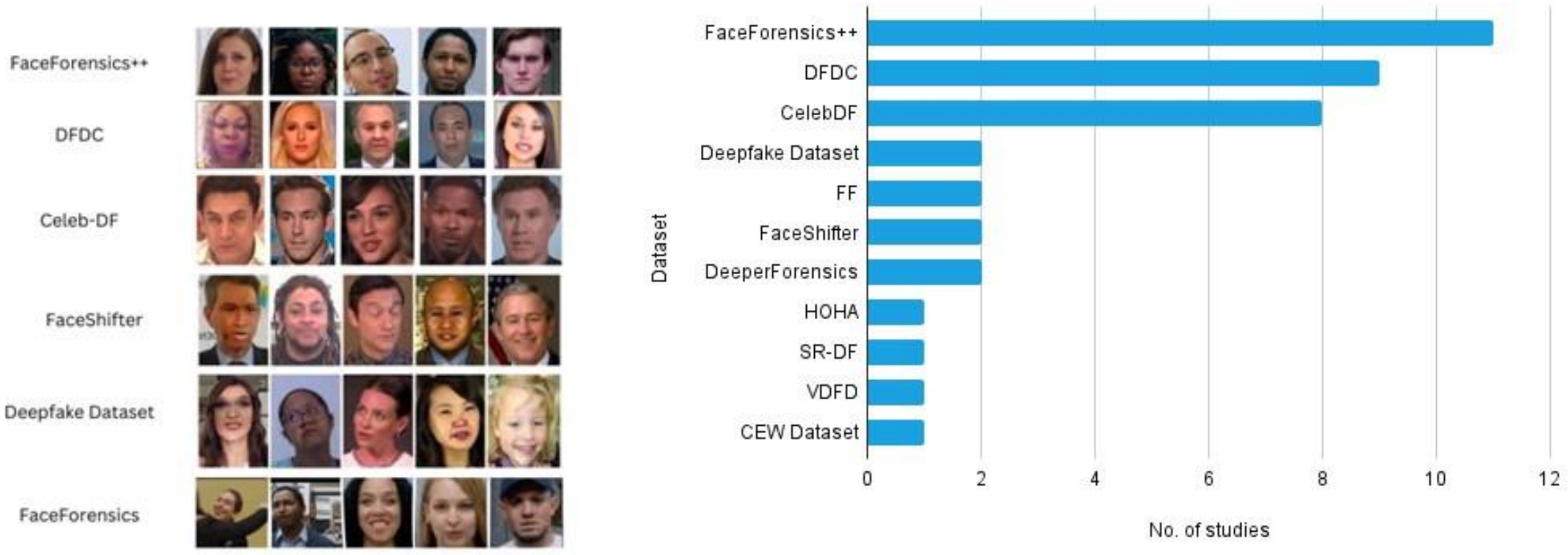

(a) (b)

**Fig. 27**: (a) Samples of some popular datasets, (b) List of datasets used in Deepfake-related studies

In Figure 28, a comparison is presented among prominent Deepfake datasets concerning the quantities of images and videos. While DF40 and ForgeryNet exhibit the largest number of videos, VFHQ stands out for offering a substantial amount of real and fake images, exceeding 10 million in total. This extensive dataset serves as a valuable resource for training and evaluating deepfake methods.

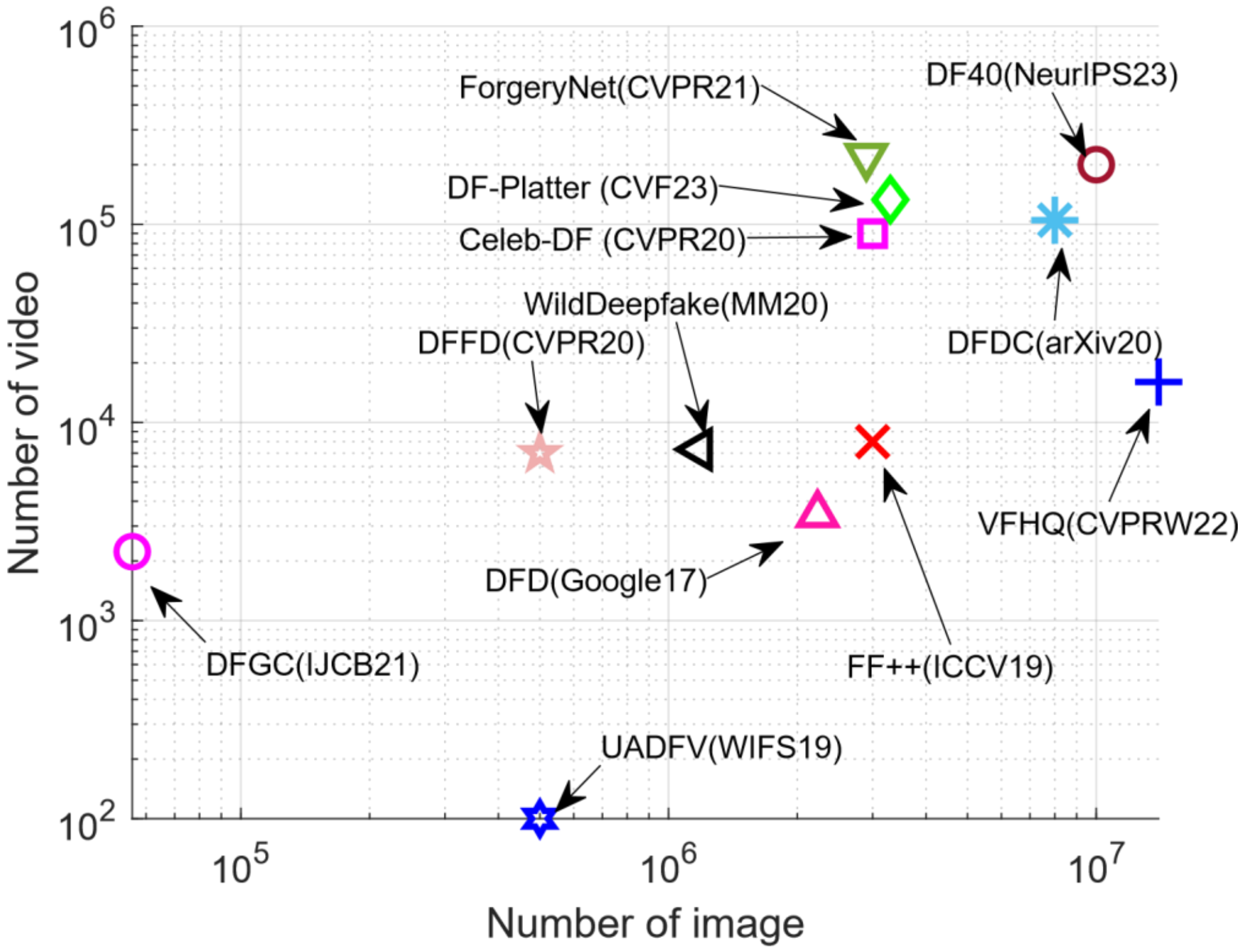

**Fig. 28**: Some popular Deepfake datasets comparison, number of images and videos.

## 7 Evaluation Metrics

Evaluation metrics hold an undeniably crucial and central position in the intricate and multifaceted world of deepfake detection systems, serving many significant functions for many compelling reasons. To begin with, these metrics act as essential and invaluable instruments that allow for a quantitative assessment of the performance and effectiveness of various deepfake detection models, which is paramount in understanding their capabilities. By establishing a consistent and standardized framework, these metrics enable researchers and developers to accurately measure the proficiency with which a model can distinguish between genuine content and content that has been manipulated or altered in some way. In addition, these metrics' importance extends to facilitating meaningful comparisons between a wide array of deepfake detection algorithms and models, thereby providing a clearer picture of their relative strengths and weaknesses. By adhering to a uniform set of evaluation standards, researchers and practitioners can conduct objective assessments and draw informed conclusions regarding the effectiveness of different methodologies. Furthermore, evaluation metrics serve as invaluable guiding lights, illuminating the path for optimizing and enhancing deepfake detection models. Through a careful and detailed examination of these performance indicators, researchers can identify specific areas where a model may be lacking and can subsequently implement targeted improvements aimed at increasing the accuracy of detection (Pei et al., 2024; Chen et al., 2019). Additionally, these metrics are foundational to the quality assurance processes in deepfake detection systems, ensuring that the models not only meet but also exceed established performance benchmarks and are fully equipped to identify altered content effectively.

One of the most critical aspects of evaluation metrics is their ability to assist in determining the most appropriate decision thresholds for deepfake detection, utilizing key metrics such as precision, recall, and F1-score to find an optimal balance between minimizing false positives and false negatives. By closely analyzing vital metrics related to accuracy and robustness, practitioners can thoroughly evaluate the practical applicability and real-world performance of Deepfake detection systems (Zhang et al., 2018), gaining valuable insights into how these systems function in everyday scenarios. Ultimately, these metrics serve as a powerful catalyst for advancing research within the field, providing a standard benchmark that facilitates the evaluation of innovative methodologies and techniques. They empower researchers to gain a comprehensive understanding of the strengths and limitations inherent in existing approaches, inspiring the continuous evolution and development of more effective and reliable detection strategies. Table 10 lists the significant performance metrics in Deepfake detection methods and their applications.

Table 11 unfolds a detailed and thorough examination of the performance evaluation results. It showcases various prominent models meticulously applied to the intricate Face Swapping task, utilizing the sophisticated FF++ dataset as a benchmark for their capabilities (Pei et al., 2024). Amongst this curated selection of models, WSC-Swap emerges as a remarkable standout, distinguished by its innovative methodology, which cleverly captures not only the external facial attribute information but also delves deep into the internal identity nuances through the deployment of two distinctly functioning encoders that work in harmony. This unique approach significantly enhances the model's ability to maintain the integrity of identity while simultaneously ensuring that the accuracy of the facial pose is preserved with remarkable precision. However, despite these notable strengths and advantages, WSC-Swap (Ren et al., 2023) does experience some limitations, particularly regarding the metrics associated with facial expression errors, which reveal a certain degree of suboptimal performance in this essential area of functionality. The model's evident proficiency in upholding identity and facial pose, when placed side-by-side with its challenges in effectively reproducing nuanced facial expressions, underscores the complex trade-offs and multifaceted challenges inherently present in the realm of advanced facial manipulation tasks. The intricate balance between these competing strengths and weaknesses paints a vivid picture of the ongoing evolution and refinement that characterizes the field of facial recognition and manipulation technologies.

# 8 Impact of Deepfake on Society and Individuals

The examination of the psychological and societal consequences of deepfakes reveals a complex interplay of both positive and negative emotional effects. On the one hand, deepfakes have demonstrated their powerful ability to evoke comfort and nostalgic feelings by resurrecting beloved individuals and celebrities (Wajid and Wajid, 2021). This technological capability holds promise for creating meaningful connections and preserving memories. However, deepfakes also pose significant threats that extend beyond public figures to impact ordinary individuals. For instance, instances have been reported where voice deepfakes were employed to deceive and defraud, resulting in substantial financial losses (Bateman, 2022). Furthermore, an in-depth analysis of deepfake conversations on Reddit, spanning from 2018 to 2021, highlighted concerns surrounding the believability of deepfakes and the moderation practices of online platforms. Surprisingly, the Reddit discussions revealed a pro-deepfake sentiment, fostering a community that actively creates, shares, and even builds marketplaces for deepfake artefacts, often disregarding the potential consequences (Kalpokas and Kalpokiene, 2022). Alarming cyber security reports from 2019 have predicted that a staggering 96% of all deepfakes will be pornographic in nature and entail other forms of harmful outcomes (Gamage et al., 2022). Counterfeit material presents substantial risks, encompassing identity theft, dissemination of revenge pornography, fraudulent activities, and potential threats to national security (Dagar and Vishwakarma, 2022; Boutadjine et al., 2023; Mirsky and Lee, 2021). The growing frequency and severity of deepfake occurrences have garnered widespread attention from scholars, policymakers, and the public, prompting endeavours to comprehend their characteristics, ramifications, and strategies for identification and mitigation (Tolosana et al., 2020; Figueira and Oliveira, 2017; Caldwell et al., 2020; Weikmann and Lecheler, 2023). These findings underscore the urgent need for comprehensive measures to address the psychological and societal impact of deepfakes. It is crucial to develop robust regulatory frameworks, promote public awareness and education, and advance deepfake detection technologies to safeguard individuals from the detrimental effects of malicious deepfake usage.

**Table 11**: A comparison among various Deepfake detection models trained by CelebA-HQ, FFHQ, VGGFace, VGGFace2, VoxCeleb2 and tested by FF++.

| Methods | Training | ID Ret.(%) | Exp Error | Pose Error | FID | Authors |
|---|---|---|---|---|---|---|
| FaceShifter | CelebA-HQ, FFHQ, VGGFace | 97.4 | 2.1 | 3.0 | - | (Li et al., 2019) |
| SimSwap | VGGFace2 | 93.0 | - | 1.5 | - | (Chen et al., 2020) |
| FaceInpainter | CelebA-HQ, FFHQ, VGGFace | 98.0 | - | 2.2 | - | (Li et al., 2021) |
| HifiFace | VGGFace2 | 98.5 | - | 2.6 | - | (Wang et al. , 2021) |
| RAFSwap | CelebA-HQ | 96.7 | 2.9 | 2.5 | - | (Xu et al., 2022b) |
| HFSwap | FFHQ | 90.0 | 2.8 | 2.5 | - | (Xu et al., 2022) |
| DiffSwap | FFHQ | 98.5 | 5.6 | 2.5 | 2.2 | (Zhao etal., 2023) |
| FlowFace | CelebA-HQ, FFHQ, VGGFace2 | 99.3 | - | 2.7 | - | (Zeng etal., 2023) |
| FlowFace++ | CelebA-HQ, FFHQ, VGGFace2 | 99.5 | - | 2.2 | - | (Zhang et al., 2023) |
| StyleIPSB | FFHQ | 95.0 | 2.2 | 3.6 | - | (Jiang etal., 2023) |
| StyleSwap | VGGFace, Vox-Celeb2 | 97.0 | 5.3 | 1.6 | 2.7 | (Xu et al., 2022c) |
| WSC-Swap | CelebA-HQ, FFHQ, VGGFace | 99.9 | 5.0 | 1.5 | - | (Ren et al., 2023) |

The emergence of deepfakes has sparked concerns about their impact on public trust, media credibility, and democratic processes. Instances where deepfakes have been used to spread disinformation have underscored the potential consequences for societal stability. For example, a deepfake video was posted on a Ukrainian news site featuring Ukrainian President Volodymyr Zelensky allegedly urging soldiers to surrender during the war with Russia (Analytica, 2019). The dissemination of such fabricated content through deepfakes poses a significant challenge, as these manipulated videos can be difficult to distinguish from authentic footage, especially for the untrained eye. The implications of deepfakes on democratic processes are particularly troubling. Manipulated videos through deepfakes can contribute to the spread of disinformation, potentially swaying public opinion and undermining the integrity of the electoral system. Detecting and countering deepfakes in the context of political campaigns becomes increasingly important to maintain the credibility of democratic processes (Diakopoulos and Johnson, 2021).

In a study by researchers (Allyn, 2022), the interconnected relationship between deepfakes and personalized micro-targeting (PMT) techniques is examined. PMT involves collecting personal data to tailor political messages and advertisements to specific individuals. Integrating deepfakes with PMT introduces new possibilities for influencing targeted segments of the electorate. The realistic and persuasive nature of deepfakes can be leveraged by political actors to manipulate public opinion, potentially disrupting democratic processes. These developments highlight the urgent need for robust safeguards and countermeasures to address the adverse effects of deepfakes on public trust, media credibility, and democratic systems. Enhancing media literacy is crucial to empower individuals to critically evaluate the authenticity of digital content. Technological advancements in deepfake detection and verification tools are essential for effective identification and debunking of manipulated media. Furthermore, strengthening legal and ethical frameworks is necessary to tackle the potential misu Table 5 provides a comprehensive overview of the legal and regulatory landscape concerning deepfakes. It summarizes the current regulations, guidelines, and policies implemented by various jurisdictions to address the challenges posed by deepfake technology (Dobber et al., 2021; Zafar and Wajid, 2019). The table highlights the diverse approaches taken by different countries and regions to mitigate the potential risks associated with deepfakes, such as misinformation, privacy violations, and the spread of harmful content.se of deepfakes, particularly in content of elections and political campaigns.

**Table 12**: Deepfake-related Laws and Policies

| **Country** | **Deepfake-related Laws and Policies** |
|---|---|
| United States (2020) | California AB-730: Prohibits distributing deepfake videos within 60 days of an election without disclosure https://leginfo.legislature.ca.gov/faces/billTextClient.xhtml?bill id=201920200AB730.<br>Virginia Code 18.2-386.2: Makes creating and distributing deepfake pornographic content illegal https://law.lis.virginia.gov/vacode/title18.2/chapter8/ section18.2-386.2/.<br>DEEPFAKES Accountability Act: Proposed federal legislation to combat malicious use of deepfakes https://www.congress.gov/bill/116th-congress/ house-bill/3230. |
| United Kingdom (2020) | The Audiovisual Media Services Regulations 2020: Requires clear labelling of deepfake content on online platforms https://www.legislation.gov.uk/uksi/2020/595/contents/made.<br>The Online Safety Bill: Proposes measures to tackle online harms, including misinformation and disinformation through deepfakes https://bills.parliament.uk/bills/3058. |
| Singapore (2019) | Protection from Online Falsehoods and Manipulation Act: Provides powers to combat online falsehoods, including deepfakes https://sso.agc.gov.sg/Act/POFMA2019.<br>Personal Data Protection Act: Addresses privacy concerns related to deepfake creation and distribution https://sso.agc.gov.sg/Act/PDPA2012. |
| European Union (2021) | EU Regulation on AI: Proposes rules for AI technologies, including provisions for deepfake detection and transparency https://eur-lex.europa.eu/eli/reg/2021/xxx/oj.<br>EU Digital Services Act: Seeks to regulate online platforms hosting deepfake content and mandate content moderation efforts https://eur-lex.europa.eu/eli/reg/2020/xxx/oj. |

**Table 12 Continued from previous page**

| Country | Deepfake-related Laws and Policies |
|---|---|
| India (2022) | Information Technology (Intermediary Guidelines and Digital Media Ethics Code) Rules, 2021: Imposes obligations on platforms to address deepfake content https://www.meity.gov.in/.<br>The Protection of Personal Information Bill: Aims to safeguard against misuse of personal data, including deepfake-related privacy concerns https://www.prsindia.org/. |
| Australia (2022) | Enhancing Online Safety (Non-consensual Sharing of Intimate Images) Act 2018: Addresses deepfake-related revenge porn issues https://www.legislation.gov.au/Series/C2018A00077. |
| | Criminal Code Amendment (Sharing of Abhorrent Violent Material) Act 2019: Criminalizes sharing violent deepfake content https://www.legislation.gov.au/Series/C2019A00047. |

By adopting a comprehensive approach encompassing technological advancements, educational initiatives, and policy interventions, we can strive to mitigate the psychological and societal consequences of deepfakes. Such efforts are vital for maintaining public trust, safeguarding democratic processes, and preserving the integrity of our media landscape in the face of this evolving threat.

# 9 Challenges and Future Directions

The realm of deepfake detection has witnessed significant advancements in recent years; however, several challenges persist, necessitating continuous exploration and development to ensure efficacy in the face of an evolving technological landscape. This paper delves into the critical aspects shaping the future of deepfake detection, highlighting both the obstacles and the promising avenues for progress. One of the primary challenges lies in the perpetual evolution of deepfake techniques. As deepfake generation methods become increasingly sophisticated, detection algorithms must adapt and improve at an equivalent pace. This necessitates ongoing research and development efforts focused on devising novel techniques capable of identifying and mitigating the latest manipulation approaches (Zhao et al., 2021; Caldelli et al., 2021; Wang et al., 2020; Cozzolino et al., 2018; Marra et al., 2019). Additionally, enhancing the generalizability of detection models is paramount. These models must be adept at handling diverse deepfake variations, datasets, and real-world scenarios to guarantee their effectiveness in practical applications.

The application of deepfakes is expanding beyond the realm of image manipulation. For instance, the burgeoning field of the metaverse proposes the use of deepfake technology to create personalized avatars that closely resemble individual users. This advancement presents researchers with a critical decision: object-specific versus object-agnostic modelling. Object-specific models excel in situations demanding high precision and accuracy for specific tasks, such as targeted applications where fine-grained control and fidelity are crucial. However, object-agnostic models offer immense potential for broader innovation. These models strive to develop an understanding and manipulation capability encompassing a wide range of objects or subjects, transcending the limitations imposed by specific categories. This broader approach fosters the development of versatile and adaptable systems with diverse applications and implications, extending beyond manipulating specific objects or individuals.

Another significant challenge emerges in the form of adversarial perturbations. In this scenario, malicious actors deliberately manipulate content to deceive deepfake detection models. Countering this threat necessitates the development of more robust and resilient detection algorithms. Researchers are actively exploring various techniques, including advanced regularization methods, adversarial training approaches, and novel defence mechanisms, with the objective of creating detection models that exhibit superior resistance to such attacks and maintain high accuracy even in the presence of sophisticated manipulations
(Gandhi and Jain, 2020; Hussain et al., 2021; Carlini and Farid, 2020; Yang et al., 2021; Yeh et al., 2020).

Furthermore, existing models often exhibit limitations in generalizability, particularly in cross-forgery and cross-dataset settings. Models trained on a specific type of forgery might not be effective against others, and focusing solely on exploiting vulnerabilities in specific deepfake generation pipelines may not be beneficial in adversarial settings where attackers actively strive to conceal their techniques. This necessitates a shift towards developing more robust, scalable, and adaptable approaches in future research endeavours.

In conclusion, the future of deepfake detection hinges on effectively addressing challenges related to adversarial perturbations, object-agnostic modelling, cost-effectiveness, whole-face analysis, and the ability to distinguish AI-generated content from human-made content. As researchers delve deeper into these areas, deepfake detection models will evolve to become more robust, versatile, and essential for safeguarding against the potential harms of deepfakes in a world increasingly driven by AI.

# 10 Concluding remarks

The advancement of deepfake technology has undergone remarkable development, fueled by breakthroughs in artificial intelligence and machine learning, particularly through innovations like GANs, VAEs, and diffusion models. These techniques have paved the way for the creation of strikingly realistic and persuasive deepfake materials. The paper underscores cutting-edge progress in deepfake detection, featuring methodologies that utilize attention mechanisms, multi-modal assessments, and hybrid frameworks to boost detection precision. Nonetheless, the realm of deepfake detection grapples with significant hurdles, especially because of the continuously advancing generative strategies that yield high-fidelity content with minimal detectable imperfections.

This review encompassed around 400 scholarly works, illustrating the extensive breadth of research in this domain. The main obstacles involve upholding generalizability across varied datasets, managing highquality synthetic media devoid of conventional manipulation signs, and ensuring temporal consistency in video sequences. The resource-intensive nature of training deepfake models also constitutes a major challenge. Ethical dilemmas like privacy breaches, misinformation, and harmful applications, including identity theft and political tampering, highlight the necessity for stringent regulatory measures to oversee the deployment of deepfake technology.

Solutions proposed in this paper include:

- Crafting more resilient detection frameworks capable of adjusting to fresh data and unrecognized manipulations. Contemporary machine learning methods are pivotal in tackling these issues.
- Transfer learning empowers detection models to tap into insights from previously trained models, facilitating quicker adaptation to novel deepfake variations with sparse data.
- Domain adaptation strategies are vital for enhancing model efficacy across distinct datasets, tackling generalizability by diminishing the gap between source and target domains.
- Foundation models, serving as expansive pre-trained frameworks that can be fine-tuned for specific objectives, offer a robust foundation for deepfake detection, enabling swift adaptation and enhanced performance across diverse manipulation methods. Reinforcement learning can be employed to refine detection tactics, where models develop optimal detection strategies through ongoing engagement with generated deepfakes.

- Additionally, contemporary methods like self-supervised learning can be harnessed to boost feature extraction from limited labelled datasets, strengthening detection reliability.
- The integration of explainable AI (XAI) methodologies is essential for boosting transparency and facilitating a clearer understanding and trust in the model's predictions by human experts.
- Adversarial training techniques are advised to bolster resilience against advancing generative tactics, as these strategies can help models better endure adversarial exploits designed to bypass detection.
- Furthermore, establishing ethical standards and regulations can alleviate the potential damage caused by misuse.

Sophisticated deep learning strategies, especially CNN-based techniques, have shown potential in recognizing deepfakes through advanced algorithms and feature extraction. Models such as YOLOCNN-XGBoost, hybrid VGG16 configurations, MesoNet, and integrations with other architectures have achieved notable success. The interpretability of these models is further augmented through techniques like Layer-wise Relevance Propagation (LRP) and Local Interpretable Model-agnostic Explanations (LIME), which enhance trust in deepfake detection frameworks by clarifying their decision-making processes.

As we gaze into the horizon, the trajectory of deepfake detection relies on developing sophisticated models that are effective and versatile enough to adapt to various types of manipulations. Synergistic collaborations among academics, industries, and governmental bodies will be vital in establishing ethical frameworks for applying deepfake technology, ensuring it is harnessed for creative and constructive ends while curbing any potential risks. The evolution of detection systems will increasingly emphasize realtime capabilities to swiftly pinpoint altered content, which is especially critical for live broadcasts and video calls. Moreover, federated learning techniques are likely to gain traction, allowing for training on decentralized data while safeguarding user privacy, a key factor in crafting resilient models from a multitude of data streams. Additionally, the promise of quantum machine learning may enhance both the training and detection timelines, leading to greater efficiency in the systems employed.

# Declarations

**Conflict of interest:** All authors certify that they have no affiliations with or involvement in any organization or entity with any financial interest or non-financial interest in the subject matter or materials discussed in this manuscript.
**Author contributions:** Amir Gandomi: Investigation, Editing, Funding acquisition, and Writing Review. Other Authors contributed equally to this work. All authors reviewed the manuscript.